\documentclass[lettersize,journal]{IEEEtran}
\usepackage{amsmath,amsfonts}
\usepackage{algorithmic}
\usepackage{tikz}
\usepackage{booktabs}
\usepackage{algorithm}
\usepackage{array}
\usepackage[caption=false,font=normalsize,labelfont=sf,textfont=sf]{subfig}
\usepackage{textcomp}
\usepackage{stfloats}
\usepackage{url}
\usepackage{verbatim}
\usepackage{graphicx}
\usepackage{cite}

\usepackage[colorlinks,
urlcolor=blue,
]{hyperref}

\begin{document}

\title{ Multiview Textured Mesh Recovery by Differentiable Rendering}

\author{Lixiang Lin, Jianke Zhu~\IEEEmembership{Senior Member,~IEEE}, and Yisu Zhang

\thanks{Lixiang Lin, Jianke Zhu and Yisu Zhang are with the School of Computer Science and Technology, ZheJiang University, 38 Zheda Road, Hangzhou, China. Jianke Zhu is also with the Alibaba-ZheJiang University Joint Research Institute of Frontier Technologies. \protect\\
Email: \{lxlin, jkzhu, 22121124\}@zju.edu.cn;}
\thanks{Jianke Zhu is the Corresponding Author.}
}

\markboth{Journal of \LaTeX\ Class Files,~Vol.~14, No.~8, August~2021}%
{Shell \MakeLowercase{\textit{et al.}}: A Sample Article Using IEEEtran.cls for IEEE Journals}


\maketitle

\begin{abstract}
Although having achieved the promising results on shape and color recovery through self-supervision, the multi-layer perceptrons-based methods usually suffer from heavy computational cost on learning the deep implicit surface representation. Since rendering each pixel requires a forward network inference, it is very computationally intensive to synthesize a whole image. To tackle these challenges, we propose an effective coarse-to-fine approach to recover the textured mesh from multi-views in this paper. Specifically, a differentiable Poisson Solver is employed to represent the object's shape, which is able to produce topology-agnostic and watertight surfaces. To account for depth information, we optimize the shape geometry by minimizing the differences between the rendered mesh and the predicted depth from multi-view stereo. In contrast to the implicit neural representation on shape and color, we introduce a physically-based inverse rendering scheme to jointly estimate the environment lighting and object's reflectance, which is able to render the high resolution image at real-time. The texture of reconstructed mesh is interpolated from a learnable dense texture grid. We have conducted the extensive experiments on several multi-view stereo datasets, whose promising results demonstrate the efficacy of our proposed approach. The code is available at \href{https://github.com/l1346792580123/diff}{https://github.com/l1346792580123/diff}. 
\end{abstract}


\section{Introduction}
\IEEEPARstart{R}{ecovering} the textured mesh from multiple views is a fundamental problem in computer vision, which is essential to a large amount of real-world applications, including virtual reality, robotics and cultural heritage digitization. 

Generally, the de-facto pipeline of 3D reconstruction is to firstly obtain the point cloud by multi-view stereo~\cite{DBLP:conf/bmvc/mvs11, DBLP:conf/eccv/colmap16, DBLP:journals/mva/mvs12}. Then, the implicit surface~\cite{DBLP:conf/sgp/psr06} is fitted from the scattered points and normals, which is further triangulated into mesh through an efficient marching cube algorithm~\cite{DBLP:conf/siggraph/Marchingcubes87}. While the overall pipeline is straightforward and robust in real world scenarios, the resulting mesh may be suboptimal during the accumulated loss in reconstruction. Usually, the textures are sampled from multi-view images by back-projecting them onto the mesh, which combine the factors like diffuse, specular and environment lighting. This leads to the unrealistic rendering results in novel views.  
\begin{figure}
	\centering
	\begin{tabular}{@{\hskip4pt}c@{\hskip4pt}@{\hskip4pt}c@{\hskip4pt}}
	    \includegraphics[width=0.47\linewidth,clip]{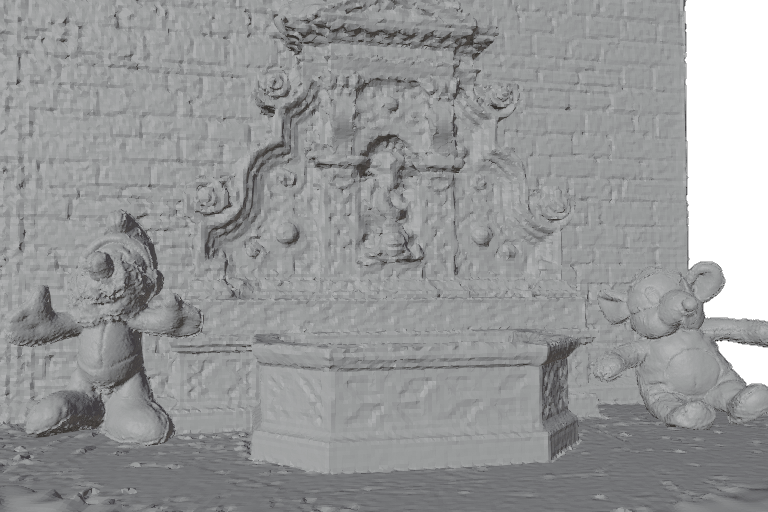} &
	    \includegraphics[width=0.47\linewidth]{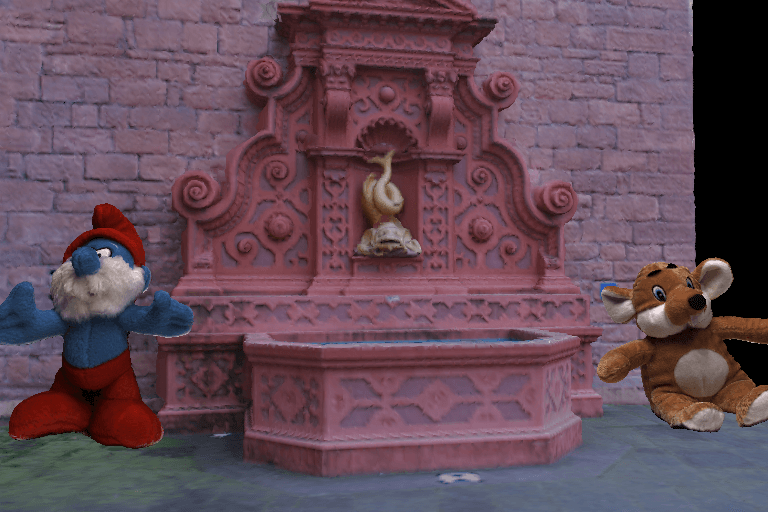} \\
	    
	    \includegraphics[width=0.47\linewidth,clip]{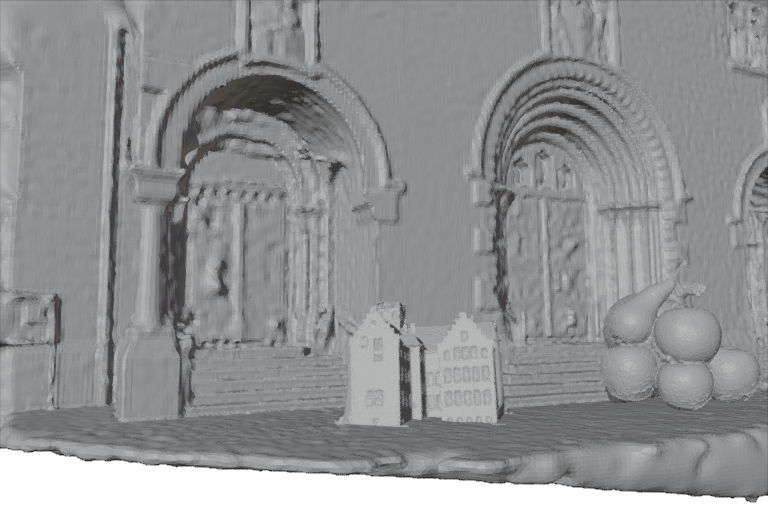} &
	    \includegraphics[width=0.47\linewidth]{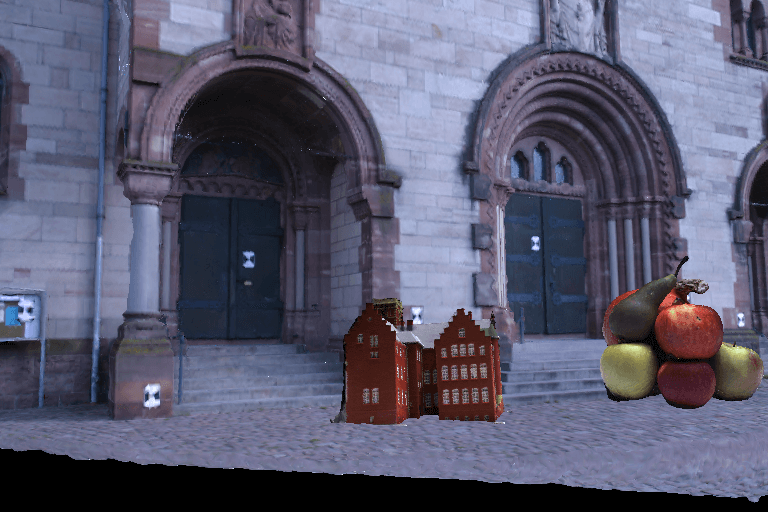} \\
	\end{tabular}
	\caption{The example reconstruction results on DTU and EPFL dataset. Our proposed method is able to reconstruct the high-quality textured mesh that can be used in standard computer graphics pipeline.}
	\label{fig:teaser}
\end{figure}

Recently, inverse rendering~\cite{DBLP:conf/cvpr/handheld20, DBLP:journals/cgf/psdr21} becomes a promising approach to recover the textured mesh with details for real-time rendering. Specifically, the photometric loss is backpropagated to the triangulated mesh through differentiable rendering~\cite{DBLP:conf/iccv/softras19, ravi2020pytorch3d, DBLP:conf/cvpr/sdfdiff20, Laine2020diffrast}, which makes it feasible to predict shape and reflectance by an inverse rendering optimization (or analysis-by-synthesis). Due to the ambiguity between geometry and appearance, the optimization in inverse rendering is highly under-constrained. To address this issue, it usually makes an assumption of single light source or the known positions of all the light sources, which is difficult to hold in real scenarios. Moreover, it is hard for the explicit mesh to deal with the complex topology changes. As it is easy to stuck at the local optima, a rough initial mesh model is usually generated by multiview stereo~\cite{DBLP:conf/eccv/colmap16, DBLP:journals/mva/mvs12}.

In contrast to the conventional approaches consisting of several stages, the implicit neural representation-based methods~\cite{DBLP:conf/nips/idr20, mvsdf21} directly learn the multi-layer perceptrons (MLP) from multiple images, which is able to recover shape and color in an end-to-end manner. The scene geometry is represented as zero level set, and the appearance is represented as the surface light field. Moreover, the neural network is trained with self-supervision through backpropagating the color consistency loss. Similarly, the triangulated mesh is extracted by marching cube algorithm via evaluating the signed distance function. Since the implicit MLP representation is not straightforward and its gradient decreases as the layer moves forward, it is very time-consuming to train the deep neural network. Typically, it takes several hours to reconstruct the shape and appearance using the implicit neural representation. More importantly, the color of each pixel is represented by the learned MLP, which is computationally intensive to render whole image. This greatly hinders them from a large amount of real-world applications.

To address the above limitations, we propose an effective coarse-to-fine approach to recover the textured mesh from multi-view in this paper. Instead of deforming a sphere like the conventional methods~\cite{DBLP:conf/nips/idr20}, we make use of visual hull to obtain an accurate initial mesh. Moreover, we take advantage of a differentiable Poisson Solver~\cite{Peng2021SAP} to represent the geometry, which is able to produce topology-agnostic and watertight surfaces. Furthermore, we introduce a physically-based inverse rendering scheme to jointly estimate the lighting and reflectance of the objects rather than the computationally intensive MLP approach, which is able to render the high resolution image in real-time. We estimate the shape geometry by minimizing the difference between the rendered mesh and the predicted depth by multi-view stereo. The texture of our reconstructed mesh is represented by a dense learnable texture grid, which can be optimized during inverse rendering. Fig.~\ref{fig:teaser} shows some example reconstruction results.
In summary, the main contributions of this paper are in the following.

\begin{itemize}
    \item We propose an effective coarse-to-fine approach to recover the textured mesh from multi-view images. 
    \item We incorporate an oriented point clouds-based geometry representation with a grid-based texture representation. This is not only lightweight and interpretable, but also greatly reduces the training time.
    \item Instead of performing neural rendering through MLP, we adopt physically-based rendering to generate the image from environment map and textured mesh. Our reconstructed textured mesh is rendered in real time, which can be used in computer graphics pipeline directly.
    \item Extensive experiments are performed on several multi-view stereo datasets. The experiments compared against state-of-the-art methods show good qualitative and quantitative results.
\end{itemize}

\section{Related Works}

During past decades, a surge of research efforts have been devoted to 3D reconstruction using multiple images. The recent approaches can be roughly divided into three categories, including Multi-view-stereo (MVS), inverse rendering and implicit neural representation. We review them in the following.

\subsection{Multi-view Stereo}

Multi-View Stereo methods try to estimate the depth map by matching feature points across different views. In general, most of them assume that the appearances of a surface point are consistent in all visible views~\cite{DBLP:conf/eccv/colmap16, DBLP:journals/mva/mvs12, DBLP:journals/tcsv/mvs12, DBLP:journals/tcsv/mvs22}. They evaluate the matching cost of image patches for all depth hypotheses in order to find the best one. Li \textit{et al}.~\cite{DBLP:journals/tcsv/mc11} propose a marker-less shape and motion capture approach for multi-view video sequences. 3D coordinates can be obtained by triangulating the correspondences through matching the image patches across all region of interests uniformly sampled or propagated from neighboring pixels and adjacent views~\cite{DBLP:conf/bmvc/mvs11, DBLP:journals/pami/mvs10}. Depth fusion and implicit surface reconstruction~\cite{DBLP:conf/sgp/psr06} are required to extract the watertight mesh from point clouds, where the surface details may be smoothed due to depth fusion. Moreover, these meshes are usually texture-less. Although the color can be estimated from the input images using camera projection, the rendered images are not fidelity enough to be viewed from free perspective. 

Recently, the learning-based MVS methods have received a bit attentions~\cite{DBLP:conf/cvpr/deepmvs18, DBLP:conf/eccv/mvsnet18, DBLP:conf/cvpr/recurmvsnet19}. DeepMVS~\cite{DBLP:conf/cvpr/deepmvs18} reprojects images onto 3D plane-sweeping volumes and performs intra-volume aggregation, where the hand-crafted features are replaced by deep ones. MVSNet~\cite{DBLP:conf/eccv/mvsnet18} warps deep features into the reference camera frustum to build a cost volume via differentiable homographies. R-MVSNet~\cite{DBLP:conf/cvpr/recurmvsnet19} replaces the 3D CNNs regularization module with 2D recurrent network. Despite these promising results, it is still difficult to build the correspondences in those texture-less pixels or non-Lambertian regions. The reliability of the estimated depth map is also important for practical applications. Su \textit{et al}.~\cite{DBLP:journals/tcsv/de22} propose an uncertainty-guided network and an uncertainty-aware loss to perceive the uncertainty in an unsupervised manner.

In order to obtain the fine detailed results, a coarse-to-fine framework is often used. Zhang \textit{et al}.~\cite{DBLP:conf/bmvc/mvs20} propose a coarse-to-fine framework to integrate pixel-wise occlusion information, which significantly improves the depth accuracy. Zhang \textit{et al}.~\cite{DBLP:journals/tcsv/mvs21} and He \textit{et al}.~\cite{DBLP:journals/tcsv/mvs212} incorporate the coarse-to-fine strategy to enable the network to estimate the high-resolution depth maps. Liu \textit{et al}.~\cite{DBLP:conf/cvpr/dist20} propose a coarse-to-fine differentiable sphere tracing to reconstruct 3D shapes from multi-view images. Liu \textit{et al}.~\cite{DBLP:journals/tip/spunet22} propose a coarse-to-fine reconstruction framework for point cloud upsampling. These methods employ the coarse-to-fine strategy to reduce computation cost. Coarse module results are sent to fine one as input to get the detailed results. Differently from these coarse-to-fine methods, we suggest a coarse-to-fine strategy to greatly reduce optimization time. In our proposed approach, the coarse module results are upsampled and optimized in fine module.

\subsection{Inverse Rendering}

By taking advantage of differentiable renderer~\cite{ravi2020pytorch3d, Laine2020diffrast}, ~\cite{DBLP:conf/iccv/softras19,DBLP:conf/cvpr/handheld20, DBLP:journals/cgf/psdr21} try to estimate the object's intrinsic color and geometry by inverse rendering, where the gradients can be backpropagated to geometry. Therefore, the objects' geometry and appearance can be recovered by minimizing the differences between the synthesized photo and input image. 

Rasterization is the key operation in rendering pipeline. Liu \textit{et al}.~\cite{DBLP:conf/iccv/softras19} propose a differentiable renderer for image-based shape fitting. The rendering is treated as a differentiable aggregating process that fuses the probabilistic contributions of all mesh triangles with respect to the rendered pixels. Wang \textit{et al}.~\cite{DBLP:journals/tog/dss19} present a differentiable renderer for point cloud that is considered as a disk. Moreover, the discontinuities are approximated by a linear function. Zhang \textit{et al}.~\cite{DBLP:journals/tcsv/pcu21} propose a progressive point cloud upsampling framework supervised by the point-based differentiable rendering. Jiang \textit{et al}.~\cite{DBLP:conf/cvpr/sdfdiff20} propose a differentiable renderer for shape optimization using signed distance functions (SDFs), where the gradients can be backpropagated to the whole SDF grid by regularization. 

As the soft rasterization operation is an approximate solution, which does not provide the gradient with respect to variables other than pixel coordinates. This may lead to the inferior reconstruction results. Li \textit{et al}.~\cite{DBLP:journals/tog/diffmcL18} present a Monte Carlo differentiable renderer that produces the unbiased gradients through edge sampling. Similarly, Luan \textit{et al}.~\cite{DBLP:journals/cgf/psdr21} propose an analysis-by-synthesis pipeline for high-quality geometry reconstruction and spatially varying reflectance. As is well-known, recovering the geometry and appearance from 2D images is a highly under-constrained problem, in which the pixel intensity is affected by lots of factors like lighting, materials, occlusions, and etc. Most of existing inverse rendering methods assume that there is a single light source or the directions of all the light sources are known. This may rarely happen in the real-world applications. In addition, it is difficult to deal with the complex topology changes using the inverse rendering optimization, where the gap between the initial geometry and target cannot be very large. To avoid this issue, MVS is usually employed to obtain a good initial shape.

\subsection{Implicit Neural Representation}

Implicit neural representation directly estimates the objects' geometry from the input image through minimizing the photometric loss, which has achieved the encouraging results in 3D reconstructions and view synthesis. Sitzmann \textit{et al}.~\cite{DBLP:conf/nips/srn19} employ LSTM to represent the scene by simulating the ray marching process. Mildenhall \textit{et al}.~\cite{DBLP:conf/eccv/nerf20} represent the scene in terms of volume density and view-dependent radiance, where the color of each pixel is obtained by accumulating the grid having the ray passed through. Due to the resolution limitations in volume rendering, the geometry extracted by the marching cube algorithm~\cite{DBLP:conf/siggraph/Marchingcubes87} is usually rough. Instead of using volumetric density, Yariv \textit{et al}.~\cite{DBLP:conf/nips/idr20} employ neural networks to represent scenes through SDF and light field implicitly, where color is only calculated at the intersection. Moreover, the two implicit networks are trained by the loss of color and silhouettes. Zhang \textit{et al}.~\cite{DBLP:conf/cvpr/physg21} use mixtures of spherical Gaussians to represent the bidirectional reflectance distribution function (BRDF) and environmental illumination for physically-based rendering. Zhang \textit{et al}.~\cite{mvsdf21} leverage the feature consistency in stereo matching to fit the implicit surface. Oechsle \textit{et al}.~\cite{unisurf21} make use of both surface rendering and volume representation, which obtain the accurate reconstruction results without masks. Although the implicit representation has the merits of watertight and topology-agnostic, it tends to produce the smoothed surface. Due to the implicit representation using neural networks, it takes long time to converge.  Since the color value of every pixel needs to be calculated by a neural network, it is quite computationally expensive to render an image. 

Recently, Peng \textit{et al}.~\cite{Peng2021SAP} propose a differentiable Poisson Solver to represent mesh with the oriented point clouds. The differentiable Poisson Solver bridges the explicit 3D point representation with 3D mesh via the implicit indicator field. In contrast to neural implicit representation of MLP, the oriented point cloud is more interpretable and lightweight, which accelerates inference time by an order of magnitude. Comparing to those explicit representations like patches or meshes, the oriented point cloud can produce topology-agnostic and watertight manifold surfaces through differentiable Poisson Surface Reconstruction. Instead of using the Chamfer distance to estimate the oriented point cloud, we optimize the oriented point cloud with depth loss and photometric loss in a weakly supervised manner. The geometry and materials can be optimized from multi-view images efficiently.

\section{Methods}
In this section, we present our proposed coarse-to-fine approach to recover the textured mesh from multi-view images. We firstly introduce the overall pipeline of our proposed framework. Secondly, we describe the point-based shape representation and grid-based texturing scheme, which is interpretable and lightweight. Thirdly, we propose the objective function for the optimization. Finally, the implementation details are given.

\begin{figure*}[t]
	\centering
	\includegraphics[width=16cm]{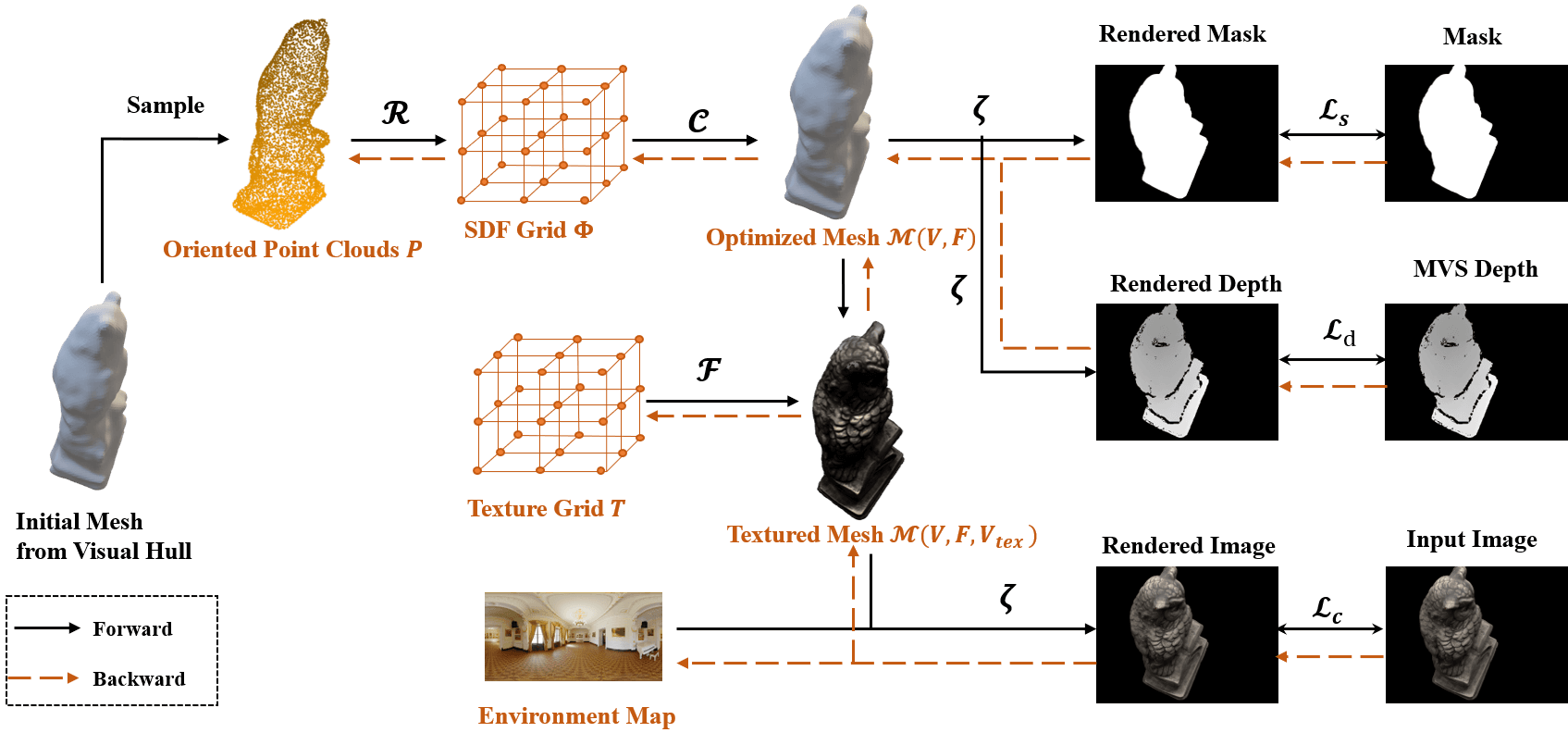}
	\caption{Overview of our proposed coarse-to-fine approach to textured mesh recovery. The coarse mesh is obtained by visual hull. Then, the oriented point clouds $P$ are sampled from the initial mesh.A differentiable Poisson Solver $\mathcal{R}$ is performed to convert $P$ to SDF grid $\Phi$. The optimized mesh $\mathcal{M} (V, F)$ is generated from $\Phi$ through differentiable marching cube $\mathcal{C}$. Differentiable renderer $\zeta$ is employed to render mask, depth and image, which is optimized by depth loss and silhouette loss. We employ physically based rendering to estimate the lighting and reflectance. The texture of the optimized mesh is trilinearly interpolated from a dense texture grid $T$. Environment map is used to represent the lighting.}
	\label{fig:overview}
\end{figure*}

\subsection{Overview}

To facilitate the effective textured mesh recovery, we firstly obtain the coarse mesh from multi-view silhouettes. Instead of deforming a sphere like IDR~\cite{DBLP:conf/nips/idr20}, we directly triangulate the visual hull~\cite{DBLP:journals/pami/visualhull94} through the marching cube algorithm. Visual hull is the maximal object that gives the same silhouette from any possible viewpoint. Each silhouette forms a cone in its corresponding camera view, and the convex hull of real object's shape is the intersection of all these visual cones. Obviously, the reconstruction results can be improved with the increasing number of views. Practically, we can get roughly correct result for convex objects. When the object is concave, the results of visual hull are far from the realistic. To tackle this critical problem, we try to recover the accurate textured mesh by taking advantage of multi-view constraints with a differentiable renderer, which is able to deform the coarse mesh to the target geometry. 

Specifically, we sample the learnable oriented point clouds from the coarse initial mesh as the shape representation, and a differentiable Poisson Surface Reconstruction~\cite{Peng2021SAP} is employed to reconstruct the mesh. Then, texture of the optimized mesh is sampled from a learnable texture grid through trilinear interpolation. Thirdly, we rasterize the predicted texture and depth map of the reconstructed mesh via differentiable rendering. Physically-based rendering is performed, and the learnable environment map is used to as the scene lighting. Finally, the textured mesh and environment map are supervised by the original images and MVS depth outputs. Fig.~\ref{fig:overview} shows the overview of our proposed coarse-to-fine framework.

The overall optimization is supervised by the following loss function, 
\begin{equation}
    \mathcal{L} = \lambda_{c}\mathcal{L}_{c} + \lambda_s \mathcal{L}_{s} + \lambda_d \mathcal{L}_{d},
\end{equation}
where $\mathcal{L}_{c}$ is the photometric loss to minimize the difference between the inverse rendered photo and input image. $\mathcal{L}_{s}$ is the silhouette loss for the mask, and $\mathcal{L}_{d}$ is the depth loss. $\lambda_{c} $, $\lambda_{s}$, and $\lambda_{d}$ are the weighting coefficients to balance the different terms.

\subsection{Shape and Texture Representation} 
Instead of using the explicit mesh with vertices and faces or implicit surface representation, we take advantage of a differentiable Poisson Solver (SAP)~\cite{Peng2021SAP} to represent the shape, which is able to produce topology-agnostic and watertight surfaces. Comparing to MLP representation, it is interpretable, lightweight and fast. The key step in Poisson Surface Reconstruction~\cite{DBLP:conf/sgp/psr06} involves solving the Poisson equation, where an oriented points set can be viewed as samples of the gradient of the underlying implicit indicator function capturing the geometry. SAP solves the Poisson equation using the spectral methods~\cite{spectral07}. As the Fast Fourier Transform (FFT) operation is well supported on GPUs, SAP can be implemented very efficiently. The mesh is generated by differentiable marching cube so that the gradients can be effectively approximated by the inverse surface normal~\cite{DBLP:conf/nips/meshsdf20}. Since all the computations are differentiable, the gradients can be backpropagated to points and normals directly.

To this end, we initialize the oriented point clouds by uniformly sampling the points and normals $P = \{ \boldsymbol{x} \in \mathbb{R}^3, \boldsymbol{n} \in \mathbb{R}^3 \} $ from the coarse mesh produced by visual hull, which is further used to build the signed distance field (SDF) through Differentiable Poisson Surface Reconstruction ($\mathcal{R}$)
\begin{equation}
    \Phi = \mathcal{R}(P),
\end{equation}
$\Phi$ is an $ r \times r \times r$ grid, which represents the SDF obtained by solving the Poisson equation from the oriented point clouds $P$. $r$ is the grid resolution. We first solve for the unnormalized SDF grid $\Phi'$ without considering the boundary conditions

\begin{equation}
    \Phi' = IFFT(\widetilde{\Phi}),
\end{equation}

\begin{equation}
    \widetilde{\Phi} = \widetilde{g}_{\sigma, r}(\mathbf{u}) \odot \frac{i \mathbf{u} \cdot \widetilde{\mathbf{v}}}{-2\pi||\mathbf{u}||^2},
\end{equation}

\begin{equation}
    \widetilde{g}_{\sigma, r}(\mathbf{u}) = exp(-2\frac{\sigma^2 ||\mathbf{}{u}||^2 }{ r^2 }),
\end{equation}
where $ IFFT $ represents the inverse fast Fourier transform. $ \widetilde{g}_{\sigma, r}(\mathbf{u}) $ is a Guassian kernel of bandwidth $\sigma$ and grid resolution $r$. $\mathbf{u} := (u,v,w)$ is the spectral frequencies corresponding to $ (x,y,z) $ spatial dimensions respectively. $\odot$ represents element-wise product. The normalized SDF grid $\Phi$ is obtained by subtracting the mean of SDF value at the point clouds $P$ and scaling by SDF value

\begin{equation}
    \Phi = \frac{m}{abs(\Phi'|\mathbf{x}=0)} (\Phi' - \frac{1}{|\{\boldsymbol{x}\}|} \sum_{\boldsymbol{x} \in \{\boldsymbol{x}\}} \Phi'|x=\boldsymbol{x}),
\end{equation}
where $ m = 0.5 $. $ \mathbf{x} \in \mathbb{R}^3 $ denotes a spatial coordinate. For more information, please refer to the original SAP paper~\cite{Peng2021SAP}.
Since the overall computation is fully differentiable, the gradient can be backpropagated to the oriented point clouds efficiently. Then, the topology-agnostic and watertight mesh can be obtained via differentiable marching cube $\mathcal{C}$
\begin{equation}
   \mathcal{M} (V, F) = \mathcal{C}(\Phi),
\end{equation}
where $V \in \mathbb{R}^{n \times 3} $ and $ F \in \mathbb{R}^{f \times 3 }$ denote the vertices and faces of the mesh $\mathcal{M}$ triangulated from SDF grid, respectively. Since $\mathcal{M}$ is generated by marching cube, the number of vertices $n$ and the number of faces $f$ are indeterminate. The forward inference of $\mathcal{C}$ is the normal Marching cube algorithm. The backward of $\mathcal{C}$ can be decomposed by the chain rule
\begin{equation}
    \frac{\partial \mathcal{L}}{\partial \Phi} = \frac{\partial \mathcal{L}}{\partial V} \frac{\partial V}{\partial \Phi}
\end{equation}
where $\mathcal{L}$ is the objective function. $\frac{\partial \mathcal{L}}{\partial V}$ can be computed by differentiable rendering, and $\frac{\partial V}{\partial \Phi}$ can be approximated by the inverse surface normal~\cite{DBLP:conf/nips/meshsdf20}
\begin{equation}
    \frac{\partial V}{\partial \Phi} = - \boldsymbol{n}.
\end{equation}

Once the mesh is reconstructed, the texture corresponding to each vertex in mesh is obtained from a learnable texture grid through trilinear interpolation $\mathcal{F}$
\begin{equation}
    V_{tex} =  \mathcal{F}(T),
\end{equation}
where $V_{tex} \in \mathbb{R}^{ n \times 7 }$ is the sampled texture. $T \in r_{tex} \times r_{tex} \times r_{tex} \times 7 $ is the learnable texture grid, and $r_tex$ represents the resolution of texture grid. Then, the textured mesh can be reconstructed under the supervision of the original multi-view images and MVS outputs via differentiable rendering.

\subsection{Silhouette Loss and Depth Loss}

To optimize the coarse mesh obtained by visual hull, we propose a differentiable rendering-based optimization framework. Given the input mesh $\mathcal{M}(V,F)$, a differentiable renderer~\cite{Laine2020diffrast} interpolates the attributes on vertices to pixels regarding to the camera parameter $\pi$. The gradients with respect to vertex positions related to occlusion, visibility, and coverage can be computed by the differentiable renderer $\zeta$. The rendered silhouette $\hat{\boldsymbol{S}}$ can be obtained by interpolating the constant value of one
\begin{equation}
    \hat{\boldsymbol{S}} = \zeta(V,F, \boldsymbol{1} ;\pi).
\end{equation}

As the differentiable renderer is able to backpropagate the gradient on the pixel back to the position of vertices, we impose the silhouette loss to limit the boundary of the generated mesh within the mask annotations,
\begin{equation}
    \mathcal{L}_{s} = \sum_{i=1}^{N} ||\boldsymbol{S}_{i} - \hat{\boldsymbol{S}_{i}}||^2_2,
\end{equation}
where $|| \cdot ||^2_2$ represents $L_2$ norm. $N$ is the number of views.

Similarly, we can render the depth map $\hat{\boldsymbol{D}}$ by a differentiable renderer $\zeta$ using the camera projection matrix and current mesh prediction, which interpolates $z$ coordinate of each vertex
\begin{equation}
    \hat{\boldsymbol{D}} = \zeta(V,F,V_z;\pi).
\end{equation}

In this work, we make use of the off-the-shelf MVS method~\cite{DBLP:conf/bmvc/mvs20} to estimate the depth map from the input images. Therefore, we can improve the mesh geometry by minimizing the difference between the rendered depth map and predictions from images as follows
\begin{equation}
    \mathcal{L}_{d} = \frac{1}{|P^{valid}|} \sum_{p \in P^{valid}} |\boldsymbol{D}_{p} - \hat{\boldsymbol{D}}_p|
\end{equation}
where $| \cdot |$ denotes $L_1$ norm. $D$ is estimated by ~\cite{DBLP:conf/bmvc/mvs20}. $P^{valid}$ is obtained from the MVS method~\cite{DBLP:conf/bmvc/mvs20}, which represents the valid indices in the estimated depth map. 

\subsection{Photometric Loss}

Instead of using MLP to represent the mesh texture like the conventional approaches~\cite{DBLP:conf/nips/idr20, mvsdf21}, we propose a physically based inverse rendering approach to jointly estimate the lighting and reflectance of the objects. 

\noindent\textbf{Physically-Based Rendering}
In computer graphics~\cite{DBLP:conf/siggraph/req86}, the color of each pixel is computed by the rendering equation based on the physical law, where we omit the radiance emitted by the object. Rendering is depicted by an integral equation determined by two factors, including the bidirectional reflectance distribution function (BRDF) representing the reflectance coefficient of object and the light emitted from the light source
\begin{equation}
    L_{o}(p,\boldsymbol{w}_o, t) = \int_{\Omega} f_{r}(p, \boldsymbol{w}_i, \boldsymbol{w}_o, t) L_{i}(p,\boldsymbol{w}_i)\boldsymbol{n} \cdot \boldsymbol{w}_i dw_{i},
\end{equation}
where $f_r(\cdot)$ is the BRDF function jointly determined by the incident light direction $\boldsymbol{w}_i$ and viewing direction $\boldsymbol{w}_o$ at the intersection point $p$. $L_{i}$ is the light intensity from direction $\boldsymbol{w}_i$. $t$ represents the texture parameter at $p$. $\boldsymbol{n}$ is the normal at point $p$. To depict an object’s spatially varying reﬂectance, we use the Cook-Torrance BRDF model~\cite{DBLP:conf/siggraph/CookT81}. We simulate the reflected lights by two parts, a certain amount of light in all directions and the other amount in a specular way
\begin{equation}
    f_{r} = f_{lambert} + f_{cook-torrance},
\end{equation}
where $f_{lambert} = a_d$ represents the diffuse component. $f_{cook-torrance}$ describes the specular component, which is usually quantified by microfacet theory. Cook-Torrance specular reflectance or microfacet BRDF has the following form
\begin{equation}
    f_{cook-torrance} = a_s \frac{DFG}{4(\boldsymbol{w}_o \cdot \boldsymbol{n})(\boldsymbol{w}_i \cdot \boldsymbol{n})},
\end{equation}
where $a_s$ denotes the specular albedo. $D$ is the normal distribution function. $F$ is the Fresnel function, and $G$ is the geometry function. $\boldsymbol{w}_i$ and $\boldsymbol{w}_o$ represent the incoming direction and outgoing direction, respectively. $D$ and $G$ are controlled by the surface roughness $\alpha$. The overall texture parameters $a_{tex} \in \mathbb{R}^7$ including $a_d \in \mathbb{R}^3$, $a_s \in \mathbb{R}^3$ and $\alpha \in \mathbb{R}^1$. We use GGX functions~\cite{DBLP:conf/rt/ggx07} to describe the normal distribution and geometry function.

The Fresnel function $F$ is used to  simulate the way light that interacts with a surface at different angles
\begin{equation}
	F = F_0 + (1-F_0)(1-\cos(\theta))^5,
\end{equation}
where $\theta$ is the angle between the viewing direction and the half vector. $F_0$ is the material response at normal incidence, which is set to $0.04$ empirically.

The distribution function $D$ is used to describe the statistical orientation of micro facets at the given point
\begin{equation}
	\chi(x) = 
	\left\{
		\begin{aligned}
			1 & \quad \mathrm{if} \, \boldsymbol{x} > 0 \\
			0 & \quad \mathrm{if} \, \boldsymbol{x} \leq 0
		\end{aligned}
	\right.
\end{equation}
\begin{equation}
	D(\boldsymbol{h}, \boldsymbol{n}, \alpha) = \frac{\alpha^2 \chi(\boldsymbol{h} \cdot \boldsymbol{n})}{\pi ((\boldsymbol{h} \cdot \boldsymbol{n})^2 (\alpha^2 + (\frac{1-(\boldsymbol{h} \cdot \boldsymbol{n})^2}{(\boldsymbol{h} \cdot \boldsymbol{n})^2})))^2}
\end{equation} 	
where $\alpha$ is the roughness of surface. $h$ is the half vector.

The geometry function $G$ is used to describe the attenuation of the light due to the microfacets shadowing each other
\begin{equation}
	G(\boldsymbol{w}_i, \boldsymbol{w}_o, \boldsymbol{h}, \boldsymbol{n}, \alpha) = G_p(\boldsymbol{w}_i, \boldsymbol{h}, \boldsymbol{n}, \alpha) G_p(\boldsymbol{w}_o, \boldsymbol{h}, \boldsymbol{n}, \alpha)
\end{equation}
\begin{equation}
	G_p(\boldsymbol{w}, \boldsymbol{h}, \boldsymbol{n}, \alpha) = \chi\left (\frac{\boldsymbol{w} \cdot \boldsymbol{h}}{\boldsymbol{w} \cdot \boldsymbol{n}}\right ) \frac{2}{1 + \sqrt{1 + \alpha^2 \frac{1 - (\boldsymbol{w} \cdot \boldsymbol{n})^2}{(\boldsymbol{h} \cdot \boldsymbol{n})^2}}}.
\end{equation}

As for the light source of the scene, we adopt an HDR light probe image~\cite{DBLP:conf/siggraph/hdrprobe98} in the latitude-longitude format, which is simple and direct. Specifically, we use a $4 \times 8$ resolution for our lighting environments. In our implementation, we create an environment map for each image. 

Let $V_{tex}$ represent the texture parameter for each vertex, which is sampled from the learnable texture grid by trilinear interpolation. Therefore, the interpolated texture map $\hat{T}$ can be obtained by
\begin{equation}
    \hat{T} = \zeta(V,F,V_{tex};\pi).
\end{equation}

Let $(u,v) = \pi(p)$ denote the 2D projection of 3D intersection point $p$. The pixel color of the rendered image $\hat{I}$ is computed by the rendering equation
\begin{equation}
    \hat{I}_{(u,v)} = L_o(p, \boldsymbol{w}_o, \hat{T}_{(u,v)}).
\end{equation}

From the above all, the inverse rendering optimization can be directly self-supervised by the input image through minimizing the photometric loss $ \mathcal{L}_{c} $ as below
\begin{equation}
    \mathcal{L}_{c} = \frac{1}{|P^{in}|} \sum_{p \in P^{in}}|I_{p} - \hat{I}_{p}|.
\end{equation}
 During the joint optimization, the probe pixels in the lighting environments are updated by the photometric loss.

\begin{table*}
	\centering
	\caption{Quantitative results on DTU dataset. Our method achieves the smallest mean Chamfer distance and comparable PSNR results.}
	\resizebox{\linewidth}{!}{%
		\begin{tabular}{l||cc|ccc||cc|ccc}
			\toprule
			& \multicolumn{5}{c||}{Chamfer (mm)}         & \multicolumn{5}{c}{PSNR}                    \\ \cline{2-11}
			& PhySG~\cite{DBLP:conf/cvpr/physg21} & Vis-MVSNet~\cite{DBLP:conf/bmvc/mvs20} & IDR~\cite{DBLP:conf/nips/idr20} & MVSDF~\cite{mvsdf21} & Ours & PhySG~\cite{DBLP:conf/cvpr/physg21} & Vis-MVSNet~\cite{DBLP:conf/bmvc/mvs20} & IDR~\cite{DBLP:conf/nips/idr20} & MVSDF~\cite{mvsdf21} & Ours  \\ \hline
			24   & 4.31   & 0.98       & 1.63 & 0.83 & \textbf{0.54} & 21.87  & 18.35      & 23.29 & \textbf{25.02} & 24.37 \\
			37   & 3.39   & 2.10       & 1.87 & 1.76 & \textbf{0.88} & 19.18  & 14.71      & \textbf{21.36} & 19.47 & 18.88 \\
			40   & 4.08   & 0.93       & 0.63 & 0.88 & \textbf{0.35} & 24.15  & 18.60      & 24.39 & \textbf{25.96} & 25.50 \\
			55   & 1.50   & 0.46       & 0.48 & 0.44 & \textbf{0.37} & 22.96  & 19.07      & 22.96 & 24.14 & \textbf{25.13} \\
			63   & 4.18   & 1.89       & 1.04 & 1.11 & \textbf{0.93} & 23.09  & 17.55      & 23.22 & 22.16 & \textbf{25.13} \\
			65   & 2.31   & \textbf{0.67}       & 0.79 & 0.90 & 0.89 & 21.01  & 17.17      & 23.94 & \textbf{26.89} & 25.49 \\
			69   & 2.87   & 0.67       & 0.77 & 0.75 & \textbf{0.57} & 18.36  & 21.81      & 20.34 & \textbf{26.38} & 24.80 \\
			83   & 2.84   & \textbf{1.08}       & 1.33 & 1.26 & 1.41 & 23.89  & 23.11      & 21.87 & 25.79 & \textbf{29.76} \\
			97   & 4.23   & \textbf{0.67}       & 1.16 & 1.02 & 0.83 & 22.12  & 18.68      & 22.95 & \textbf{26.22} & 22.31 \\
			105  & 2.31   & 0.95       & \textbf{0.76} & 1.35 & 0.74 & 22.51  & 21.68      & 22.71 & 27.29 & \textbf{27.55} \\
			106  & 3.20   & 0.66  & 0.67 & 0.87 & \textbf{0.44} & 21.13  & 21.03      & 22.81 & 27.78 & \textbf{30.01} \\
			110  & 1.79   & 0.85       & 0.90 & \textbf{0.84} & 0.98 & 21.43  & 18.41      & 21.26 & \textbf{23.82} & 21.26 \\
			114  & 1.47   & \textbf{0.30} & 0.42 & 0.34 & 0.33 & 23.74  & 19.42      & 25.35 & \textbf{27.79} & 26.08 \\
			118  & 5.03   & 0.45 & 0.51 & 0.47 & \textbf{0.42} & 22.30  & 23.85      & 23.54 & 28.60 & \textbf{31.93} \\
			122  & 1.87   & 0.51       & 0.53 & \textbf{0.46} & 0.47 & 26.07  & 24.29      & 27.98 & 31.49 & \textbf{32.44} \\ \hline
			Mean & 3.02   & 0.88       & 0.90 & 0.88 & \textbf{0.68} & 22.25  & 19.85      & 23.20 & 25.92 & \textbf{26.04} \\
			\bottomrule
	\end{tabular}}
	\label{tab:dtu}
\end{table*}

\subsection{Implementation Details}

In this paper, we firstly compute the visual hull at the resolution of $128^3$. Secondly, we uniformly sample 10,000 oriented points from the initial mesh triangulated from the visual hull using marching cube, which are used to minimize the depth and silhouette differences through the differentiable rendering-based optimization. It is performed at a resolution of $128^3$ for 150 epochs. To obtain the fine details, we increase the resolution to $256^3$ and the number of sampled oriented points to 60,000 for the extra 150 epochs. Moreover, we resample points and normals every 50 epochs in order to increase the robustness of the optimization process, and replace the original point cloud with the resampled ones. Thirdly, the texture of the generated mesh is interpolated from a learnable texture grid with an initial resolution of $128^3$, which is interpolated into $256^3$ after 150 epochs. Finally, the texture grid and environment map are supervised by the photometric loss. In our implementation, we use Adam optimizer in minimization. The learning rates for the oriented point clouds and texture grid are $5e^{-4}$ and $1e^{-4}$, respectively. The weights for the loss term are $\lambda_{c}=5$, $\lambda_{s}=10$, and $\lambda_{d} = 30$. The optimization and inverse rendering are implemented by PyTorch. We use the differentiable renderer nvdiffrast~\cite{Laine2020diffrast} to obtain the silhouette and texture map. Then, we perform Cook-Torrance BRDF rendering to get the rendered image.

\section{Experiments}

In this section, we discuss the results on the textured mesh recovery from multiple views. We compare our method against the state-of-the-art methods on DTU and EPFL datasets. Moreover, we also show the additional qualitative results on BlendedMVS and Tanks and Temples datasets. Additionally, we demonstrate that our reconstructed meshes can be combined and rendered with arbitrary lighting. Finally, we conduct ablation studies on DTU dataset to investigate the effect of depth loss and mask supervision.

\begin{figure*}
	\centering
	\begin{tabular}{@{\hskip2pt}c@{\hskip2pt}@{\hskip2pt}c@{\hskip2pt}@{\hskip2pt}c@{\hskip2pt}@{\hskip2pt}c@{\hskip2pt}@{\hskip2pt}c@{\hskip2pt}}
		\begin{tikzpicture}\node[above right, inner sep=0](image) at (0,0) {\includegraphics[width=0.19\linewidth]{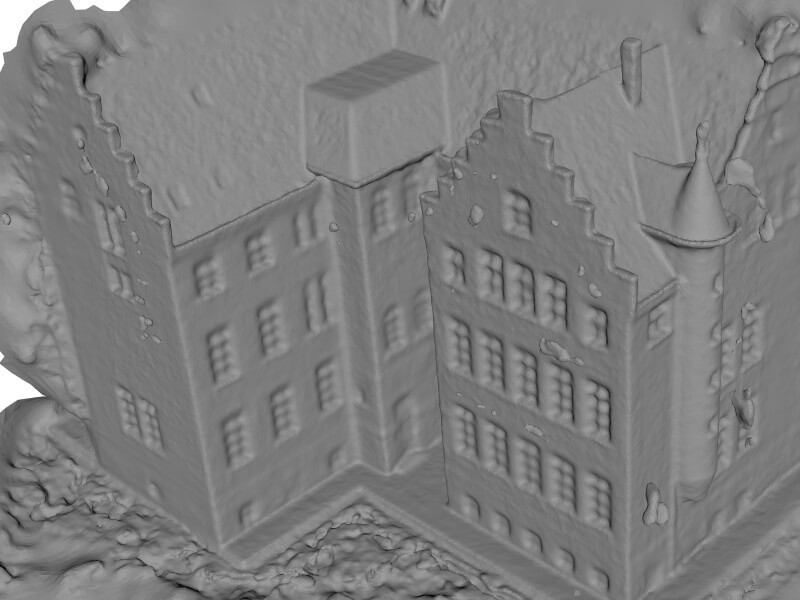}};
		\end{tikzpicture} &
		\begin{tikzpicture}\node[above right, inner sep=0](image) at (0,0) {\includegraphics[width=0.19\linewidth]{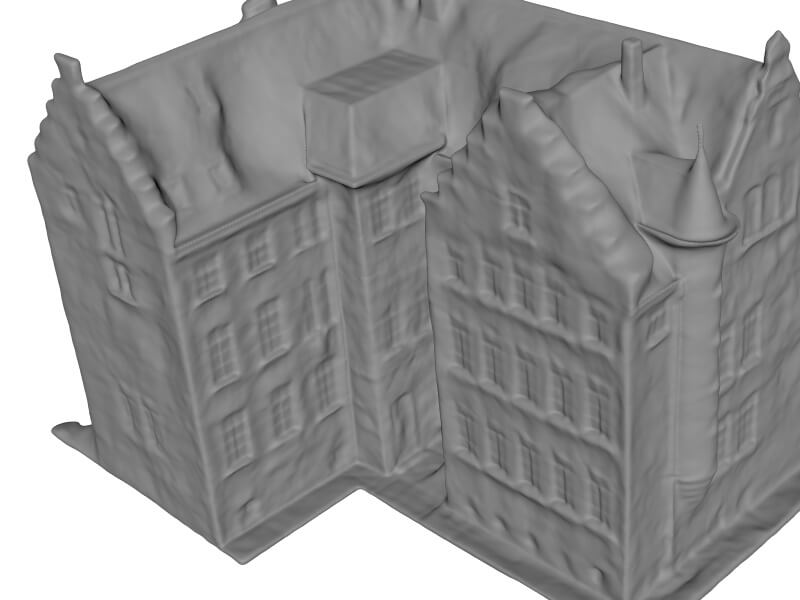}};
		\end{tikzpicture} &
		\begin{tikzpicture}\node[above right, inner sep=0](image) at (0,0) {\includegraphics[width=0.19\linewidth]{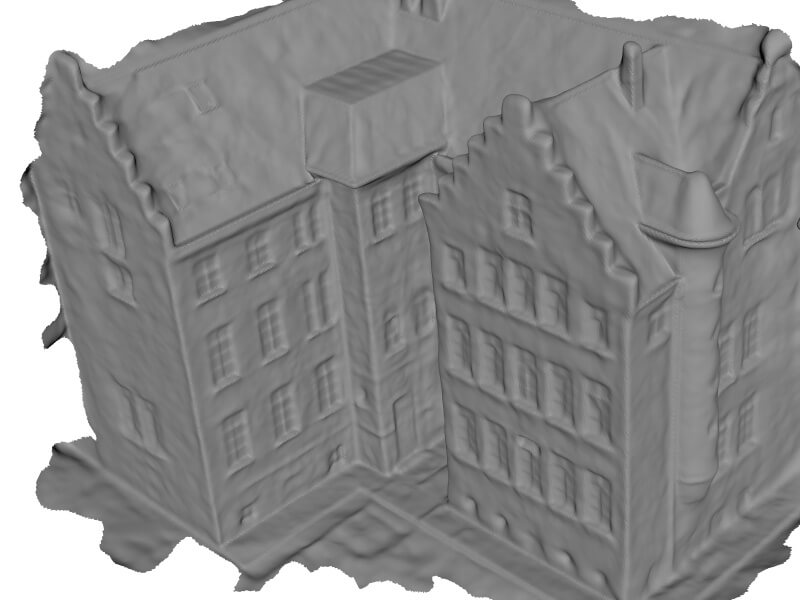}};
		\end{tikzpicture} &
		\begin{tikzpicture}\node[above right, inner sep=0](image) at (0,0) {\includegraphics[width=0.19\linewidth]{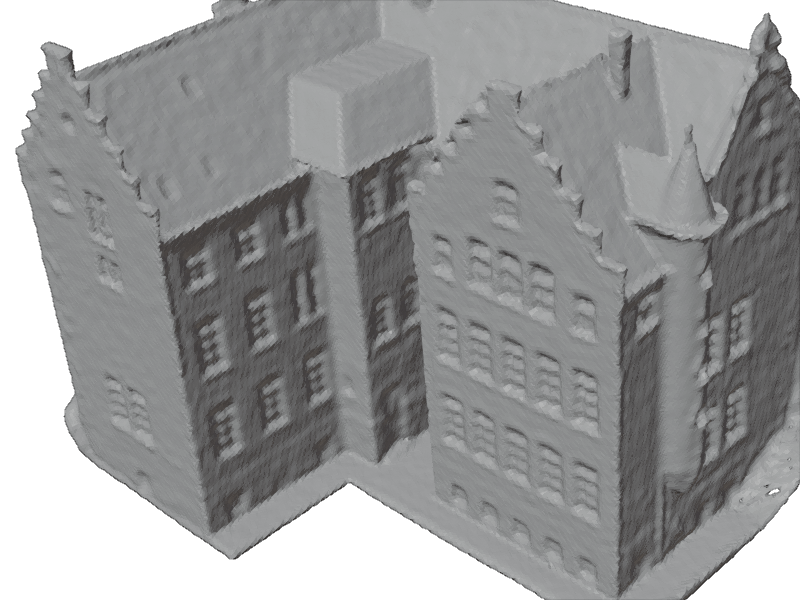}};
		\end{tikzpicture} & 
		\begin{tikzpicture}\node[above right, inner sep=0](image) at (0,0) {\includegraphics[width=0.19\linewidth]{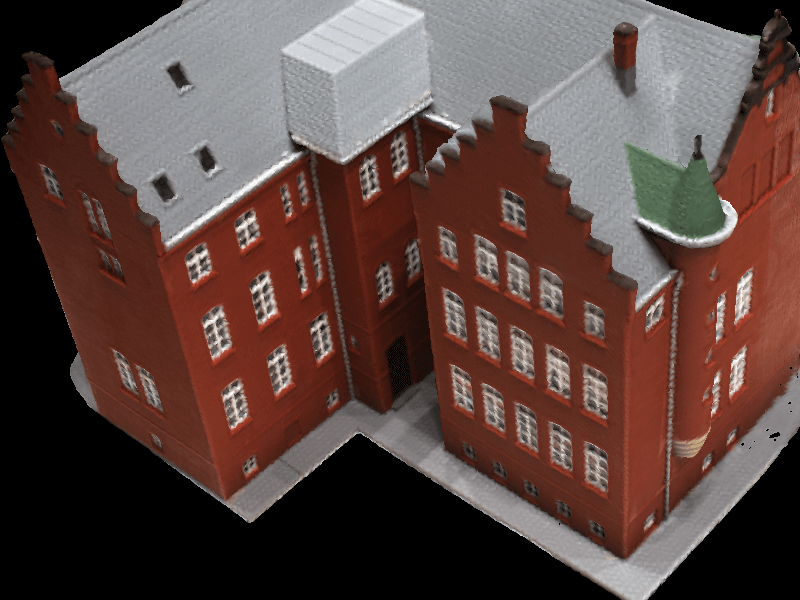}}; 
		\end{tikzpicture} \\
		
		\begin{tikzpicture}\node[above right, inner sep=0](image) at (0,0) {\includegraphics[width=0.19\linewidth]{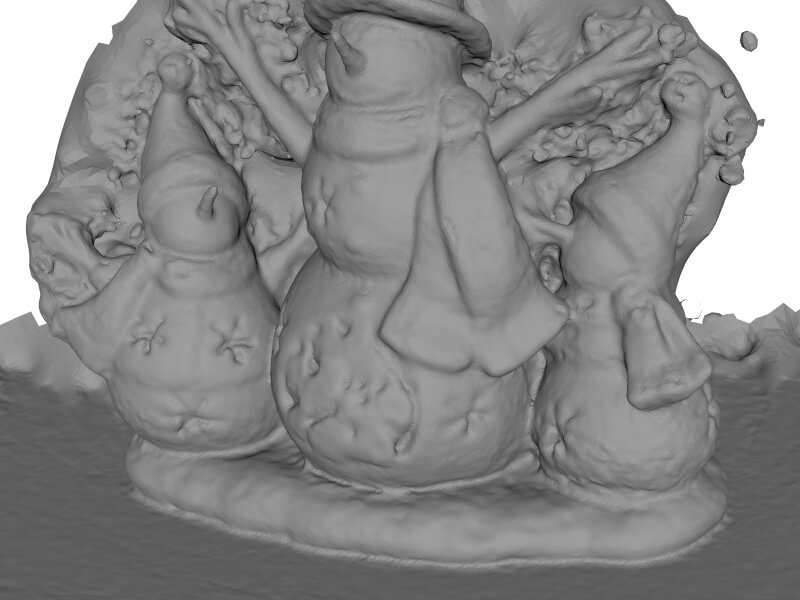}};
		\end{tikzpicture} &
		\begin{tikzpicture}\node[above right, inner sep=0](image) at (0,0) {\includegraphics[width=0.19\linewidth]{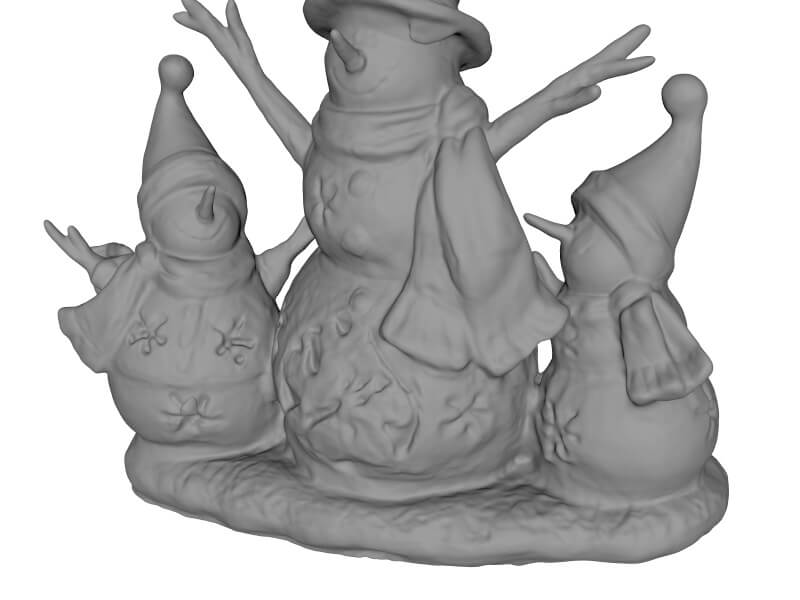}};
		\end{tikzpicture} &
		\begin{tikzpicture}\node[above right, inner sep=0](image) at (0,0) {\includegraphics[width=0.19\linewidth]{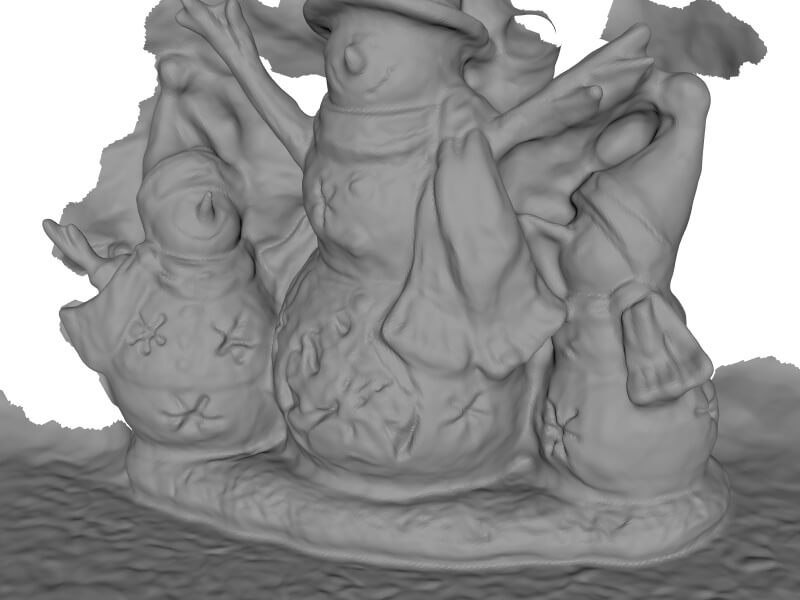}}; \end{tikzpicture} &
		\begin{tikzpicture}\node[above right, inner sep=0](image) at (0,0) {\includegraphics[width=0.19\linewidth]{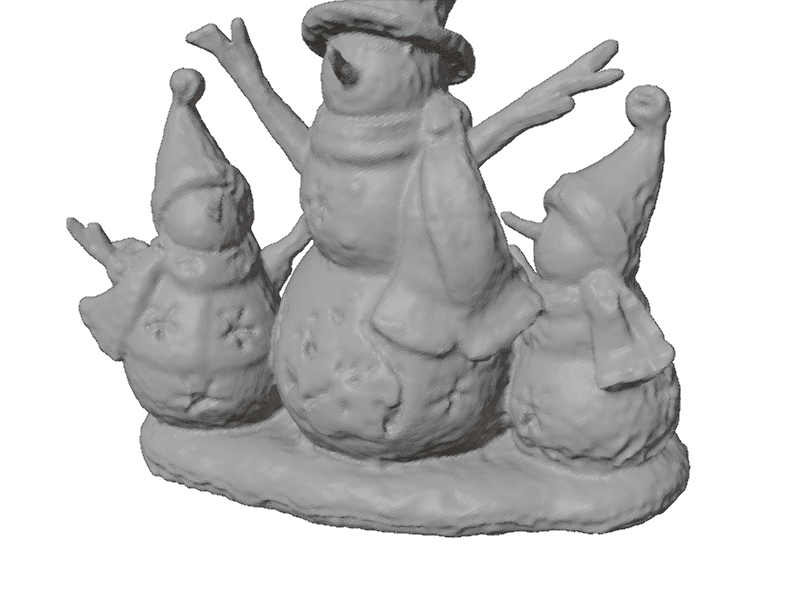}};
		\end{tikzpicture} & 
		\begin{tikzpicture}\node[above right, inner sep=0](image) at (0,0) {\includegraphics[width=0.19\linewidth]{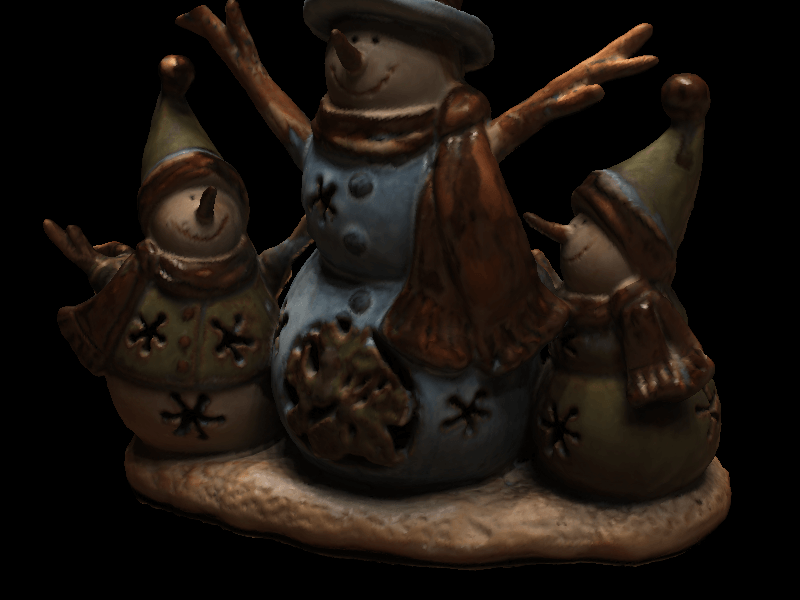}}; \end{tikzpicture} \\
		
		\begin{tikzpicture}\node[above right, inner sep=0](image) at (0,0) {\includegraphics[width=0.19\linewidth]{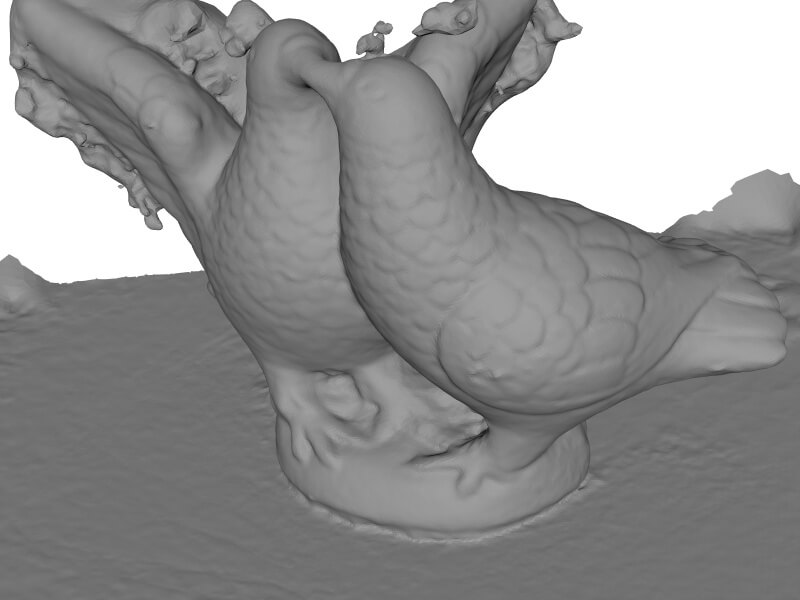}};
		\end{tikzpicture} &
		\begin{tikzpicture}\node[above right, inner sep=0](image) at (0,0) {\includegraphics[width=0.19\linewidth]{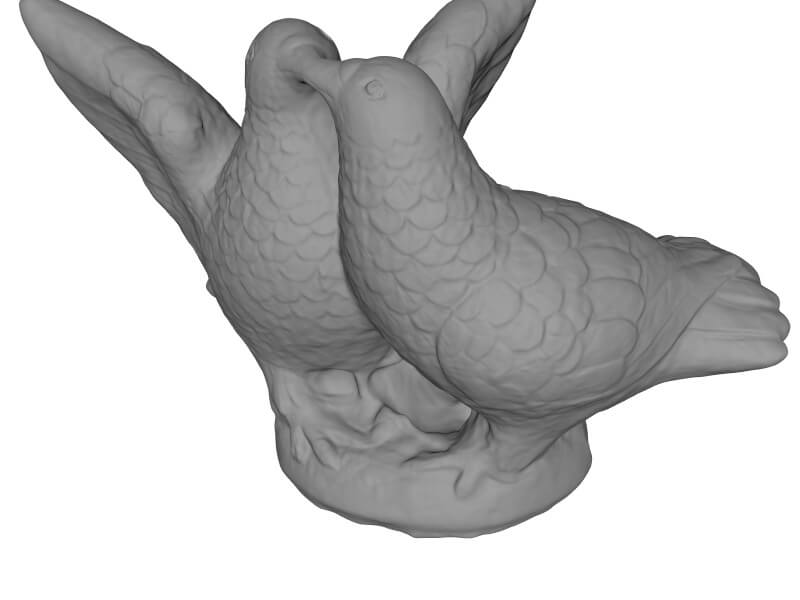}};
		\end{tikzpicture} &
		\begin{tikzpicture}\node[above right, inner sep=0](image) at (0,0) {\includegraphics[width=0.19\linewidth]{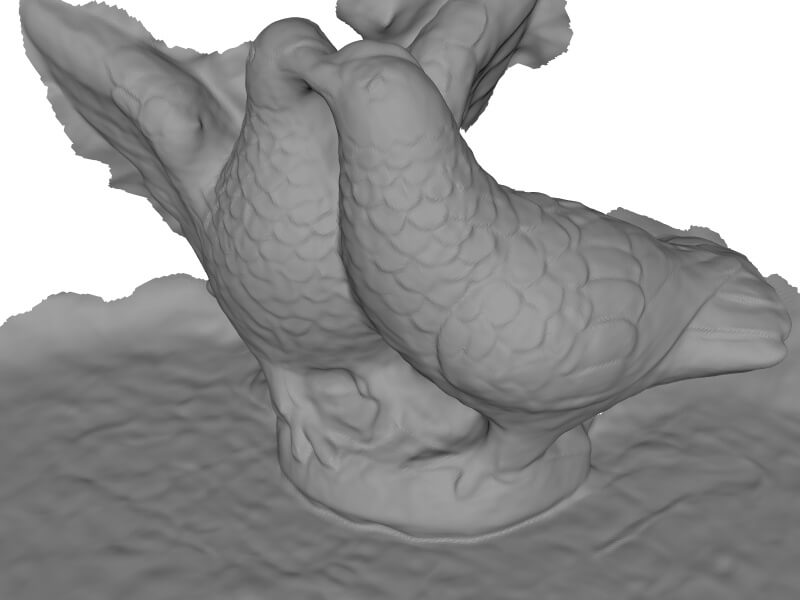}};
		\end{tikzpicture} &
		\begin{tikzpicture}\node[above right, inner sep=0](image) at (0,0) {\includegraphics[width=0.19\linewidth]{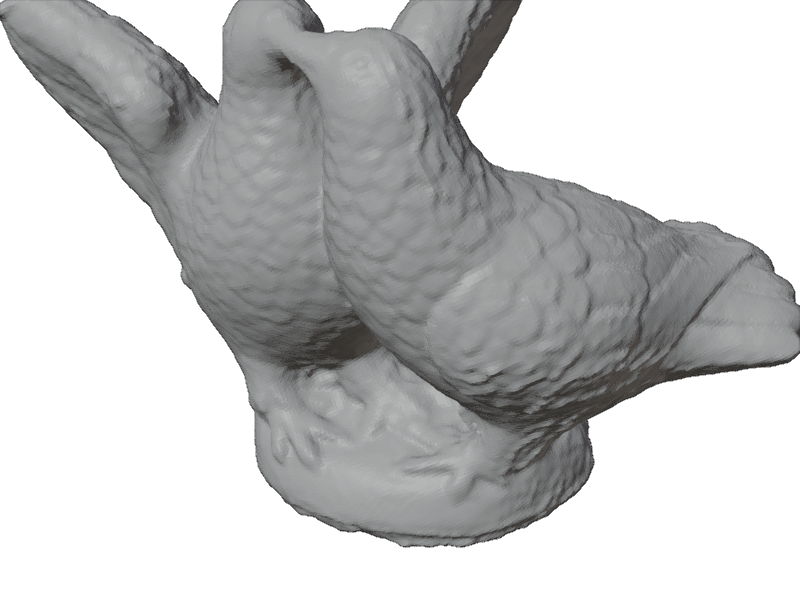}};
		\end{tikzpicture} & 
		\begin{tikzpicture}\node[above right, inner sep=0](image) at (0,0) {\includegraphics[width=0.19\linewidth]{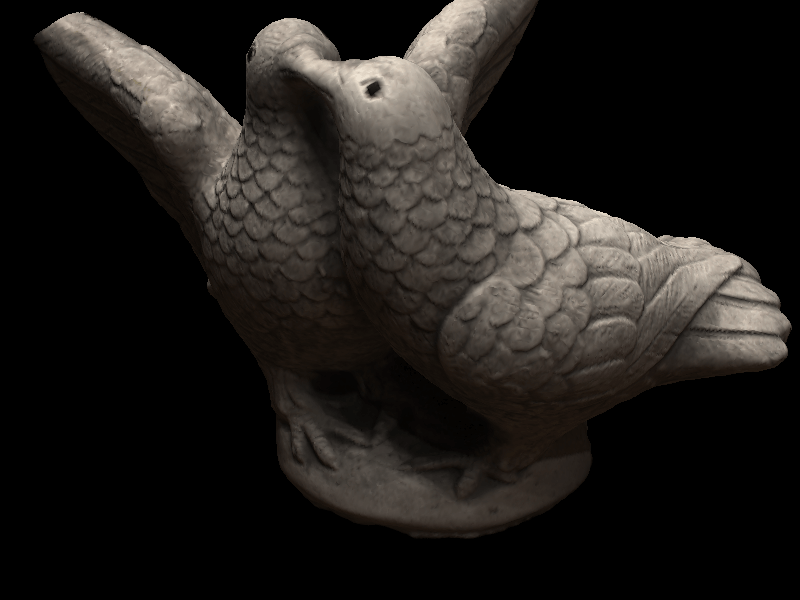}}; 
		\end{tikzpicture} \\
		
		\begin{tikzpicture}\node[above right, inner sep=0](image) at (0,0) {\includegraphics[width=0.19\linewidth]{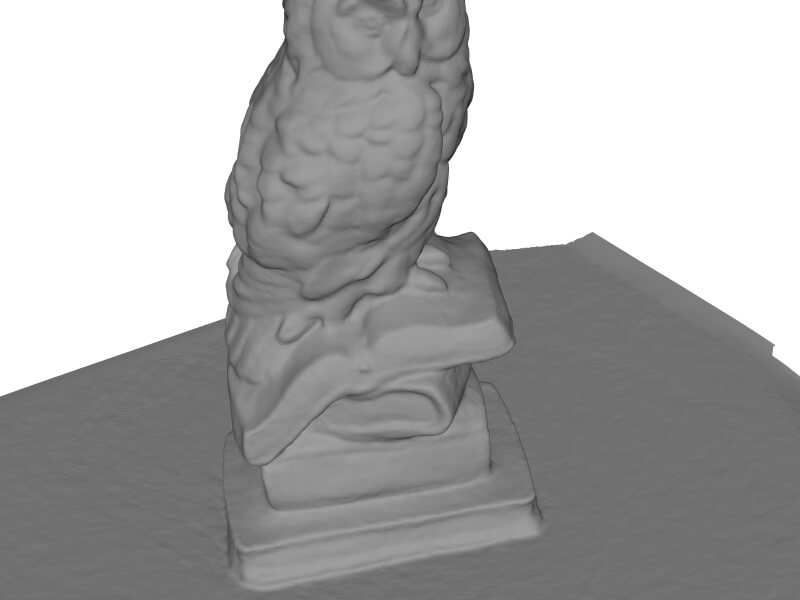}};
		\end{tikzpicture} &
		\begin{tikzpicture}\node[above right, inner sep=0](image) at (0,0) {\includegraphics[width=0.19\linewidth]{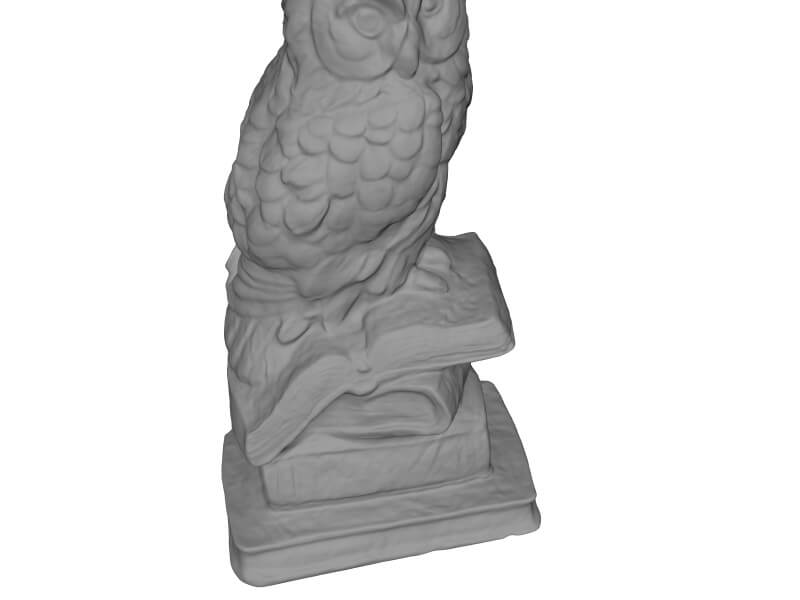}};
		\end{tikzpicture} &
		\begin{tikzpicture}\node[above right, inner sep=0](image) at (0,0) {\includegraphics[width=0.19\linewidth]{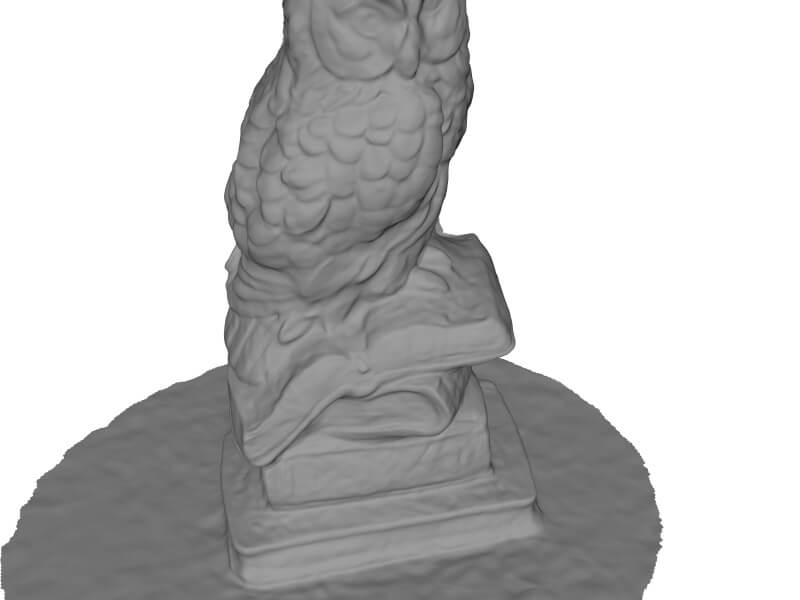}};
		\end{tikzpicture} &
		\begin{tikzpicture}\node[above right, inner sep=0](image) at (0,0) {\includegraphics[width=0.19\linewidth]{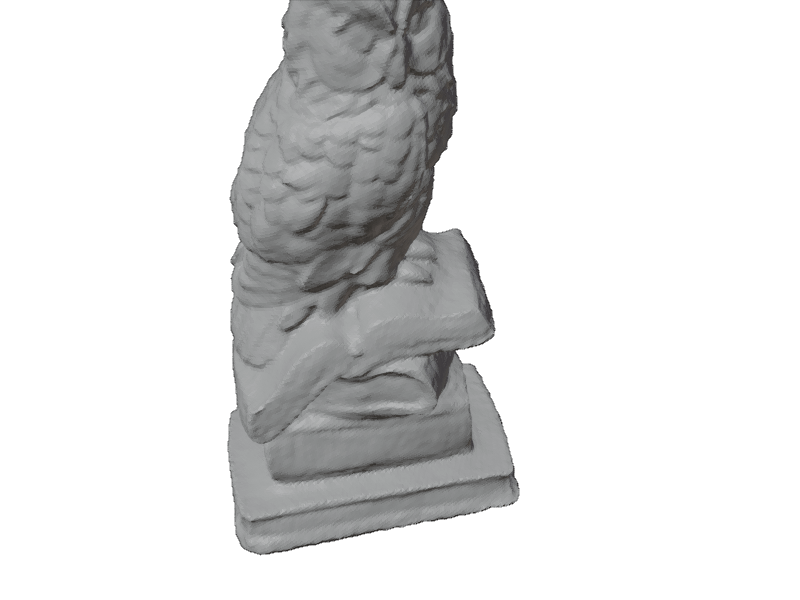}};
		\end{tikzpicture} & 
		\begin{tikzpicture}\node[above right, inner sep=0](image) at (0,0) {\includegraphics[width=0.19\linewidth]{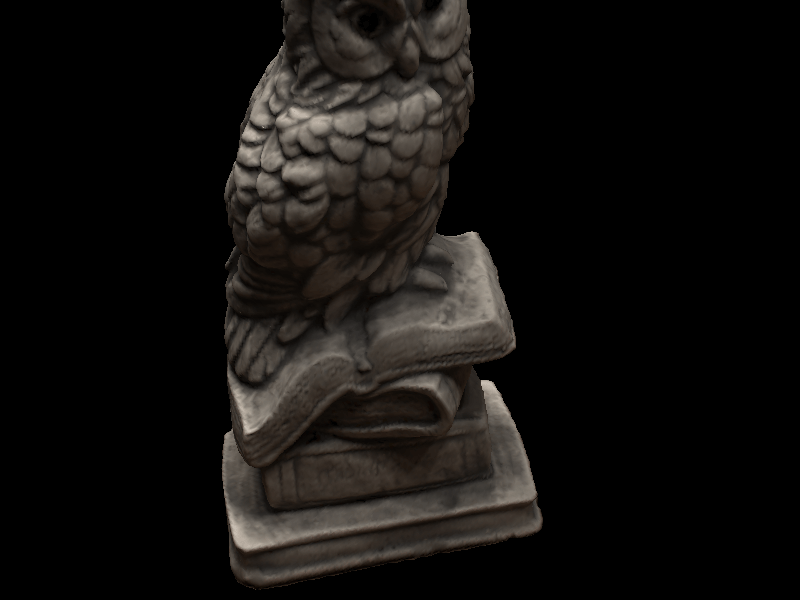}}; 
		\end{tikzpicture} \\
		
		Vis-MVSNet~\cite{DBLP:conf/bmvc/mvs20} & IDR~\cite{DBLP:conf/nips/idr20} & MVSDF~\cite{mvsdf21} & Ours & Ours Render
	\end{tabular}
	\caption{Qualitative results on DTU dataset. We compare our method with Vis-MVSNet~\cite{DBLP:conf/bmvc/mvs20}, IDR~\cite{DBLP:conf/nips/idr20} and MVSDF~\cite{mvsdf21}.}
	\label{fig:dtu}
\end{figure*}

\subsection{Evaluation on DTU Dataset}

We firstly evaluate our proposed approach on the DTU MVS dataset~\cite{DBLP:conf/cvpr/DTU14}, which contains 128 scans. For each scan, there are 49 calibrated cameras surrounding the captured object. We evaluate both the generated mesh and rendered image using the same set of scans as the conventional method~\cite{DBLP:conf/nips/idr20,mvsdf21}.

To facilitate fair comparison, Chamfer distance is employed as the evaluation metric to measure the accuracy of generated mesh, and PSNR is used to evaluate the rendered images. We compare our method against the recent state-of-the-art 3D reconstruction methods, including PhySG~\cite{DBLP:conf/cvpr/physg21}, Vis-MVSNet~\cite{DBLP:conf/bmvc/mvs20}, IDR~\cite{DBLP:conf/nips/idr20} and MVSDF~\cite{mvsdf21}. Depth maps from Vis-MVSNet are fused into point clouds, which are further converted into meshes by the screened Poisson Surface Reconstruction (sPSR)~\cite{DBLP:journals/tog/spsr13} with trim parameter 5. The color is assigned from input images by back projection. The reconstruction results of Vis-MVSNet and MVSDF may include some extrapolated surfaces as they do not use the pre-defined mask. Therefore, a graph-cut based algorithm is performed to trim the extra faces for fair comparison.

Table~\ref{tab:dtu} shows the quantitative results. It can be clearly observed that our proposed approach achieves the lowest mean Chamfer distance on DTU dataset. This indicates that we achieve the most accurate reconstruction results. Due to the lack of GPU RAM, we use an environment map of $4 \times 8$, which cannot simulate the complex lighting with high fidelity. Our rendered results on the image quality are a bit inferior in some objects with high reflectance like scan 37 and 97. While Cook-Torrance BRDF model can simulate rough surfaces well, this leads to good PSNR results, such as scan 83 and 105. Moreover, our proposed method achieves the highest mean PSNR on DTU dataset.

\begin{table}[htbp]
	\centering
	\caption{Image synthesis evaluation on DTU dataset.}
	\resizebox{\linewidth}{!}{%
		\begin{tabular}{l||ccc||ccc}
			\toprule
			& \multicolumn{3}{c||}{SSIM}  & \multicolumn{3}{c}{LPIPS}  \\ 
			\cline{2-7}
			& IDR~\cite{DBLP:conf/nips/idr20} & MVSDF~\cite{mvsdf21} & Ours & IDR~\cite{DBLP:conf/nips/idr20} & MVSDF~\cite{mvsdf21} & Ours  \\ \hline
			24   & 0.74   & 0.75 & \textbf{0.81} & 0.32 & 0.31 & \textbf{0.21} \\
			37   & 0.81   & 0.80 & \textbf{0.83} & 0.22 & 0.23 & \textbf{0.19} \\
			40   & 0.73   & 0.72 & \textbf{0.78} & 0.40 & 0.41 & \textbf{0.31} \\
			55   & 0.86   & 0.84 & \textbf{0.91} & 0.20 & 0.23 & \textbf{0.11} \\
			63   & 0.94   & 0.94 & \textbf{0.95} & 0.13 & 0.13 & \textbf{0.09} \\
			65   & \textbf{0.95}   & \textbf{0.95} & 0.94 & 0.13 & \textbf{0.12} & \textbf{0.12} \\
			69   & 0.91   & \textbf{0.92} & \textbf{0.92} & 0.25 & 0.21 & \textbf{0.17} \\
			83   & 0.96   & 0.96 & \textbf{0.97} & 0.10 & 0.09 & \textbf{0.06} \\
			97   & \textbf{0.92}   & 0.91 & 0.90 & 0.15 & 0.15 & \textbf{0.13} \\
			105  & 0.81   & 0.90 & \textbf{0.92} & 0.26 & 0.19 & \textbf{0.14} \\
			106  & 0.92   & 0.90  & \textbf{0.93} & 0.21 & 0.22 & \textbf{0.13} \\
			110  & \textbf{0.93}   & 0.92 & 0.91 & 0.20 & 0.18 & \textbf{0.15} \\
			114  & 0.90 & 0.90 & \textbf{0.91} & 0.23 & 0.22 & \textbf{0.17} \\
			118  & \textbf{0.95} & 0.93 & \textbf{0.95} & 0.17 & 0.17 & \textbf{0.11} \\
			122  & \textbf{0.96} & 0.95 & \textbf{0.96} & 0.12 & 0.12 & \textbf{0.09} \\ \hline
			Mean & 0.89  & 0.89 & \textbf{0.91} & 0.21 & 0.19 & \textbf{0.15} \\
			\bottomrule
	\end{tabular}}
	\label{tab:ssim}
\end{table}

In order to better measure the quality of synthesized images, we also employ SSIM and LPIPS as evaluation metrics to further evaluate the image synthesis results. We compare our proposed approach with IDR and MVSDF, as they perform well on PSNR. Table~\ref{tab:ssim} shows the experimental results. It can be seen that our proposed method outperforms other methods under SSIM and LPIPS evaluation metrics. These results show that our proposed can generate the photo-realistic images. Fig.~\ref{fig:dtu} gives the example reconstruction results. It can be seen that our method obtains more accurate meshes and synthesizes the photo-realistic images.

\begin{figure*}[htbp]
	\centering
	\begin{tabular}{@{\hskip2pt}c@{\hskip2pt}@{\hskip2pt}c@{\hskip2pt}@{\hskip2pt}c@{\hskip2pt}@{\hskip2pt}c@{\hskip2pt}@{\hskip2pt}c@{\hskip2pt}}
		\begin{tikzpicture}\node[above right, inner sep=0](image) at (0,0) {\includegraphics[width=0.23\linewidth]{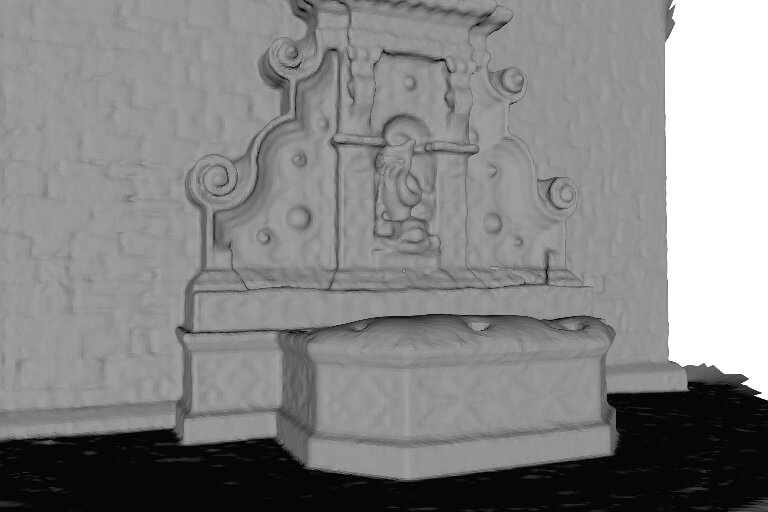}}; \end{tikzpicture} &
		\begin{tikzpicture}\node[above right, inner sep=0](image) at (0,0) {\includegraphics[width=0.23\linewidth]{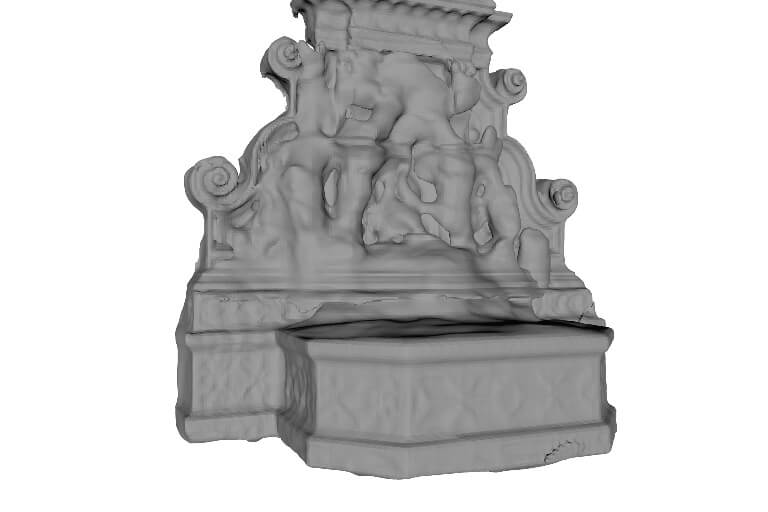}}; \end{tikzpicture} &
		\includegraphics[width=0.23\linewidth]{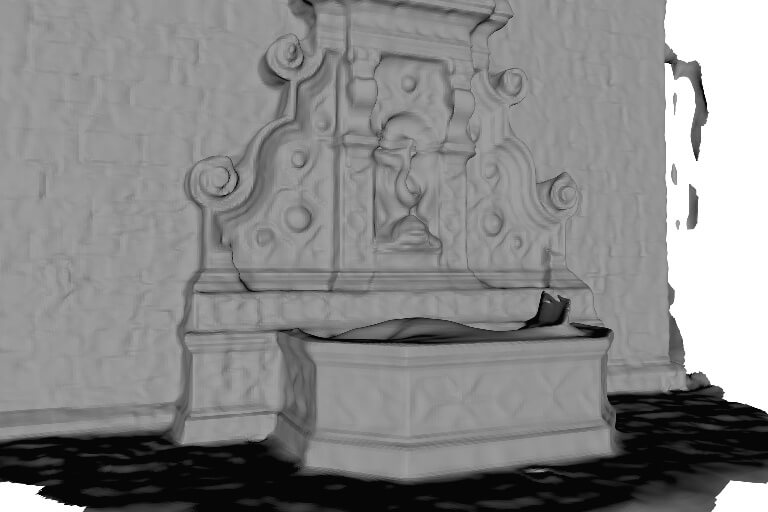} &
		
		\begin{tikzpicture}\node[above right, inner sep=0](image) at (0,0) {\includegraphics[width=0.23\linewidth]{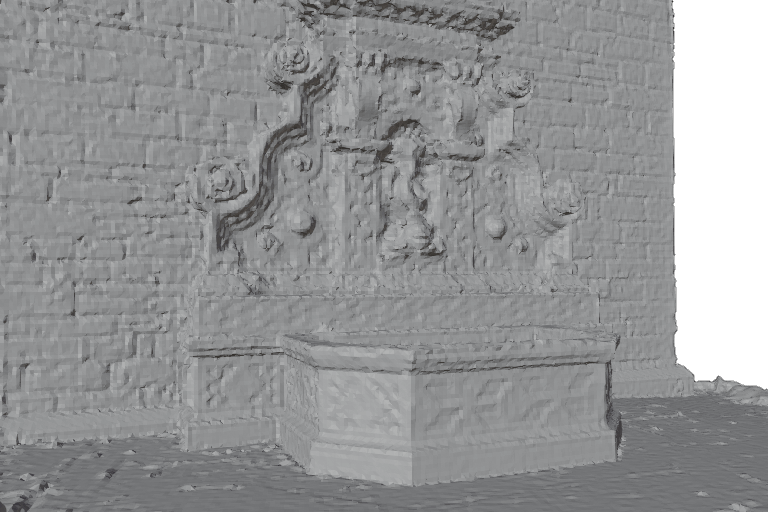}};
		\end{tikzpicture} \\
		
		\includegraphics[width=0.23\linewidth]{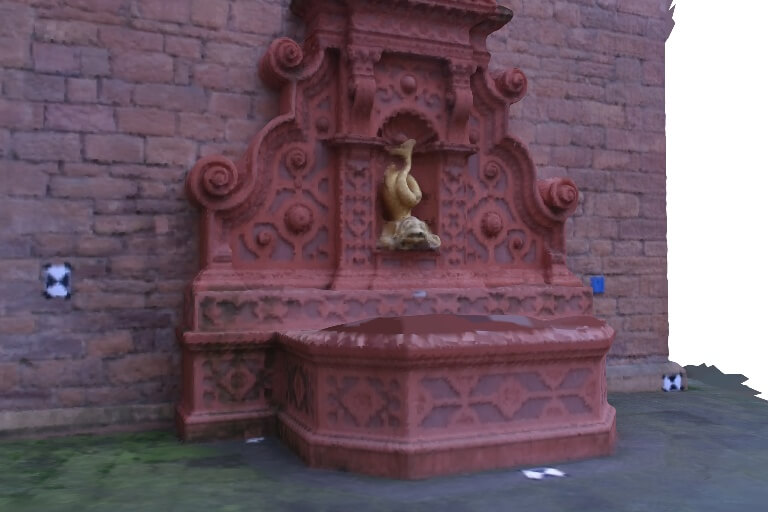} &
		\includegraphics[width=0.23\linewidth]{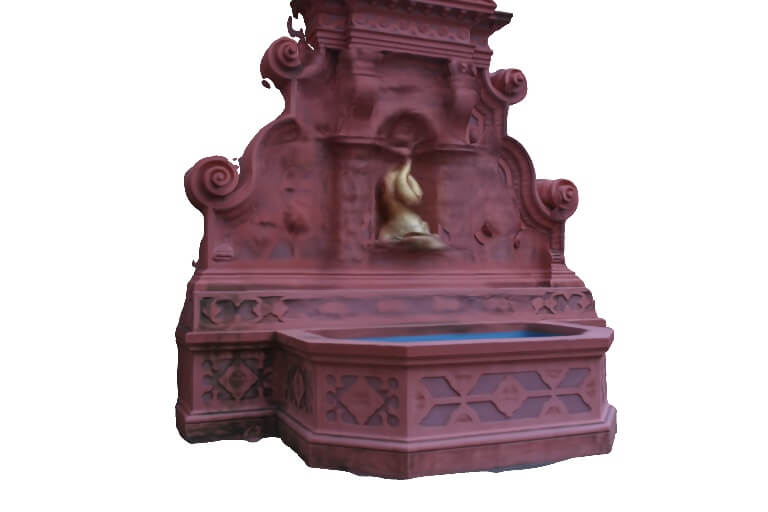} &
		\includegraphics[width=0.23\linewidth]{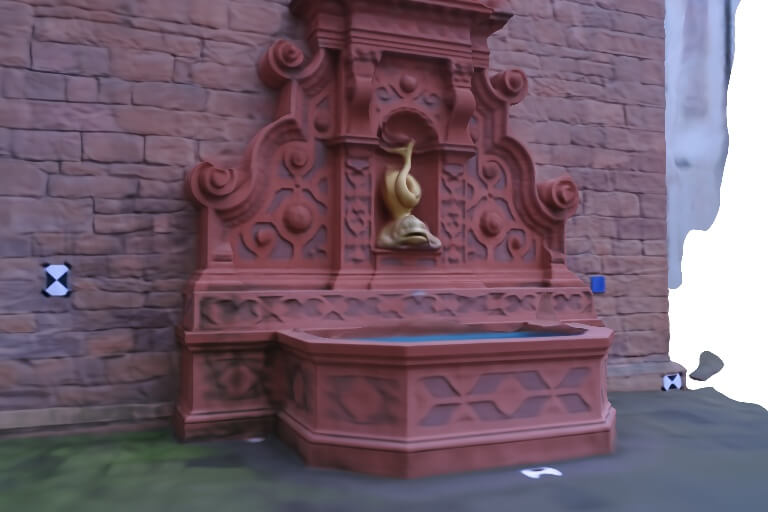} &
		\includegraphics[width=0.23\linewidth]{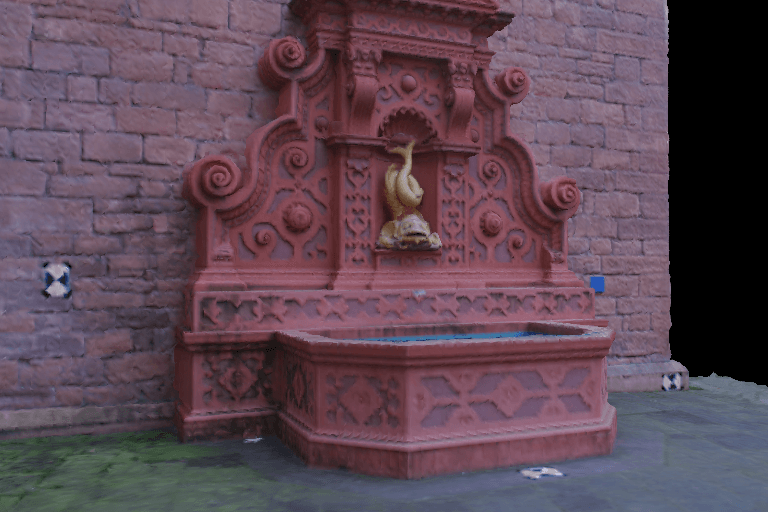} \\
		
		\includegraphics[width=0.23\linewidth]{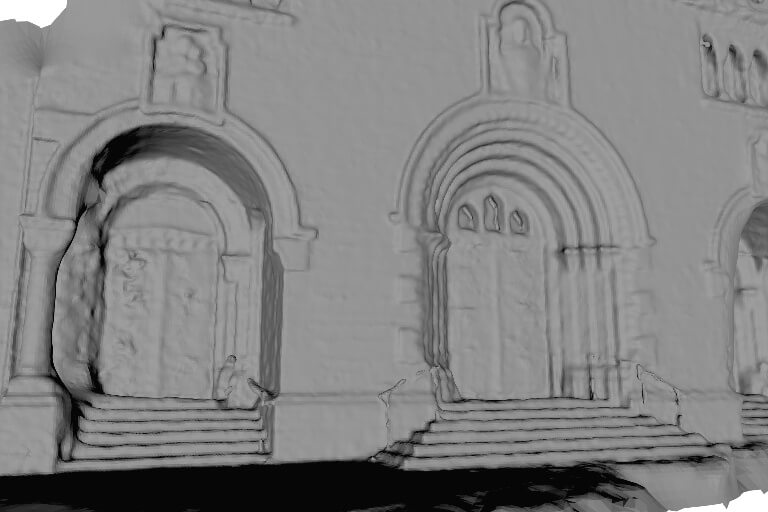} &
		\includegraphics[width=0.23\linewidth]{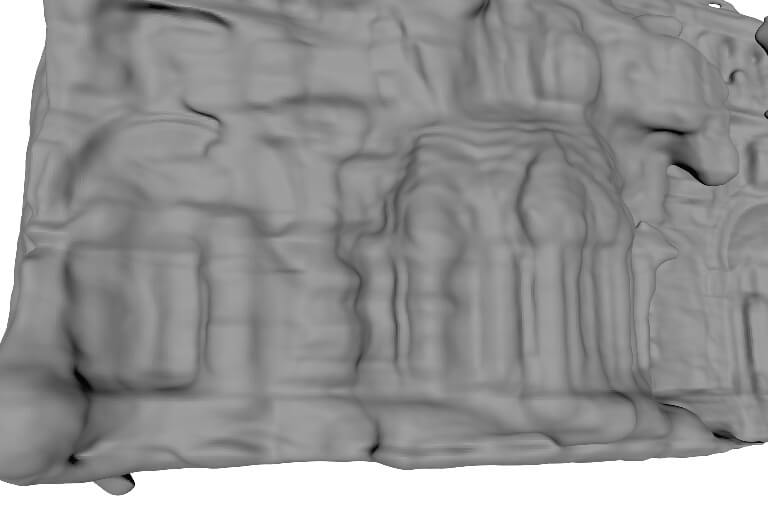} &
		\includegraphics[width=0.23\linewidth]{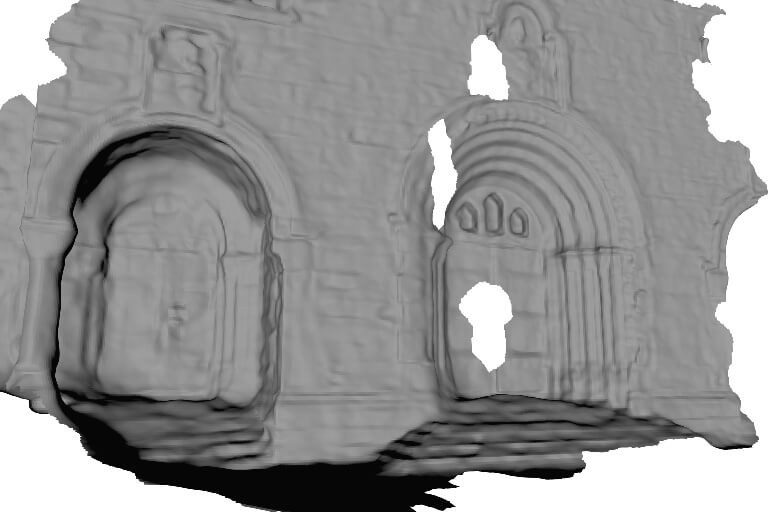} &
		\includegraphics[width=0.23\linewidth]{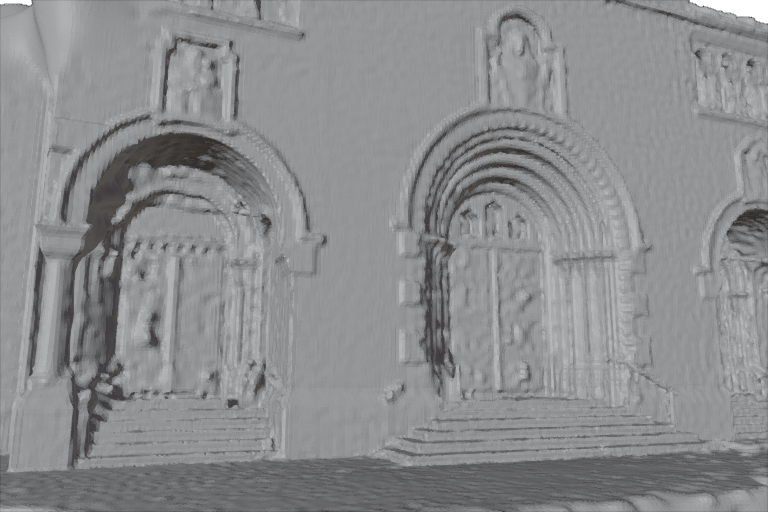} \\
		
		\includegraphics[width=0.23\linewidth]{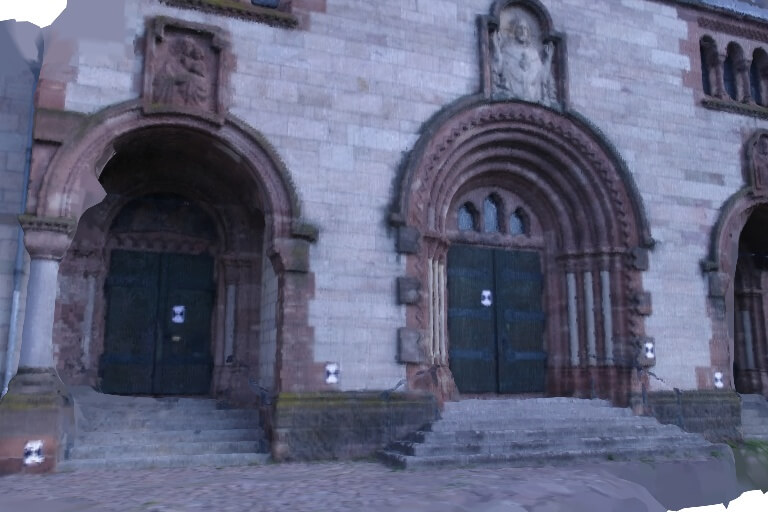} &
		\includegraphics[width=0.23\linewidth]{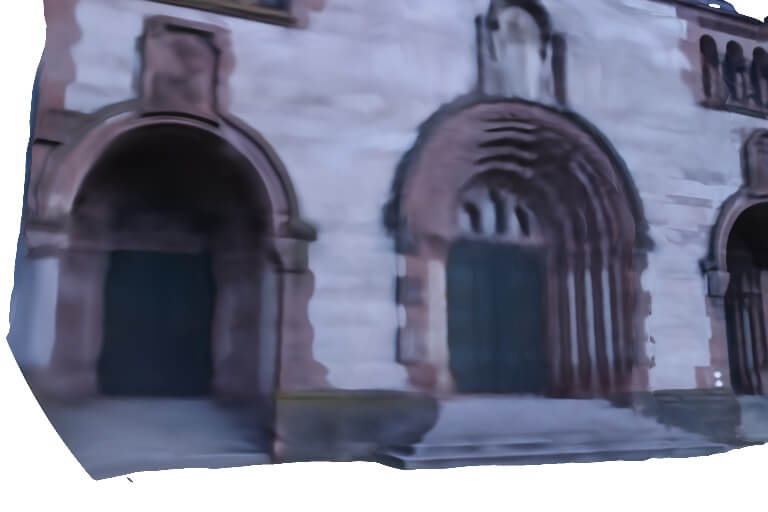} &
		\includegraphics[width=0.23\linewidth]{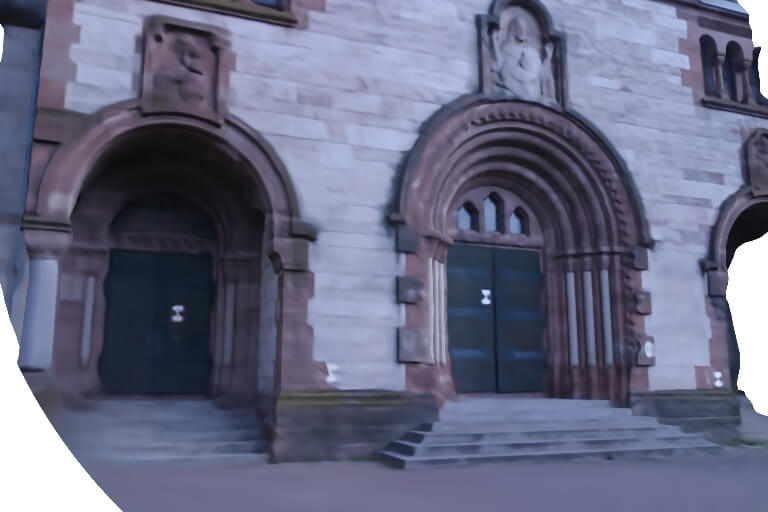} &
		\includegraphics[width=0.23\linewidth]{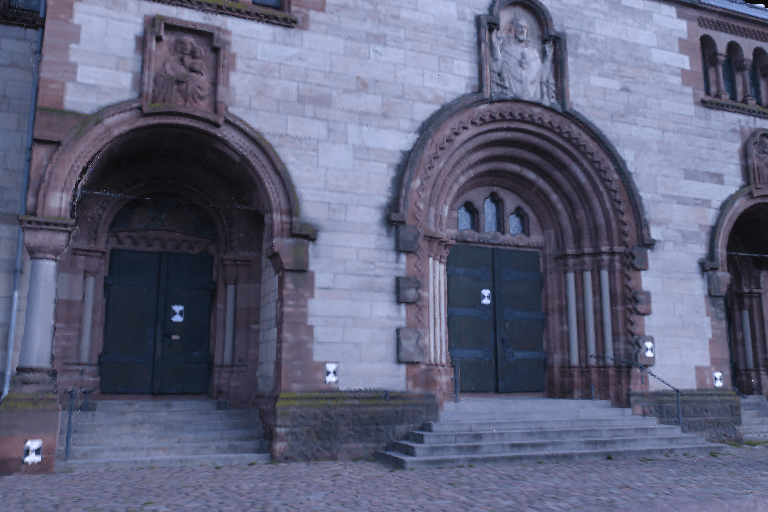} \\
		
		Vis-MVSNet & IDR & MVSDF & Ours
	\end{tabular}
	\caption{Qualitative results on EPFL dataset. Our method is able to generate both high quality mesh and render high quality image.}
	\label{fig:epfl}
\end{figure*}

We further evaluate the computational cost on DTU dataset. All the experiments on computational time are conducted on the same machine with an Intel i9 CPU (32GB RAM) and a NVIDIA 2080Ti GPU (11GB RAM). Table~\ref{tab:timeconsumption} summarizes the computational cost for the compared methods. It can be seen that our proposed approach requires 30 minutes to obtain the 3D reconstruction results while the implicit representation based-methods need several hours. This is because the gradients of MLP decreases as the layer goes forward. The parameters of MLP only change a little at each iteration, which leads to longer training time. Due to the implicit shape representation of MLP, a sphere tracing algorithm is performed to find the intersection of rays and the optimized mesh at each iteration, which consumes a lot of time. We employ the oriented point clouds as the shape representation, where the gradient can be efficiently backpropagated to points and normals. At each iteration, we obtain the optimized mesh from marching cube while rasterization is completed by the differentiable renderer. This is much more efficient than sphere tracing algorithm. 

\begin{table}[htbp]
	\centering
	\caption{Computational time on DTU dataset.}
	\begin{tabular}{c|c|c}
		\toprule
		Methods & Training time & Rendering time\\
		\midrule
		PhySG~\cite{DBLP:conf/cvpr/physg21} & 5.0h & 30s \\
		IDR~\cite{DBLP:conf/nips/idr20} & 6.5h & 30s \\
		MVSDF~\cite{mvsdf21} & 5.5h & 30s \\
		Ours & 30 min & ~0.04s \\
		\bottomrule
	\end{tabular}
	\label{tab:timeconsumption}
\end{table}

In terms of rendering time, our approach is able to run at 25Hz by taking advantage of the physically-based rendering. Due to the short rendeirng time of our method, we render 1000 times and measure the average time. Moreover, we can export mesh and texture from the oriented point clouds and texture grid. In addition, we can export the reconstruction results into the existing rendering engines to enable real-time rendering. For the implicit representation-based methods, it takes dozens of seconds to render an image of resolution $1600 \times 1200$, since a forward network inference is required for each valid pixel. 

\begin{table*}[htbp]
	\centering
	\caption{Quantitative results on EPFL dataset. Our method performs comparable to the state-of-the-art methods.}
	
	\begin{tabular}{l||cc|cc||cc|cc}
		\toprule
		& \multicolumn{4}{c||}{Chamfer ($ \times10^{-2} $)}     & \multicolumn{4}{c}{PSNR}             \\ \cline{2-9} 
		& Vis-MVSNet & IDR   & MVSDF & Ours & Vis-MVSNet & IDR  & MVSDF & Ours  \\ \hline
		Fountain-P11        & 6.12   & 7.88       & 6.84  & \textbf{5.81} & 24.33  & 23.43      & 25.27 & \textbf{26.53} \\
		Herzjesu-P8         & 7.47  & 32.19       & 6.38 & \textbf{6.20} & 23.45  & 24.75      & \textbf{28.75} & 28.51 \\ \hline
		Mean                & 6.80   & 25.30 & 6.61 & \textbf{6.01} & 23.89  & 24.67      & 27.01 & \textbf{27.52} \\
		\bottomrule
	\end{tabular}
	\label{tab:epfl}
\end{table*}

\subsection{Evaluation on EPFL Dataset}

To compare the performance of the scene reconstruction, we evaluate our proposed method on EPFL dataset. We conduct experiments on Fountain-P11 and Herzjesu-P8, which have ground truth meshes. We compare our method against Vis-MVSNet, IDR and MVSDF. As EPFL dataset does not provide the mask, we generate it by projecting the ground truth mesh onto the image. The experimental results of Vis-MVSNet and IDR are from MVSDF~\cite{mvsdf21}.

\begin{figure*}[htbp]
	\centering
	\begin{tabular}{@{\hskip2pt}c@{\hskip2pt}@{\hskip2pt}c@{\hskip2pt}@{\hskip2pt}c@{\hskip2pt}@{\hskip2pt}c@{\hskip2pt}@{\hskip2pt}c@{\hskip2pt}}
		\begin{tikzpicture}\node[above right, inner sep=0](image) at (0,0) {\includegraphics[width=0.19\linewidth]{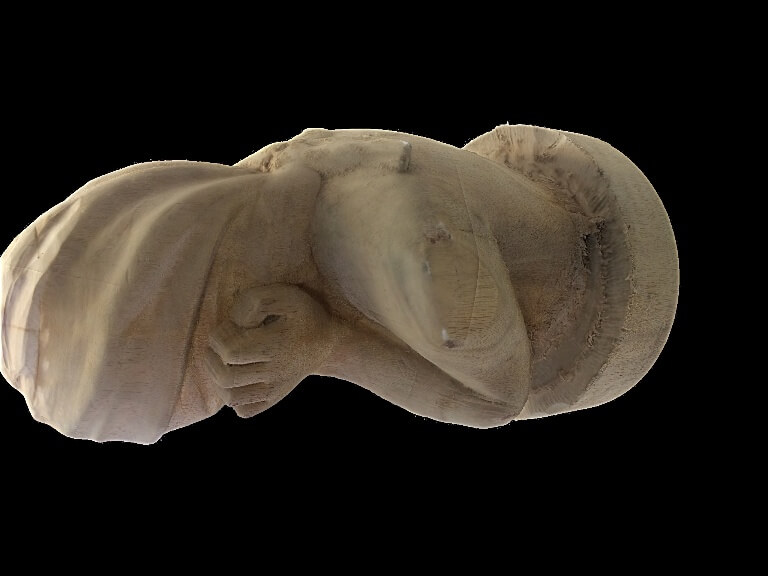}};
		\end{tikzpicture} &
		\begin{tikzpicture}\node[above right, inner sep=0](image) at (0,0) {\includegraphics[width=0.19\linewidth]{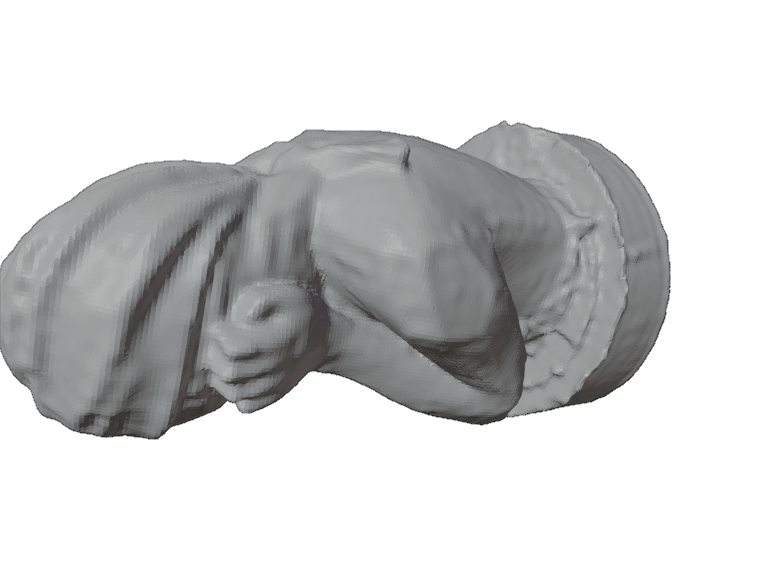}};
		\end{tikzpicture} &
		\begin{tikzpicture}\node[above right, inner sep=0](image) at (0,0) {\includegraphics[width=0.19\linewidth]{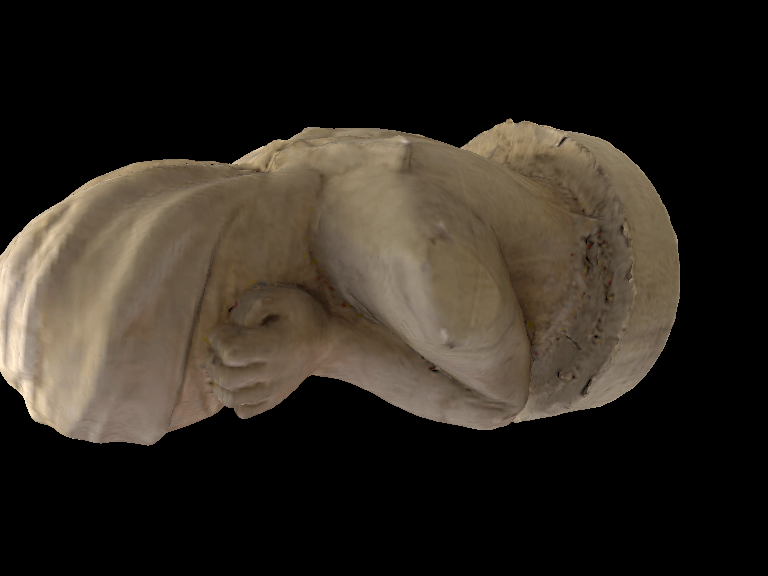}};
		\end{tikzpicture} &
		\begin{tikzpicture}\node[above right, inner sep=0](image) at (0,0) {\includegraphics[width=0.19\linewidth]{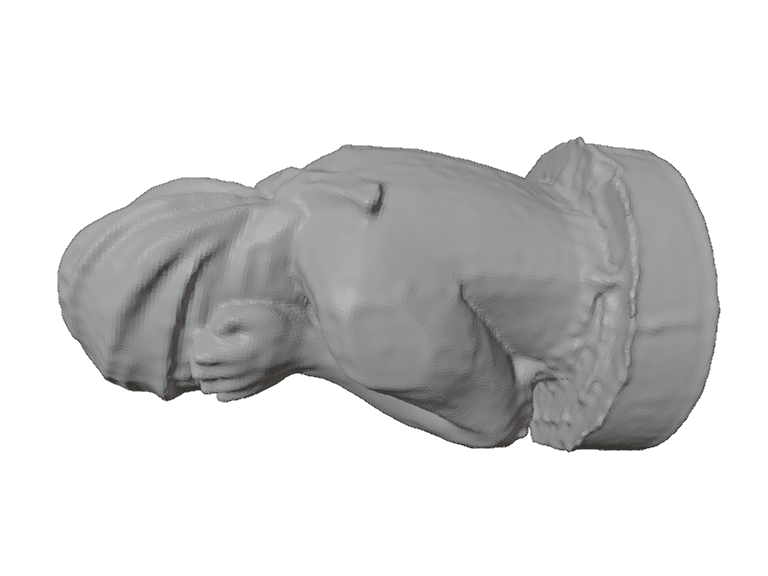}};
		\end{tikzpicture} & 
		\begin{tikzpicture}\node[above right, inner sep=0](image) at (0,0) {\includegraphics[width=0.19\linewidth]{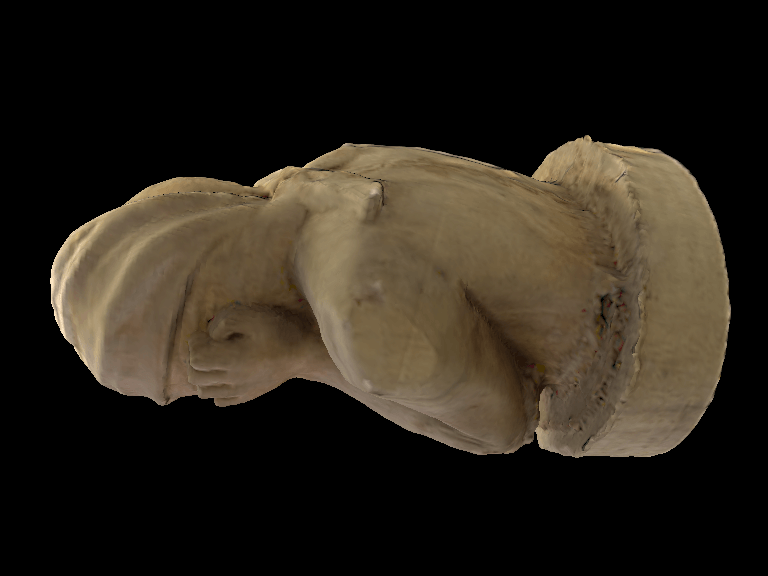}};
		\end{tikzpicture} \\
		
		\begin{tikzpicture}\node[above right, inner sep=0](image) at (0,0) {\includegraphics[width=0.19\linewidth]{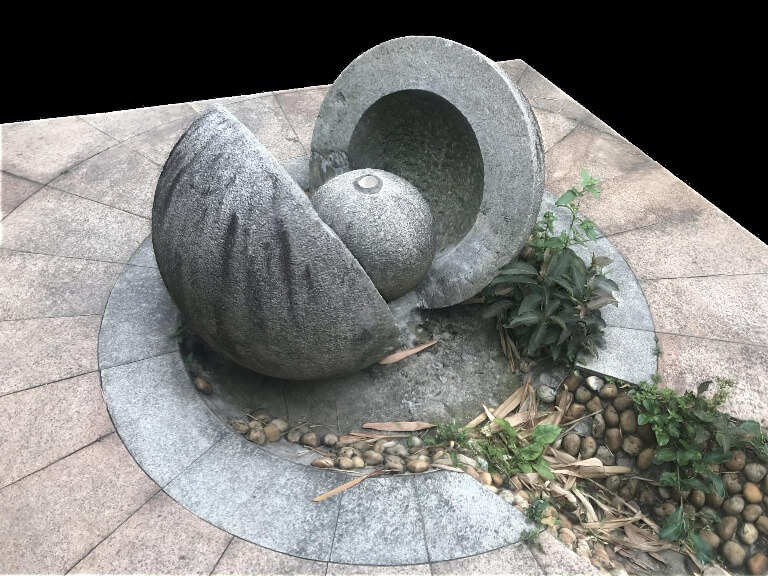}};
		\end{tikzpicture} &
		\begin{tikzpicture}\node[above right, inner sep=0](image) at (0,0) {\includegraphics[width=0.19\linewidth]{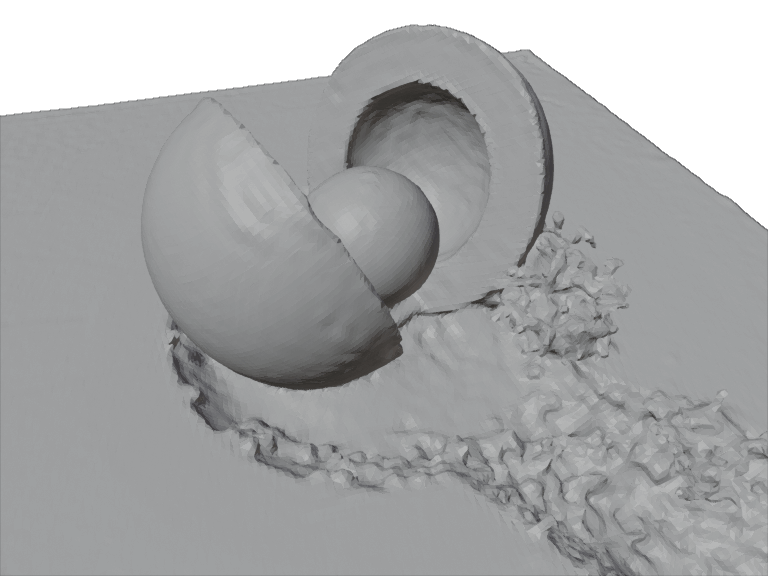}};
		\end{tikzpicture} &
		\begin{tikzpicture}\node[above right, inner sep=0](image) at (0,0) {\includegraphics[width=0.19\linewidth]{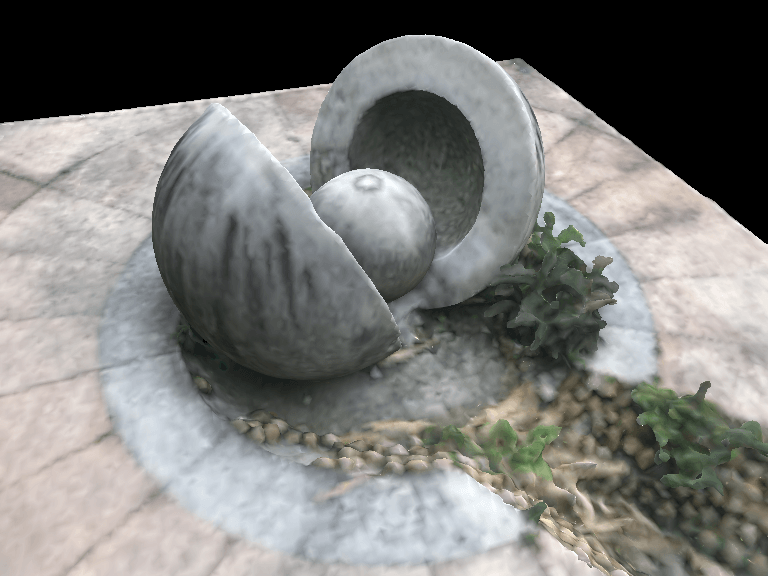}}; 
		\end{tikzpicture} &
		\begin{tikzpicture}\node[above right, inner sep=0](image) at (0,0) {\includegraphics[width=0.19\linewidth]{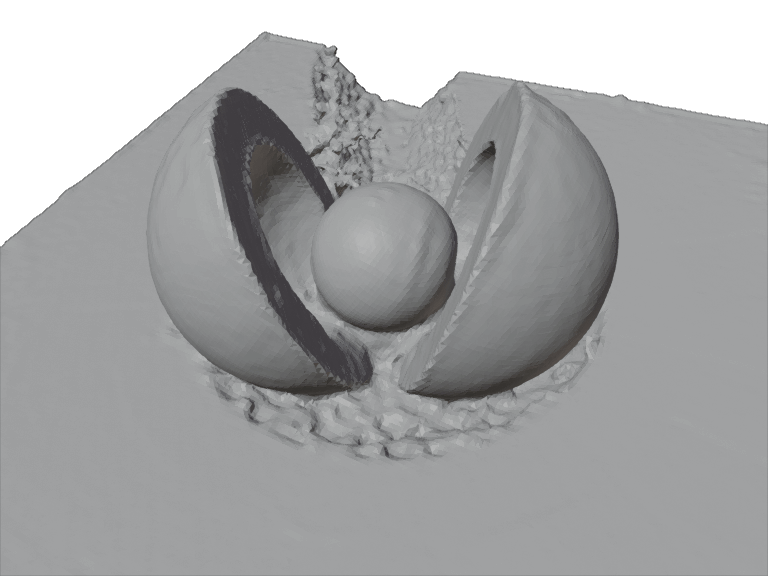}};
		\end{tikzpicture} & 
		\begin{tikzpicture}\node[above right, inner sep=0](image) at (0,0) {\includegraphics[width=0.19\linewidth]{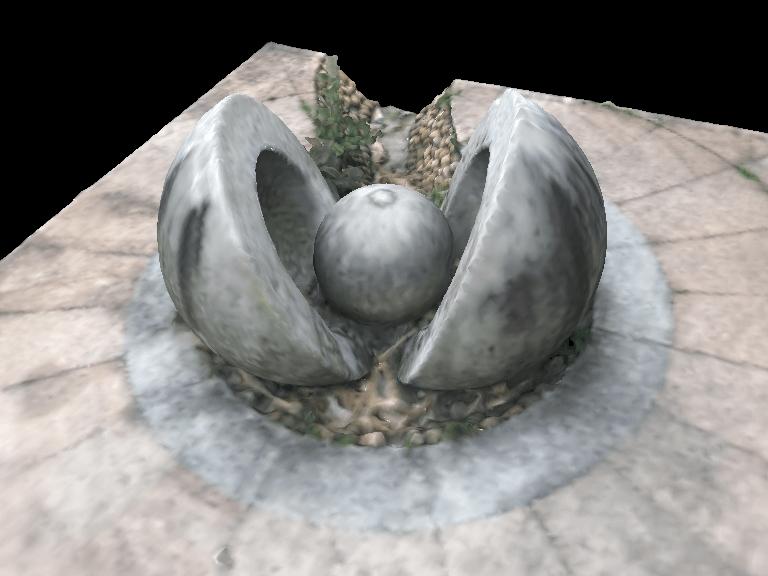}}; 
		\end{tikzpicture} \\
		
		\begin{tikzpicture}\node[above right, inner sep=0](image) at (0,0) {\includegraphics[width=0.19\linewidth]{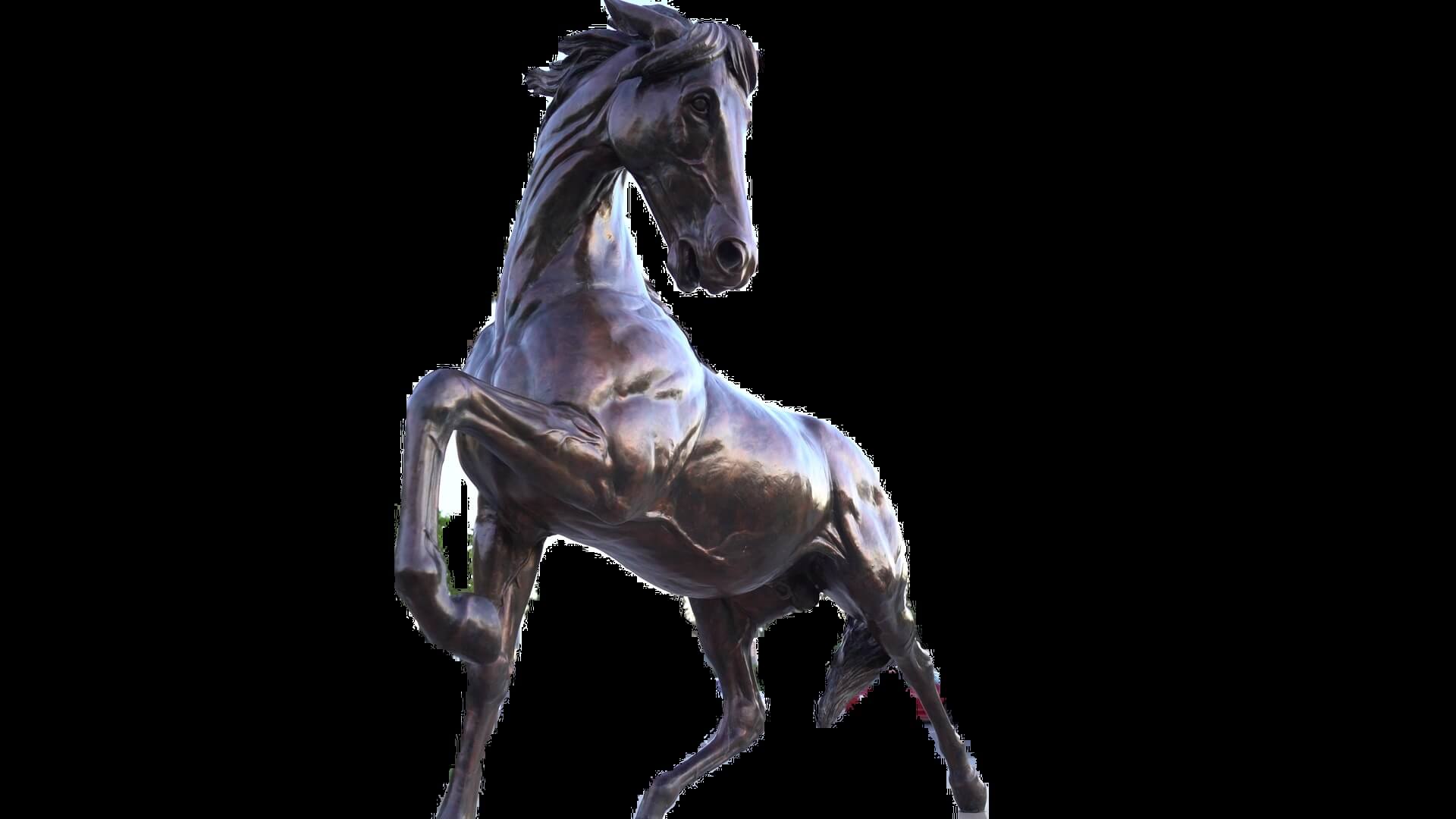}};
		\end{tikzpicture} &
		\begin{tikzpicture}\node[above right, inner sep=0](image) at (0,0) {\includegraphics[width=0.19\linewidth]{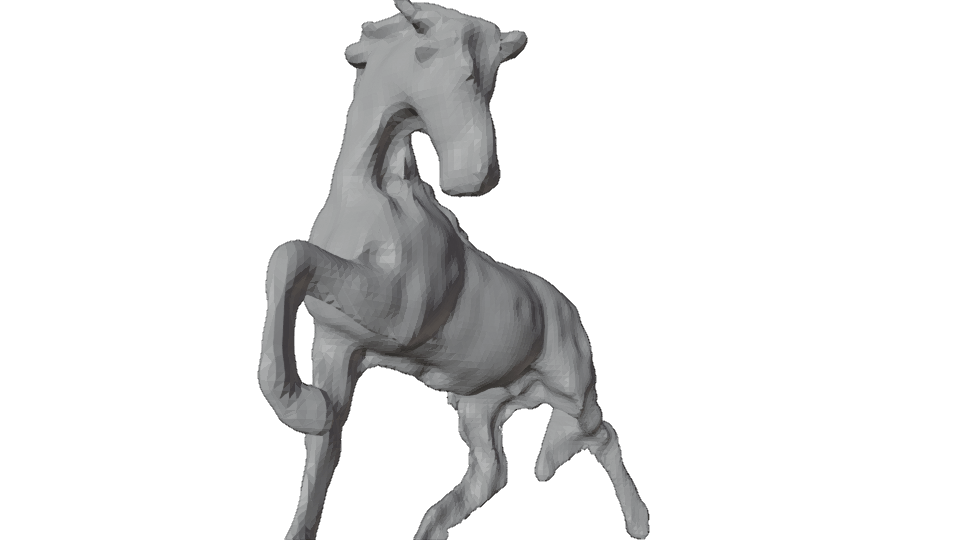}};
		\end{tikzpicture} &
		\begin{tikzpicture}\node[above right, inner sep=0](image) at (0,0) {\includegraphics[width=0.19\linewidth]{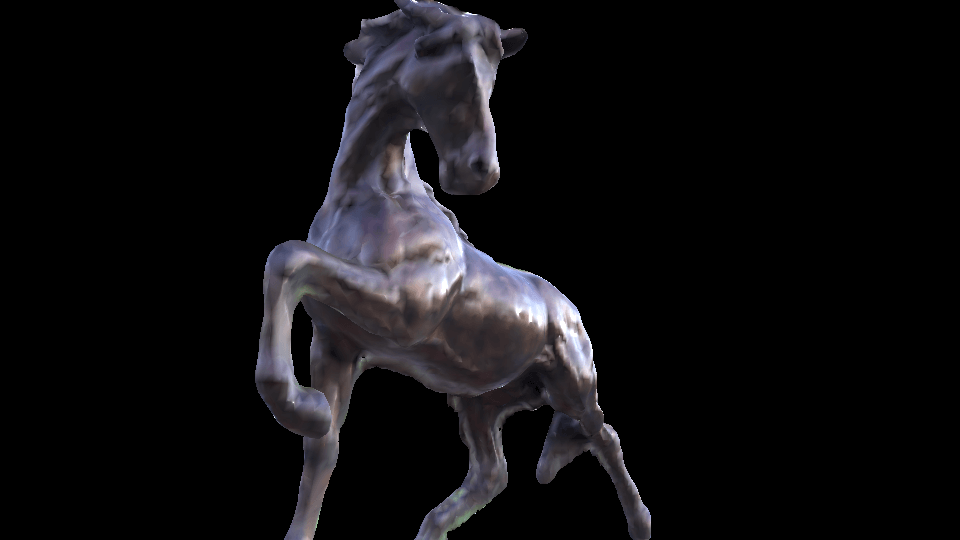}};
		\end{tikzpicture} &
		\begin{tikzpicture}\node[above right, inner sep=0](image) at (0,0) {\includegraphics[width=0.19\linewidth]{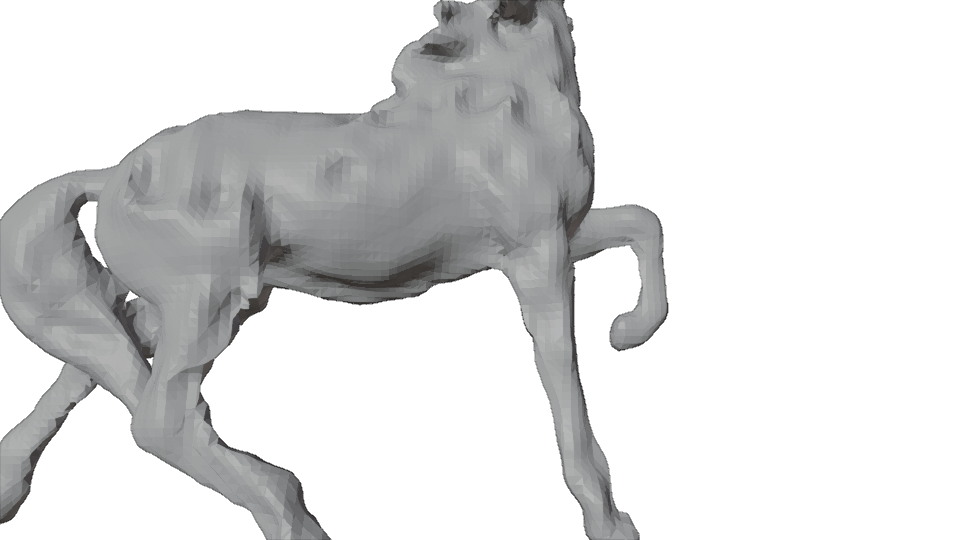}};
		\end{tikzpicture} & 
		\begin{tikzpicture}\node[above right, inner sep=0](image) at (0,0) {\includegraphics[width=0.19\linewidth]{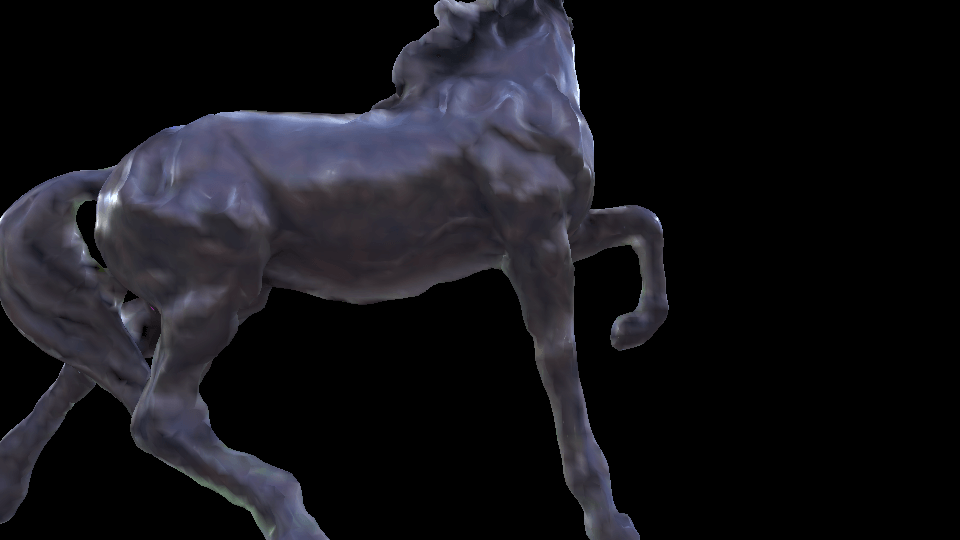}}; 
		\end{tikzpicture} \\
		
		\begin{tikzpicture}\node[above right, inner sep=0](image) at (0,0) {\includegraphics[width=0.19\linewidth]{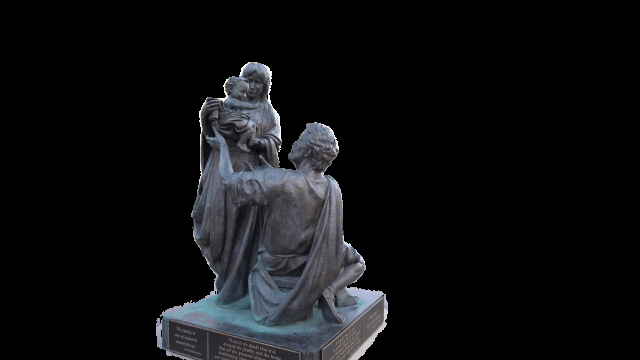}};
		\end{tikzpicture} &
		\begin{tikzpicture}\node[above right, inner sep=0](image) at (0,0) {\includegraphics[width=0.19\linewidth]{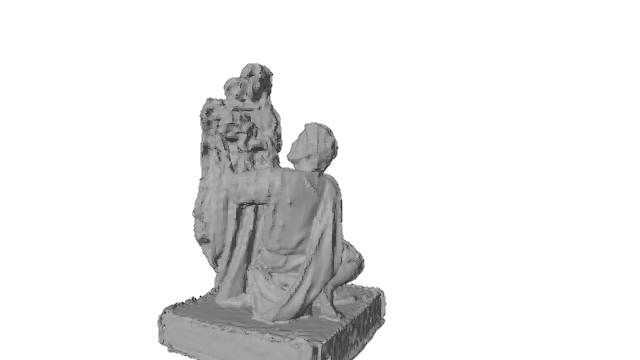}};
		\end{tikzpicture} &
		\begin{tikzpicture}\node[above right, inner sep=0](image) at (0,0) {\includegraphics[width=0.19\linewidth]{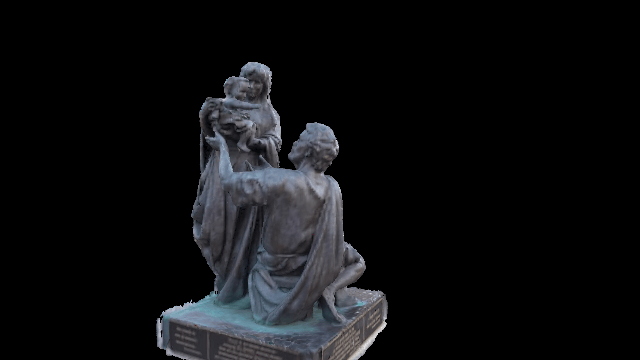}};
		\end{tikzpicture} &
		\begin{tikzpicture}\node[above right, inner sep=0](image) at (0,0) {\includegraphics[width=0.19\linewidth]{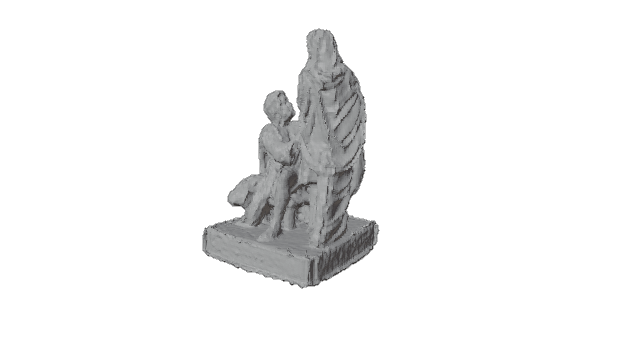}};
		\end{tikzpicture} & 
		\begin{tikzpicture}\node[above right, inner sep=0](image) at (0,0) {\includegraphics[width=0.19\linewidth]{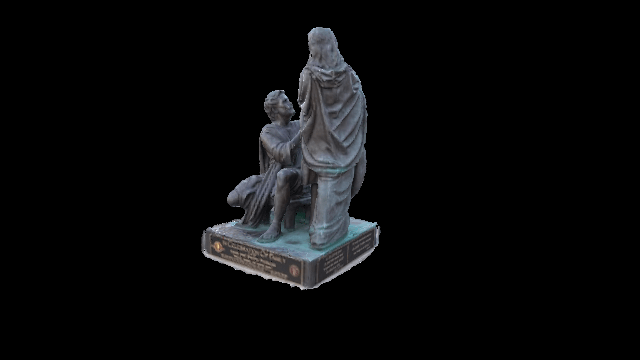}}; 
		\end{tikzpicture} \\
		
		Original image & recovered shape & rendered image & recovered shape & rendered image
	\end{tabular}
	\caption{Qualitative results on BlendedMVS dataset and Tanks and Temples dataset. The top two rows are from the BlendedMVS dataset, and the bottom two rows are from Tanks and Temples dataset.}
	\label{fig:additional}
\end{figure*}

\begin{figure*}[htbp]
	\centering
	\begin{tabular}{@{\hskip1pt}c@{\hskip1pt}@{\hskip1pt}c@{\hskip1pt}@{\hskip1pt}c@{\hskip1pt}@{\hskip1pt}c@{\hskip1pt}@{\hskip1pt}c@{\hskip1pt}}
		\begin{tikzpicture}\node[above right, inner sep=0](image) at (0,0) {\includegraphics[width=0.23\linewidth]{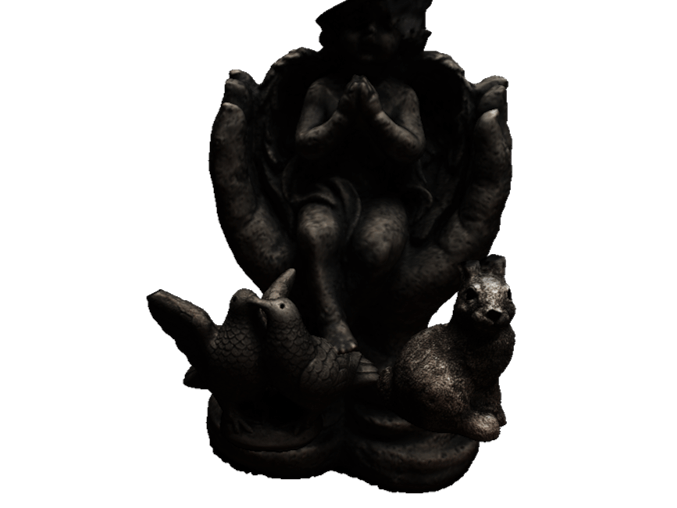}};
		\end{tikzpicture} &
		\begin{tikzpicture}\node[above right, inner sep=0](image) at (0,0) {\includegraphics[width=0.23\linewidth]{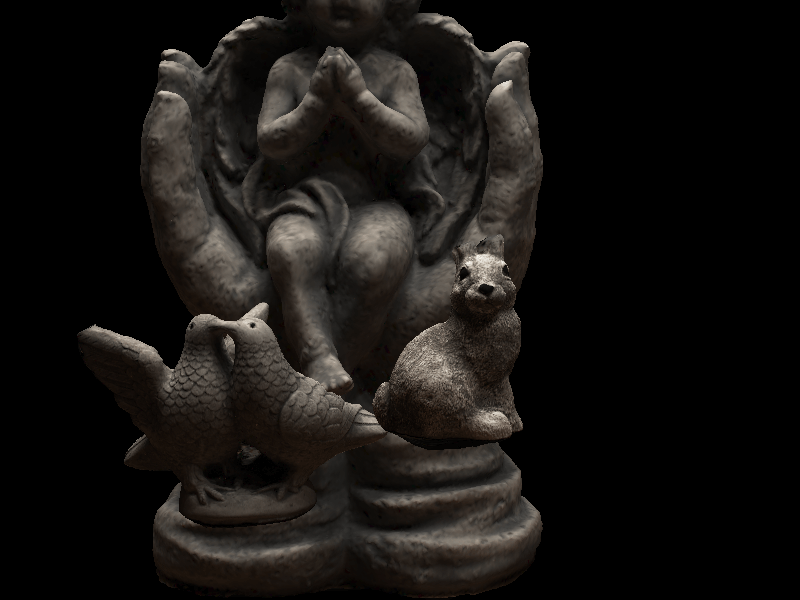}};
		\end{tikzpicture} &
		\begin{tikzpicture}\node[above right, inner sep=0](image) at (0,0) {\includegraphics[width=0.23\linewidth]{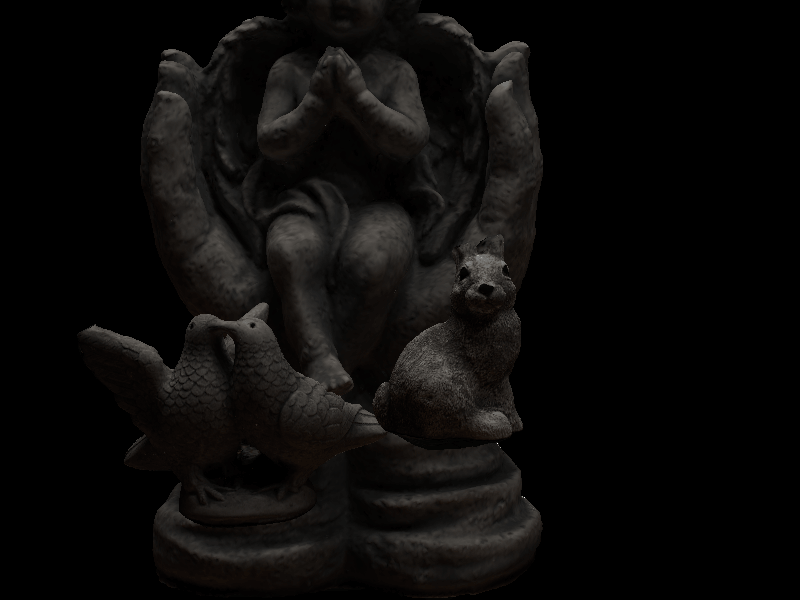}};
		\end{tikzpicture} &
		\begin{tikzpicture}\node[above right, inner sep=0](image) at (0,0) {\includegraphics[width=0.23\linewidth]{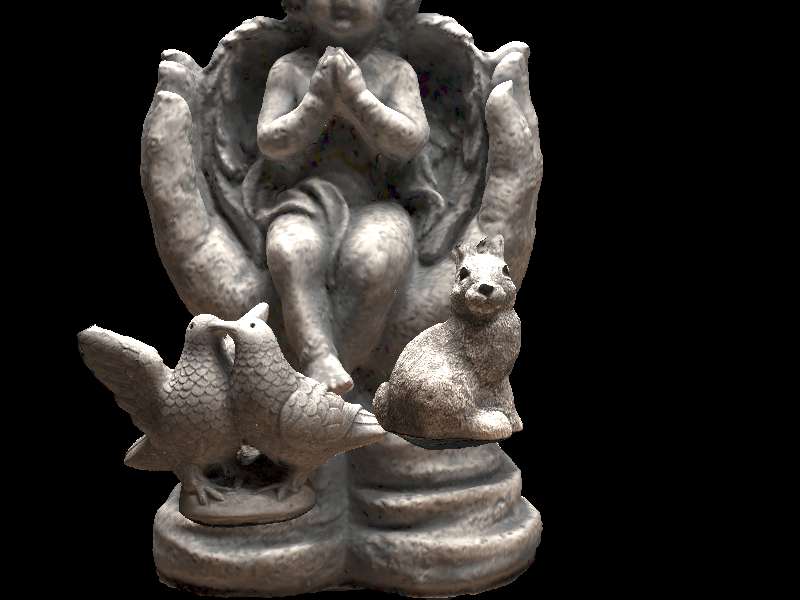}};
		\end{tikzpicture} & \\
		
		\begin{tikzpicture}\node[above right, inner sep=0](image) at (0,0) {\includegraphics[width=0.23\linewidth]{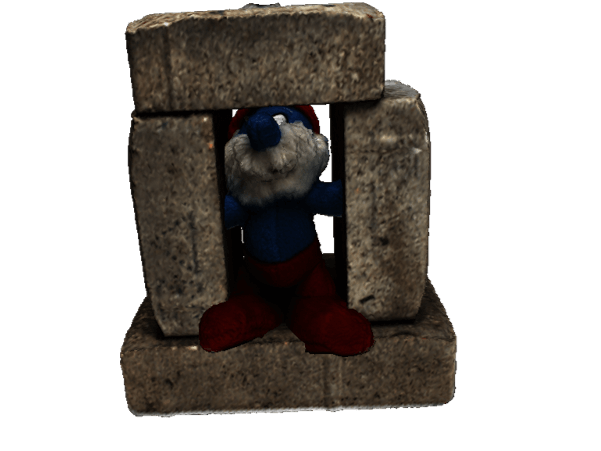}};
		\end{tikzpicture} &
		\begin{tikzpicture}\node[above right, inner sep=0](image) at (0,0) {\includegraphics[width=0.23\linewidth]{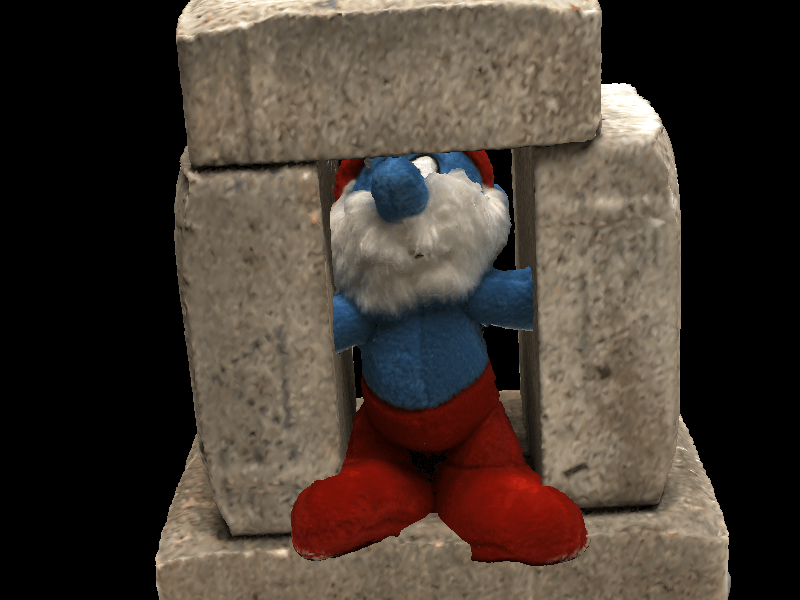}};
		\end{tikzpicture} &
		\begin{tikzpicture}\node[above right, inner sep=0](image) at (0,0) {\includegraphics[width=0.23\linewidth]{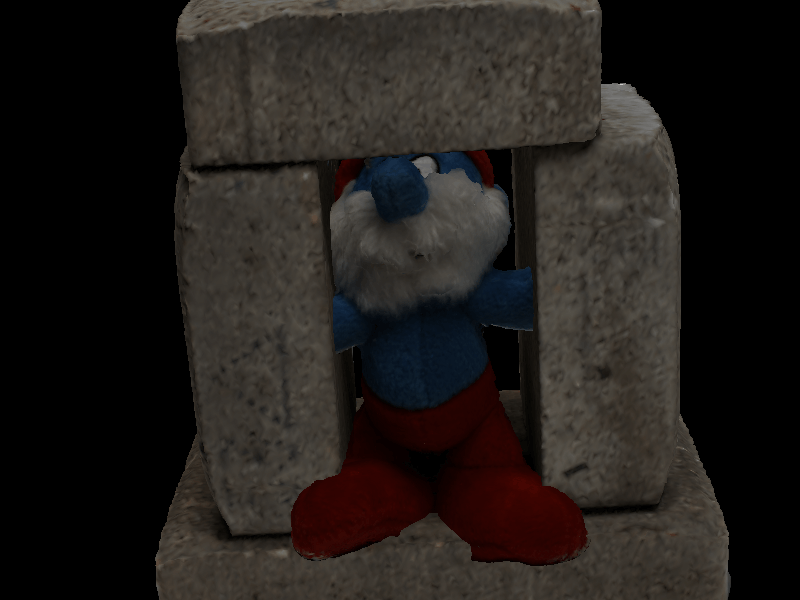}};
		\end{tikzpicture} &
		\begin{tikzpicture}\node[above right, inner sep=0](image) at (0,0) {\includegraphics[width=0.23\linewidth]{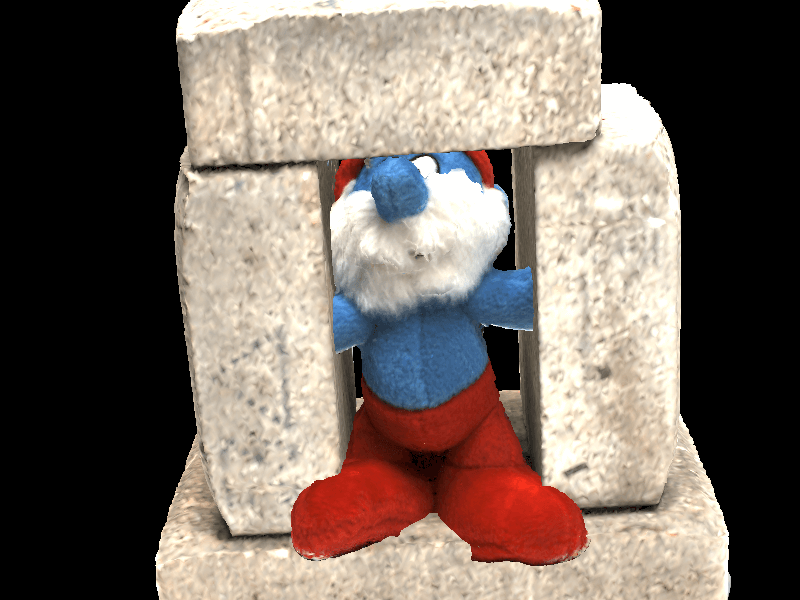}};
		\end{tikzpicture} & \\

		\begin{tikzpicture}\node[above right, inner sep=0](image) at (0,0) {\includegraphics[width=0.23\linewidth]{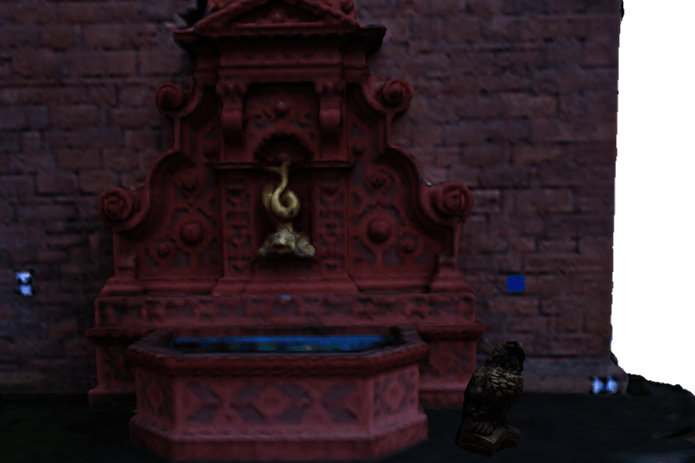}};
		\end{tikzpicture} &
		\begin{tikzpicture}\node[above right, inner sep=0](image) at (0,0) {\includegraphics[width=0.23\linewidth]{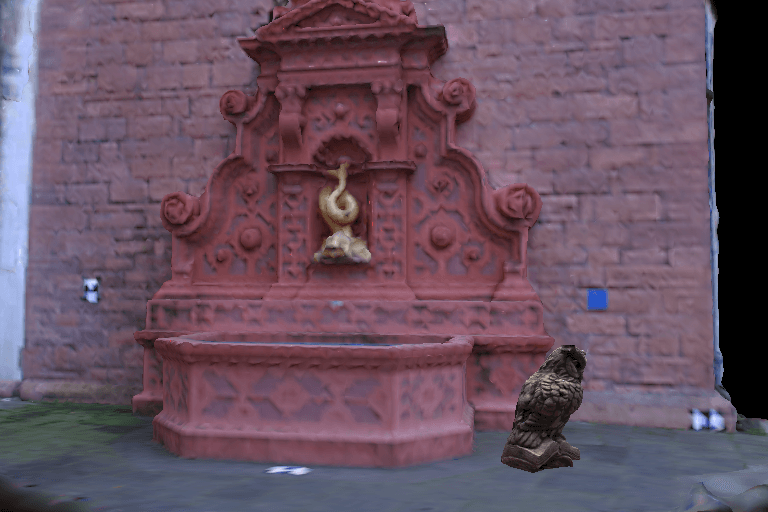}};
		\end{tikzpicture} &
		\begin{tikzpicture}\node[above right, inner sep=0](image) at (0,0) {\includegraphics[width=0.23\linewidth]{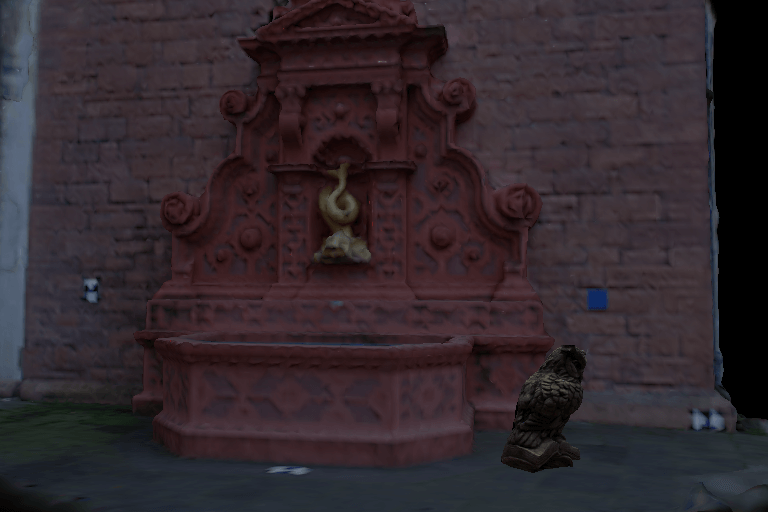}};
		\end{tikzpicture} &
		\begin{tikzpicture}\node[above right, inner sep=0](image) at (0,0) {\includegraphics[width=0.23\linewidth]{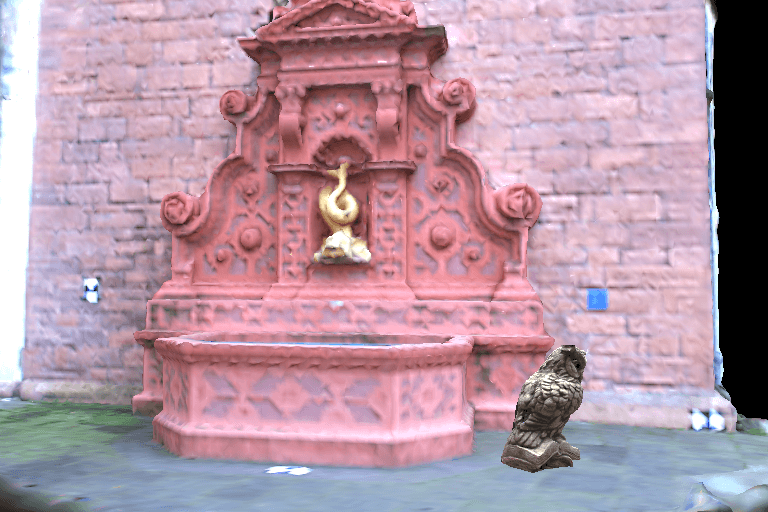}};
		\end{tikzpicture} &  \\
		
		\begin{tikzpicture}\node[above right, inner sep=0](image) at (0,0) {\includegraphics[width=0.23\linewidth]{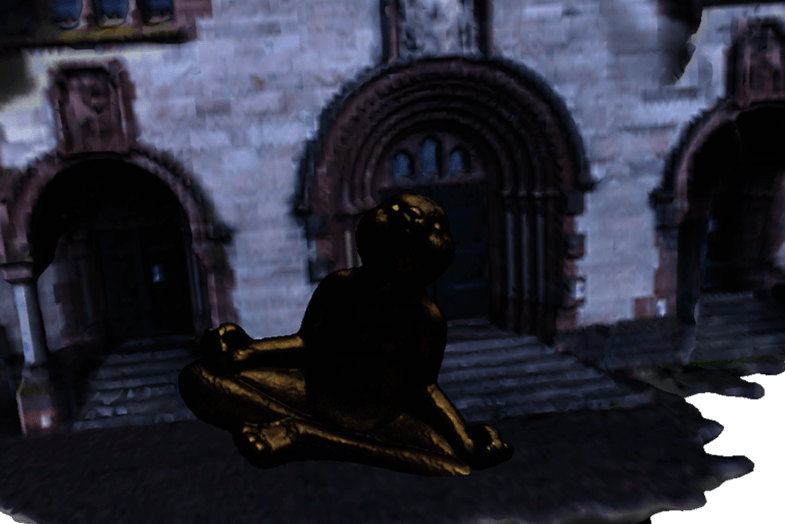}};
		\end{tikzpicture} &
		\begin{tikzpicture}\node[above right, inner sep=0](image) at (0,0) {\includegraphics[width=0.23\linewidth]{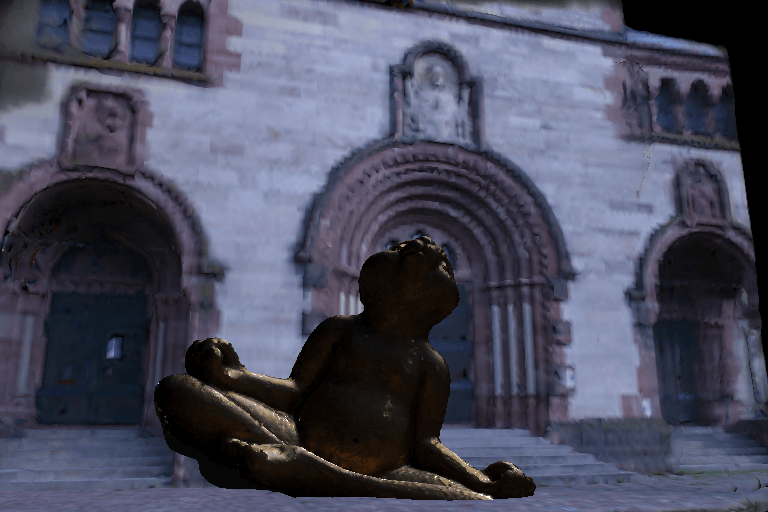}};
		\end{tikzpicture} &
		
		\begin{tikzpicture}\node[above right, inner sep=0](image) at (0,0) {\includegraphics[width=0.23\linewidth]{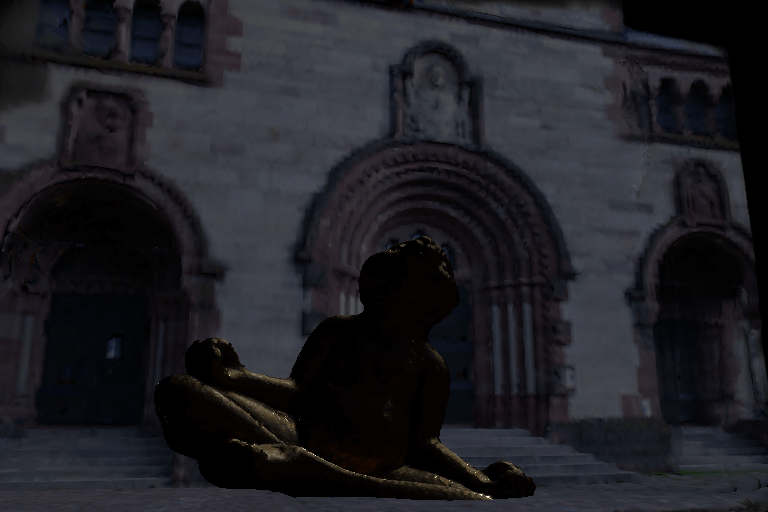}};
		\end{tikzpicture} &
		\begin{tikzpicture}\node[above right, inner sep=0](image) at (0,0) {\includegraphics[width=0.23\linewidth]{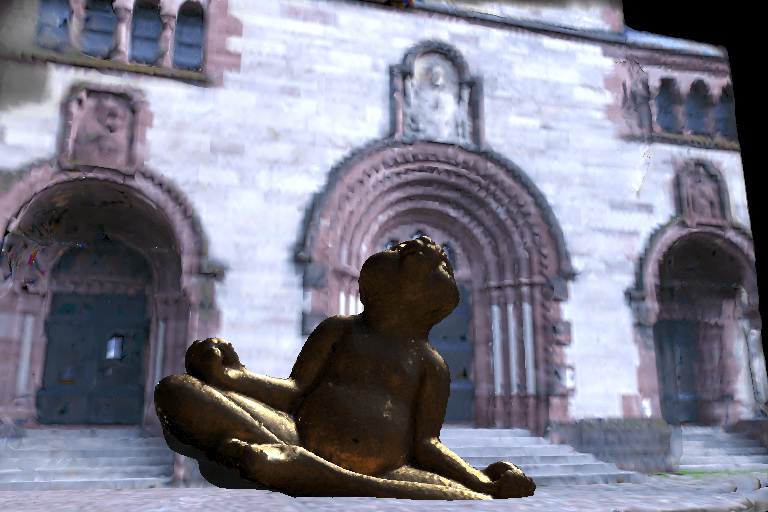}};
		\end{tikzpicture} &  \\
		
		Textured mesh & rendered image & relighting & relighting
	\end{tabular}
	\caption{Qualitative results on mixing and relighting rendering. We present the textured mesh and the rendered images in different lighting.}
	\label{fig:mix}
\end{figure*}

Table~\ref{tab:epfl} gives the quantitative results. We use the same evaluation metrics as the DTU dataset. It can be seen that our approach performs better than the state-of-the-art methods on both synthesized images and reconstructed meshes. Fig.~\ref{fig:epfl} shows the reconstruction results on the EPFL dataset. Obviously, our method both estimates high-quality mesh and renders the high quality images. Compared with other methods, our proposed approach recovers more details such like the trace of blocks, a nd the decoration on door. The reconstructed mesh of IDR has the inflated surfaces, since it is difficult to recover the accurate shape only from color constraints. By taking advantage of the extra supervision from multi-view stereo, both MVSDF and our presented method can recover the correct geometry. The experimental results on EPFL dataset show that our proposed method is not only effective for small indoor objects, but also achieves promising results for in-the-wild  outdoor scenes.

\subsection{Qualitative Results}

We provide the qualitative results on BlendedMVS, Tanks and Temples dataset. The masks of the Horse and Family are generated by an off-the-shelf image segmentation framework~\cite{DBLP:conf/cvpr/pointrend20}, and the camera parameters are estimated by COLMAP~\cite{DBLP:conf/eccv/colmap16}. Fig.~\ref{fig:additional} shows several examples. It demonstrates that our method is able to reconstruct the accurate geometry and texture from both synthesized images and in-the-wild images. The results in the second row of Fig.~\ref{fig:additional} show that our proposed method is able to extract the accurate mesh and textures from multi-view images without masks as well.

As our method explicitly extracts the triangulated mesh and texture, which can be combined and rendered with the arbitrary lighting. Instead of using two MLP to represent the implicit surface and radiance field, we employ the point-based mesh and physically-based rendering that decouples the texture and illumination. We can synthesize the images by combing the extracted meshes under various illuminations. Mesh composition and illumination changes are difficult for those MLP representation as the geometry and color information are encoded in MLP. While our proposed method can extract mesh and texture from the oriented point clouds and texture grid efficiently. We arbitrarily combine meshes and change the environment map, where the hidden surface removal can be done efficiently by the renderer. Fig.~\ref{fig:mix} illustrates the qualitative results. We present the textured meshes and rendered images in different lighting. It shows that we can synthesize the images with different meshes and various lighting. Moreover, the synthesized images are with high quality. 

\begin{figure*}[htbp]
	\centering
	\begin{tabular}{@{\hskip1pt}c@{\hskip1pt}@{\hskip1pt}c@{\hskip1pt}@{\hskip1pt}c@{\hskip1pt}@{\hskip1pt}c@{\hskip1pt}@{\hskip1pt}c@{\hskip1pt}}
		\begin{tikzpicture}\node[above right, inner sep=0](image) at (0,0) {\includegraphics[width=0.23\linewidth]{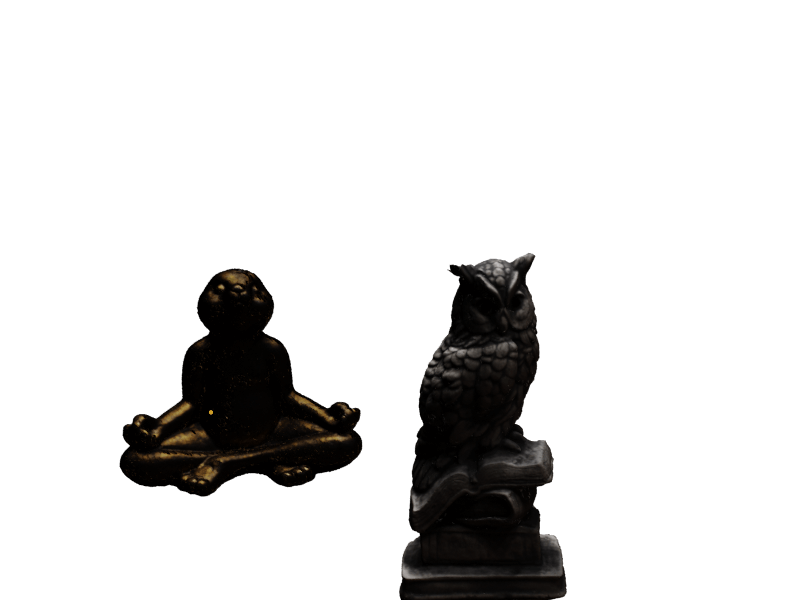}};
		\end{tikzpicture} &
		\begin{tikzpicture}\node[above right, inner sep=0](image) at (0,0) {\includegraphics[width=0.23\linewidth]{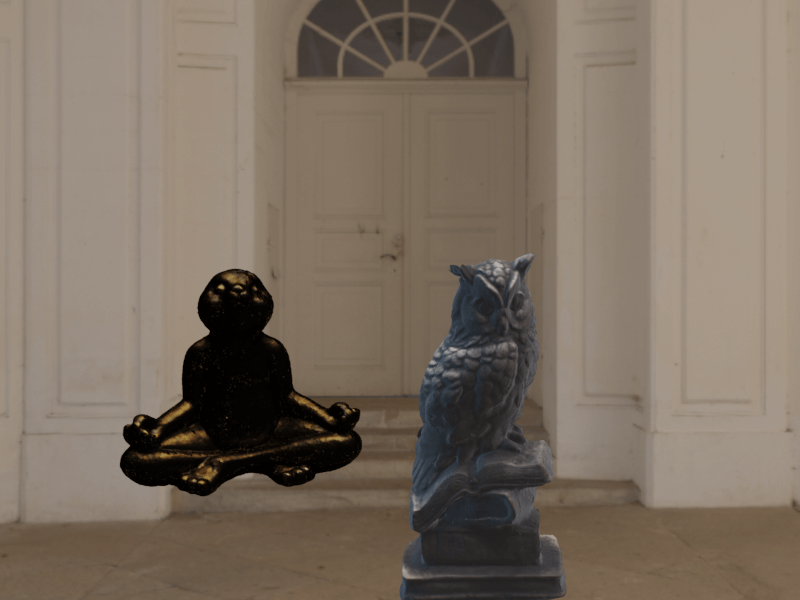}};
		\end{tikzpicture} &
		\begin{tikzpicture}\node[above right, inner sep=0](image) at (0,0) {\includegraphics[width=0.23\linewidth]{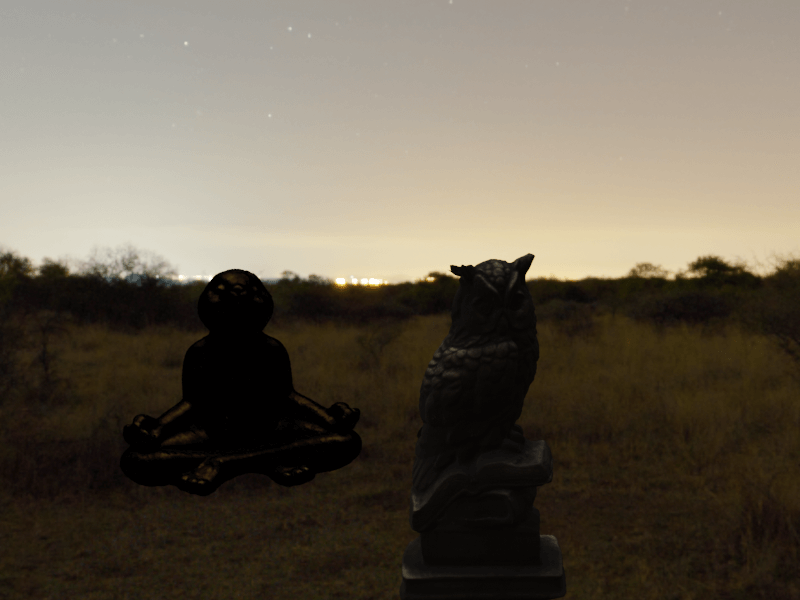}};
		\end{tikzpicture} &
		\begin{tikzpicture}\node[above right, inner sep=0](image) at (0,0) {\includegraphics[width=0.23\linewidth]{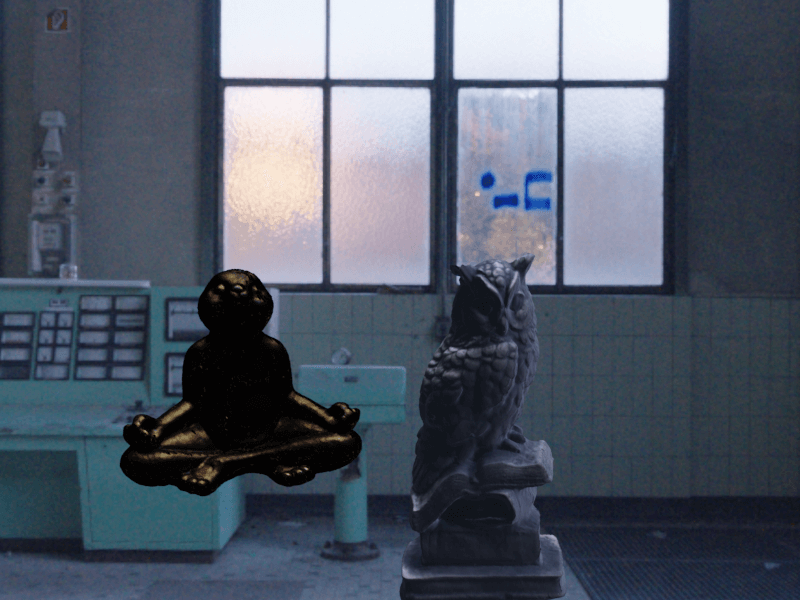}};
		\end{tikzpicture} & \\
		
		\begin{tikzpicture}\node[above right, inner sep=0](image) at (0,0) {\includegraphics[width=0.23\linewidth]{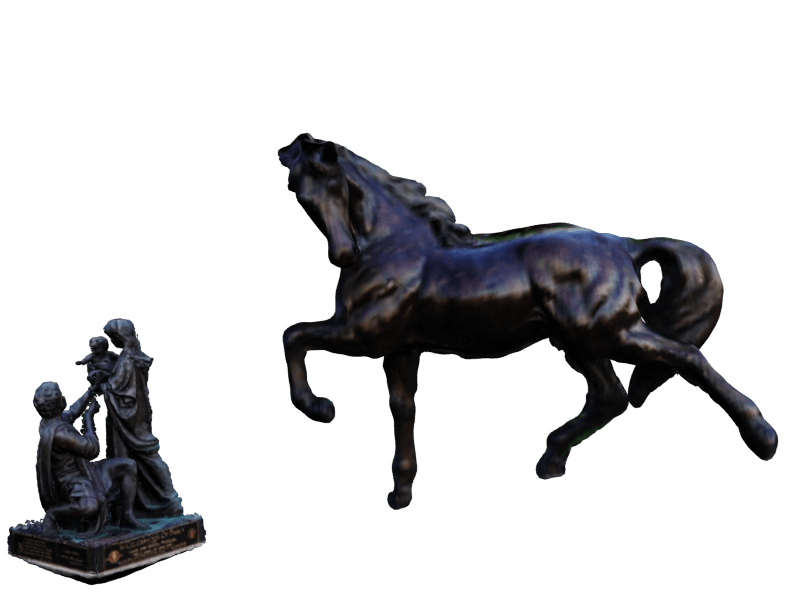}};
		\end{tikzpicture} &
		\begin{tikzpicture}\node[above right, inner sep=0](image) at (0,0) {\includegraphics[width=0.23\linewidth]{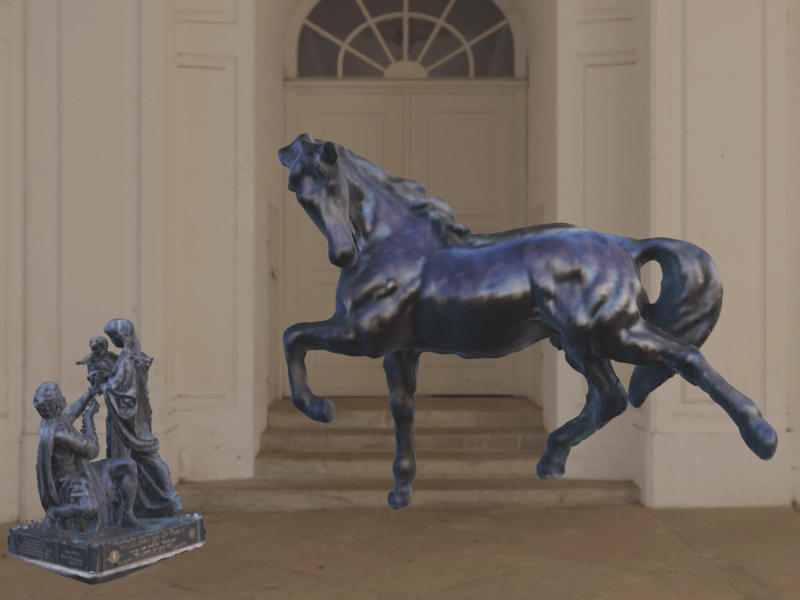}};
		\end{tikzpicture} &
		\begin{tikzpicture}\node[above right, inner sep=0](image) at (0,0) {\includegraphics[width=0.23\linewidth]{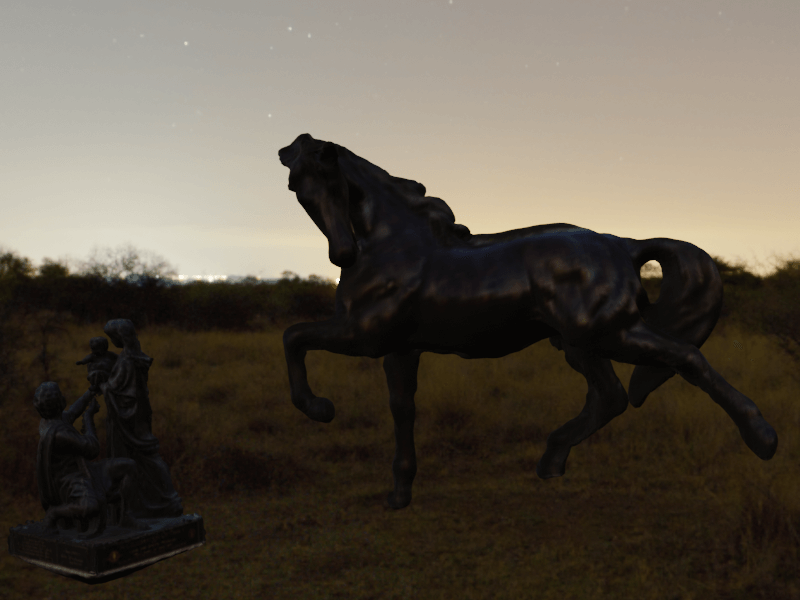}};
		\end{tikzpicture} &
		\begin{tikzpicture}\node[above right, inner sep=0](image) at (0,0) {\includegraphics[width=0.23\linewidth]{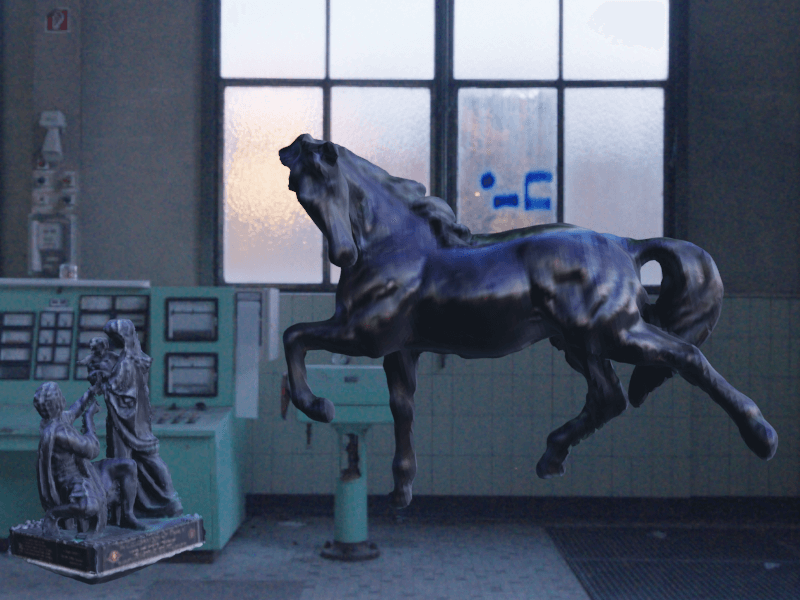}};
		\end{tikzpicture} & \\
		
		\begin{tikzpicture}\node[above right, inner sep=0](image) at (0,0) {\includegraphics[width=0.23\linewidth]{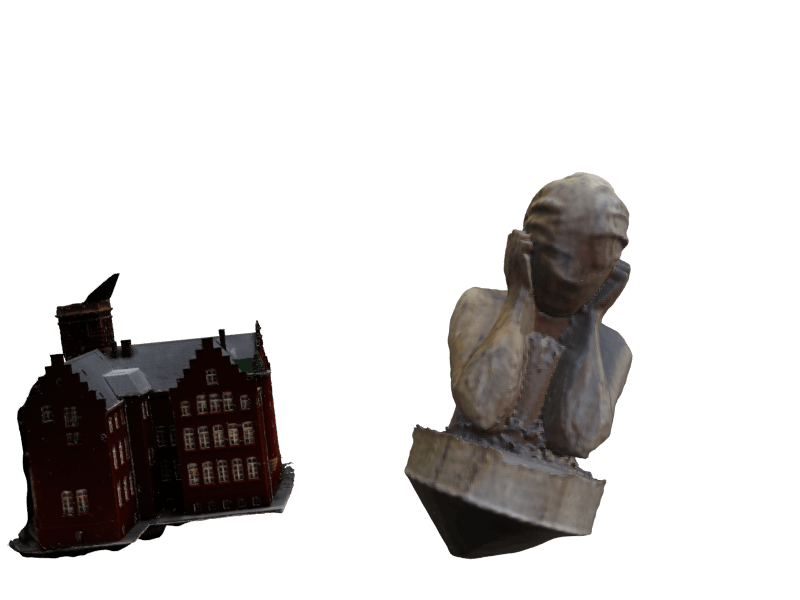}};
		\end{tikzpicture} &
		\begin{tikzpicture}\node[above right, inner sep=0](image) at (0,0) {\includegraphics[width=0.23\linewidth]{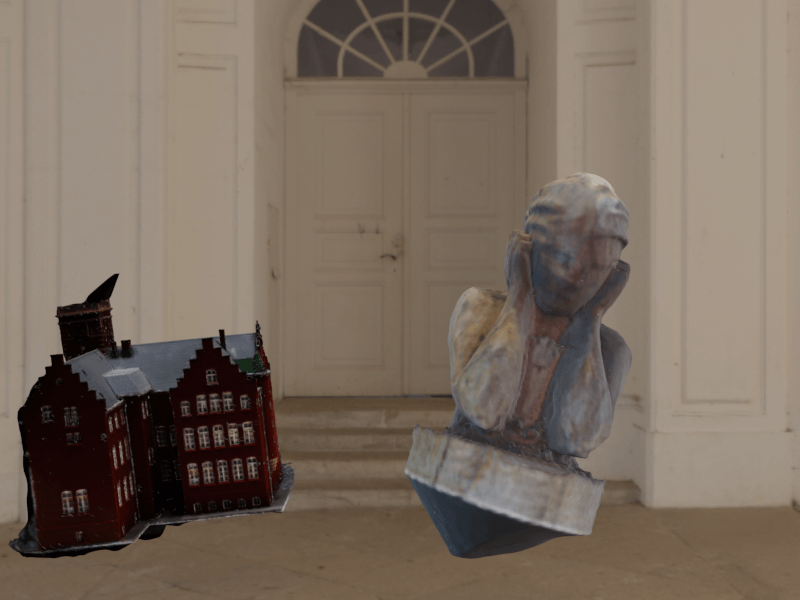}};
		\end{tikzpicture} &
		\begin{tikzpicture}\node[above right, inner sep=0](image) at (0,0) {\includegraphics[width=0.23\linewidth]{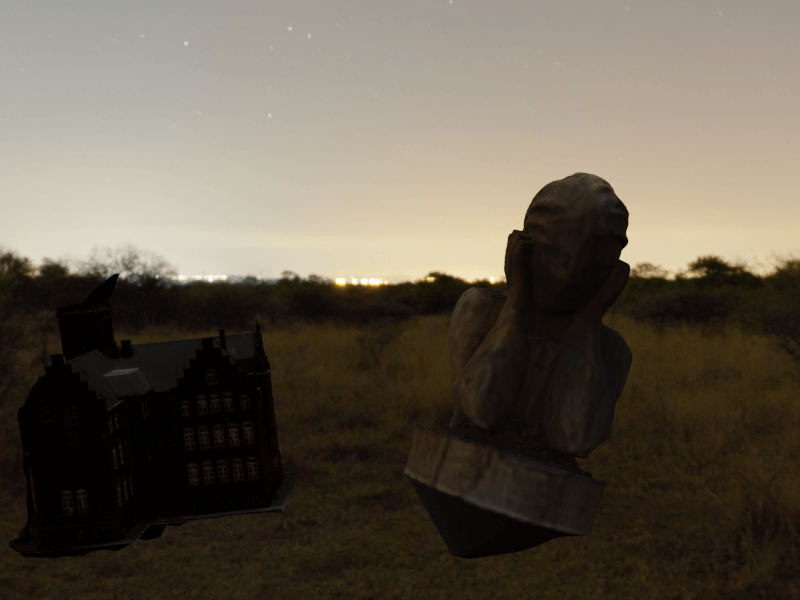}}; 
		\end{tikzpicture} &
		\begin{tikzpicture}\node[above right, inner sep=0](image) at (0,0) {\includegraphics[width=0.23\linewidth]{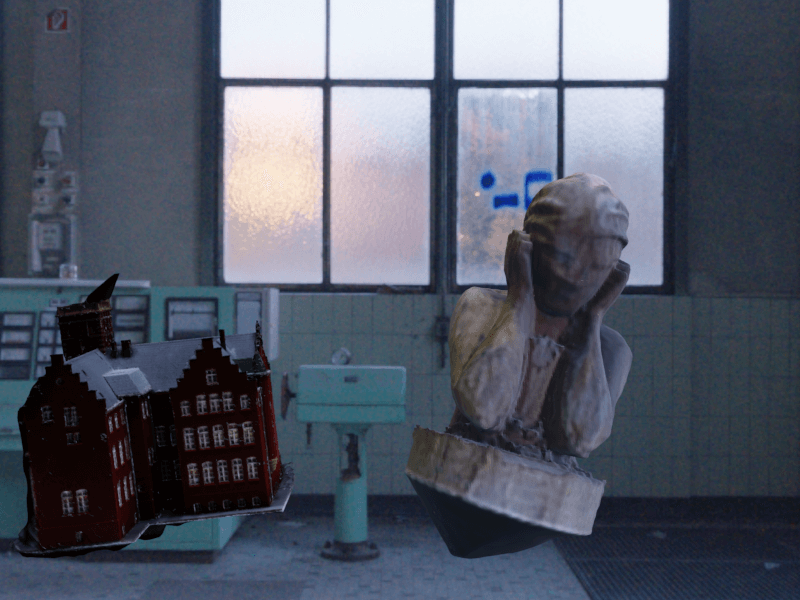}};
		\end{tikzpicture} &  \\
		
		Textured mesh & rendered image & relighting & relighting
	\end{tabular}
	\caption{Qualitative results on mixing and relighting rendered by Blender. We present the textured mesh and the rendered images in different lighting.}
	\label{fig:blendermix}
\end{figure*}

We treat every pixel of the environment map as a light source, which greatly increases the GPU memory consumption. As the resolution of environment map is quite small, the relighting results are not very obvious. Our reconstruction outputs are fully compatible with the existing rendering engines, such as Unreal Engine~\cite{unreal}, Blender~\cite{blender} and so on. Although PhysG~\cite{DBLP:conf/cvpr/physg21} can render realistic images with different lighting, Sphere Gaussians are used to represent texture and lighting. It is not compatible with current rendering engines. We can export the reconstruction results into these engines to generate the photo-realistic images with complex lighting. Fig.~\ref{fig:blendermix} shows the results rendered by Blender. It can be seen that we can render photo-realistic images by our reconstructed meshes under arbitrary complex lighting.

\subsection{Ablation Studies}

\begin{figure}[htbp]
	\centering
	\begin{tabular}{@{\hskip2pt}c@{\hskip2pt}@{\hskip2pt}c@{\hskip2pt}@{\hskip2pt}c@{\hskip2pt}@{\hskip2pt}c@{\hskip2pt}@{\hskip2pt}c@{\hskip2pt}}
		
		\begin{tikzpicture}\node[above right, inner sep=0](image) at (0,0) {\includegraphics[width=0.3\linewidth]{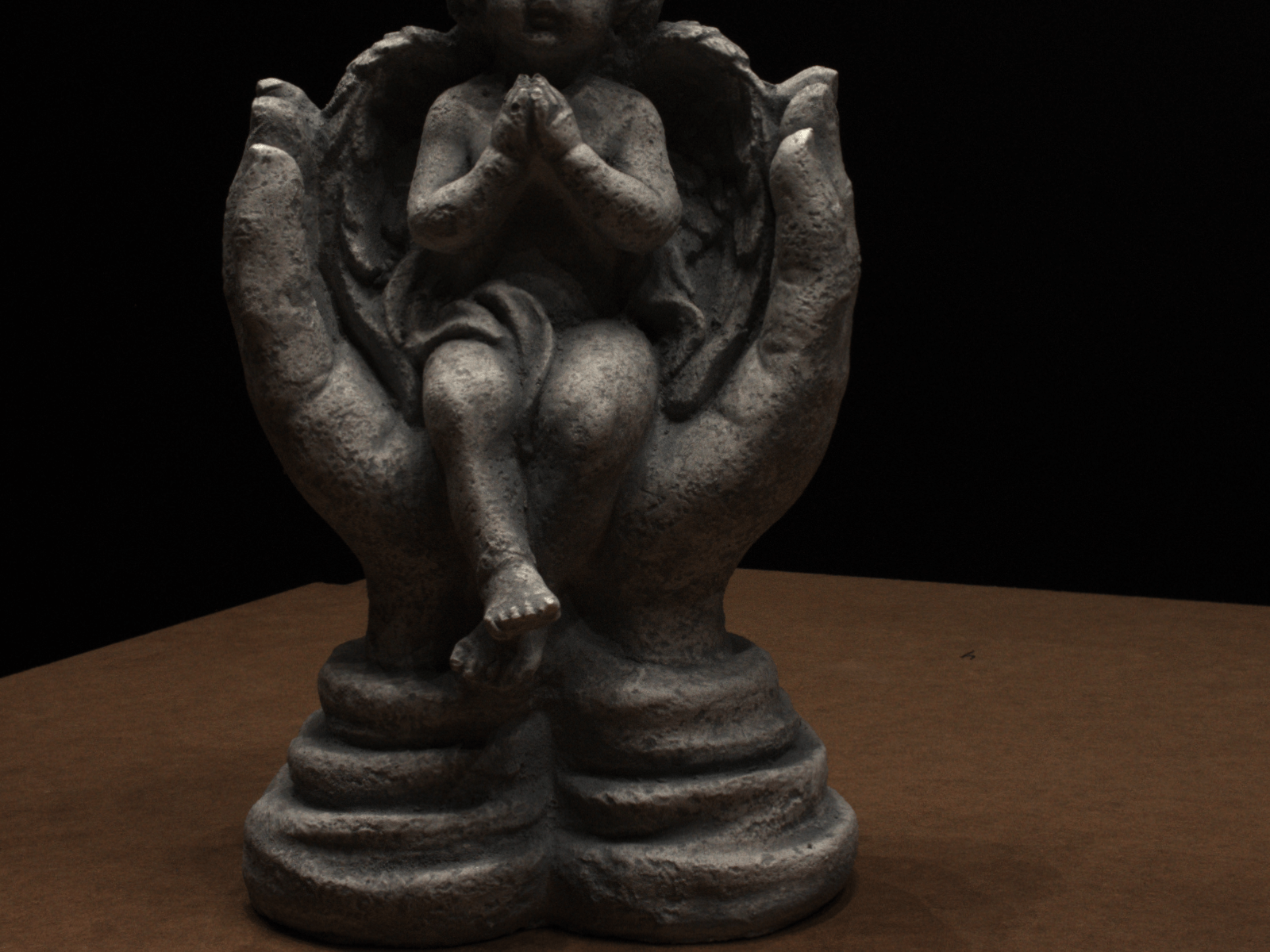}};
		\end{tikzpicture} &
		\begin{tikzpicture}\node[above right, inner sep=0](image) at (0,0) {\includegraphics[width=0.3\linewidth]{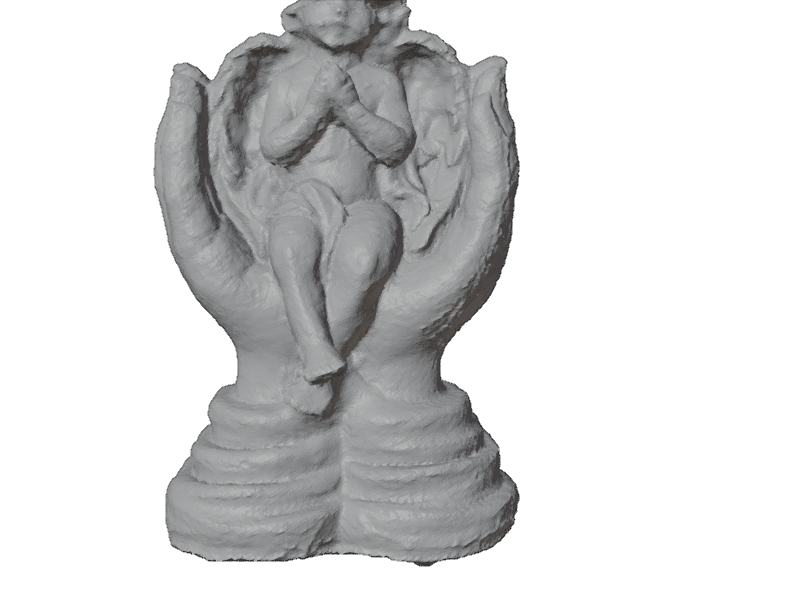}};
		\end{tikzpicture} &
		\begin{tikzpicture}\node[above right, inner sep=0](image) at (0,0) {\includegraphics[width=0.3\linewidth]{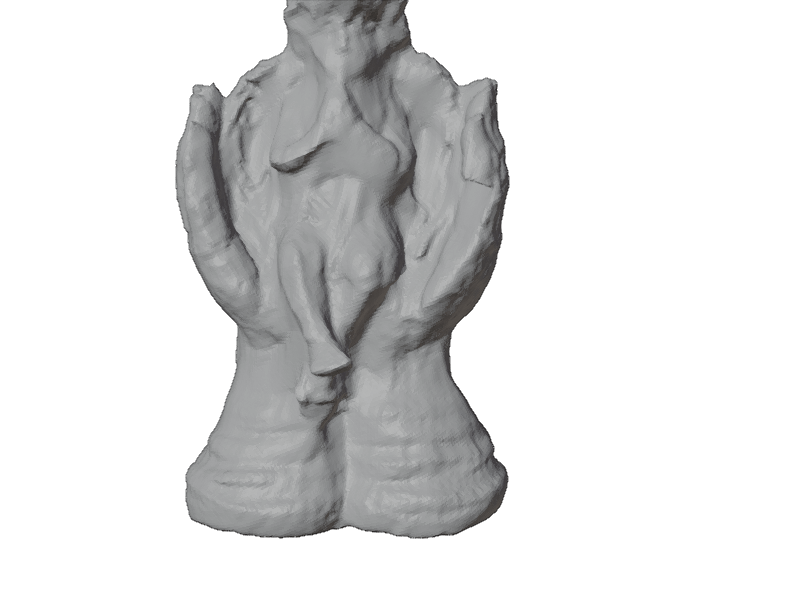}}; 
		\end{tikzpicture} & \\
		
		\begin{tikzpicture}\node[above right, inner sep=0](image) at (0,0) {\includegraphics[width=0.3\linewidth]{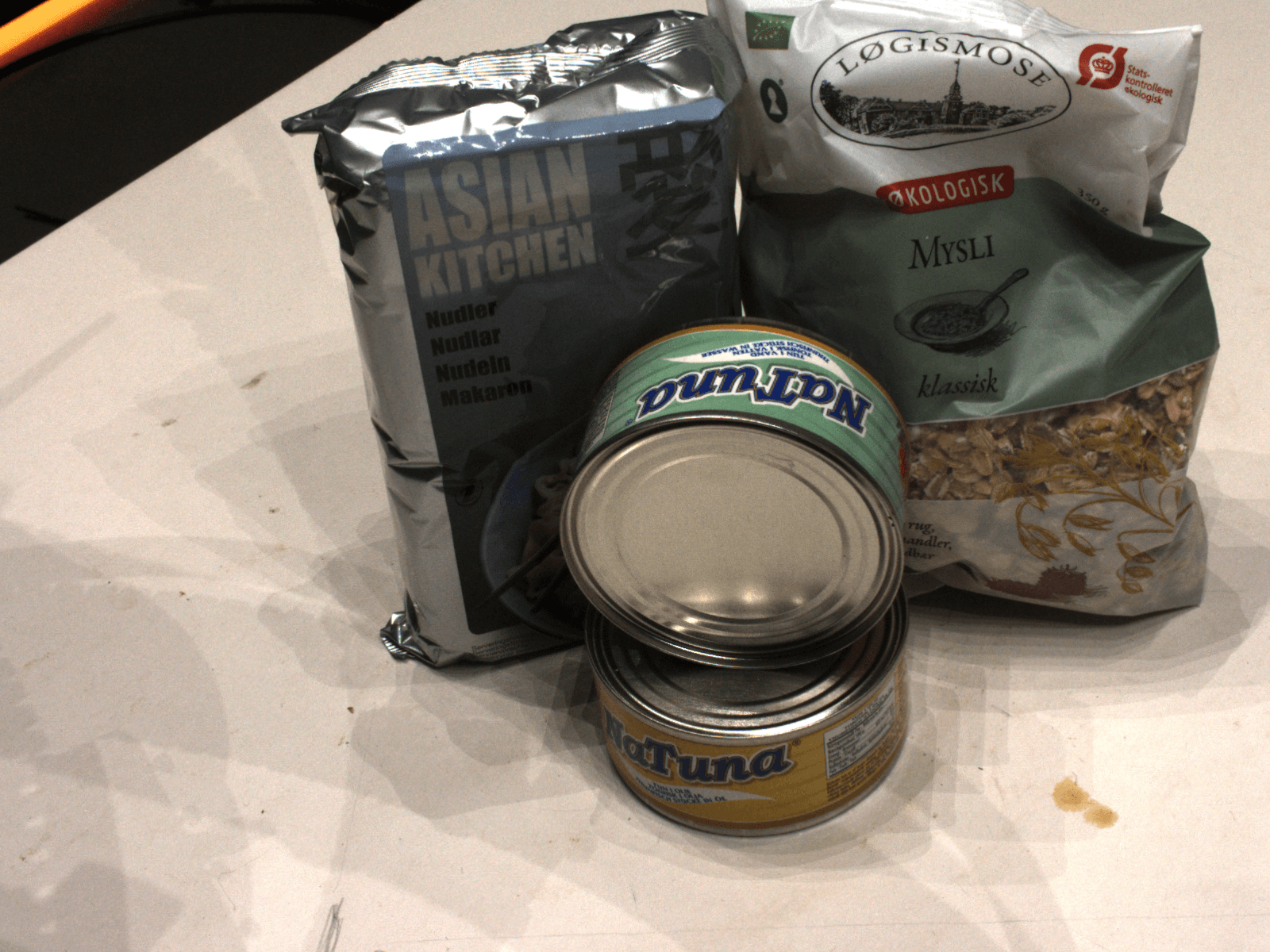}};
		\end{tikzpicture} &
		\begin{tikzpicture}\node[above right, inner sep=0](image) at (0,0) {\includegraphics[width=0.3\linewidth]{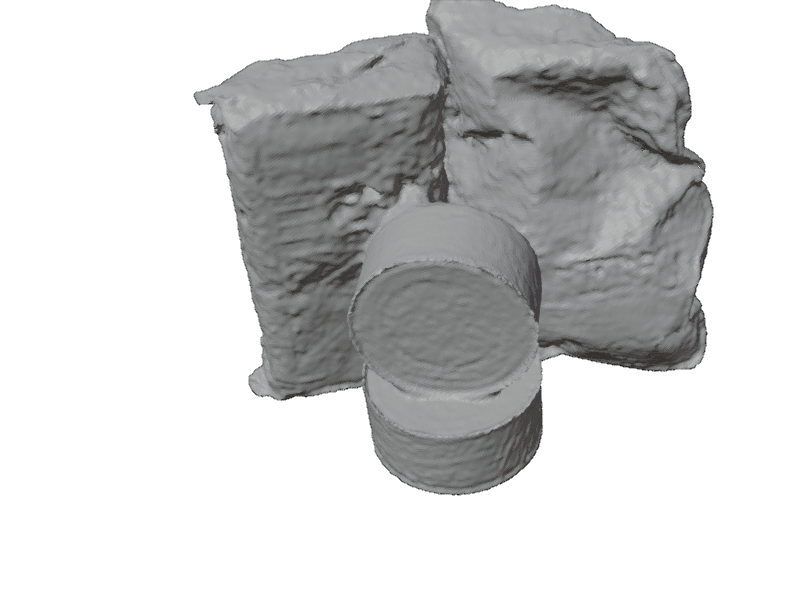}};
		\end{tikzpicture} &
		\begin{tikzpicture}\node[above right, inner sep=0](image) at (0,0) {\includegraphics[width=0.3\linewidth]{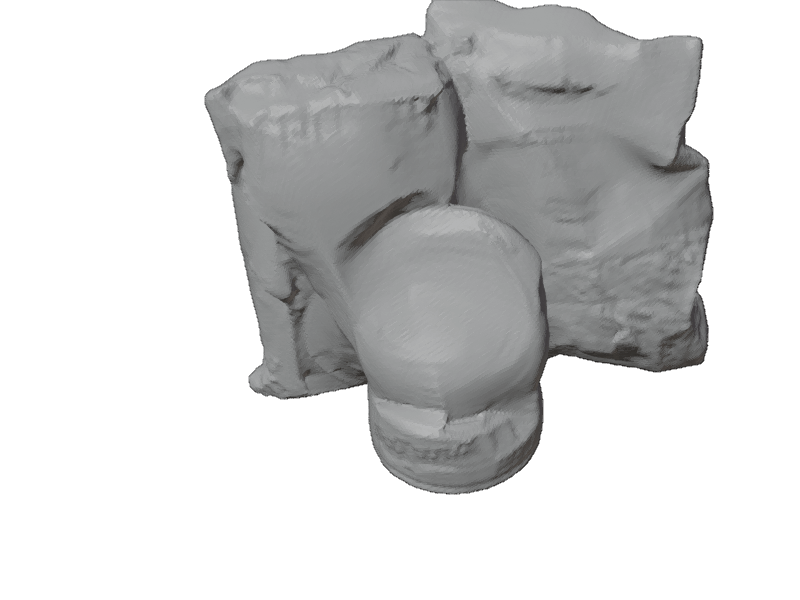}};
		\end{tikzpicture} & \\
		
		Original image & depth loss & rendering loss
	\end{tabular}
	\caption{Ablation studies on the depth loss. The mesh obtained by network rendering loss loses a lot of details. While the depth loss provides the extra geometry information so that the recovered mesh is accurate with the details.}
	\label{fig:ablation}
\end{figure}

In this section, we firstly discuss the effect of depth loss on the surface recovery. Previous methods~\cite{DBLP:conf/nips/idr20, DBLP:conf/cvpr/physg21} mainly try to minimize the rendering loss, where the shape is just a by-product. Although the geometry can be estimated accurately in some cases, it is usually difficult to recover the accurate shape due to the ambiguity between geometry and appearance. We replace the depth loss with the rendering loss used in previous work to learn the shape. Specifically, we employ an MLP to estimate the color of each pixel. The gradient is backpropagated to the shape through the proposed intersection points representation described in~\cite{DBLP:conf/nips/idr20}. The SDF value is obtained from the SDF grid by trilinear interpolation, which is optimized by the rendering loss. The qualitative results are shown in Fig.~\ref{fig:ablation}. We can get roughly correct results with rendering loss. When the topology or texture of the optimized mesh is complex, it is hard for the rendering loss to handle the ambiguity of shape and appearance. The depth maps predicted by multi-view stereo provide the correct geometry information, which result in more accurate results and enable the fast convergence. Table~\ref{tab:abl} shows the quantitative results. It can be seen that we cannot get the accurate shape using the rendering loss only.

\begin{figure}[htbp]
	\centering
	\begin{tabular}{@{\hskip2pt}c@{\hskip2pt}@{\hskip2pt}c@{\hskip2pt}@{\hskip2pt}c@{\hskip2pt}@{\hskip2pt}c@{\hskip2pt}@{\hskip2pt}c@{\hskip2pt}}
		
		\begin{tikzpicture}\node[above right, inner sep=0](image) at (0,0) {\includegraphics[width=0.3\linewidth]{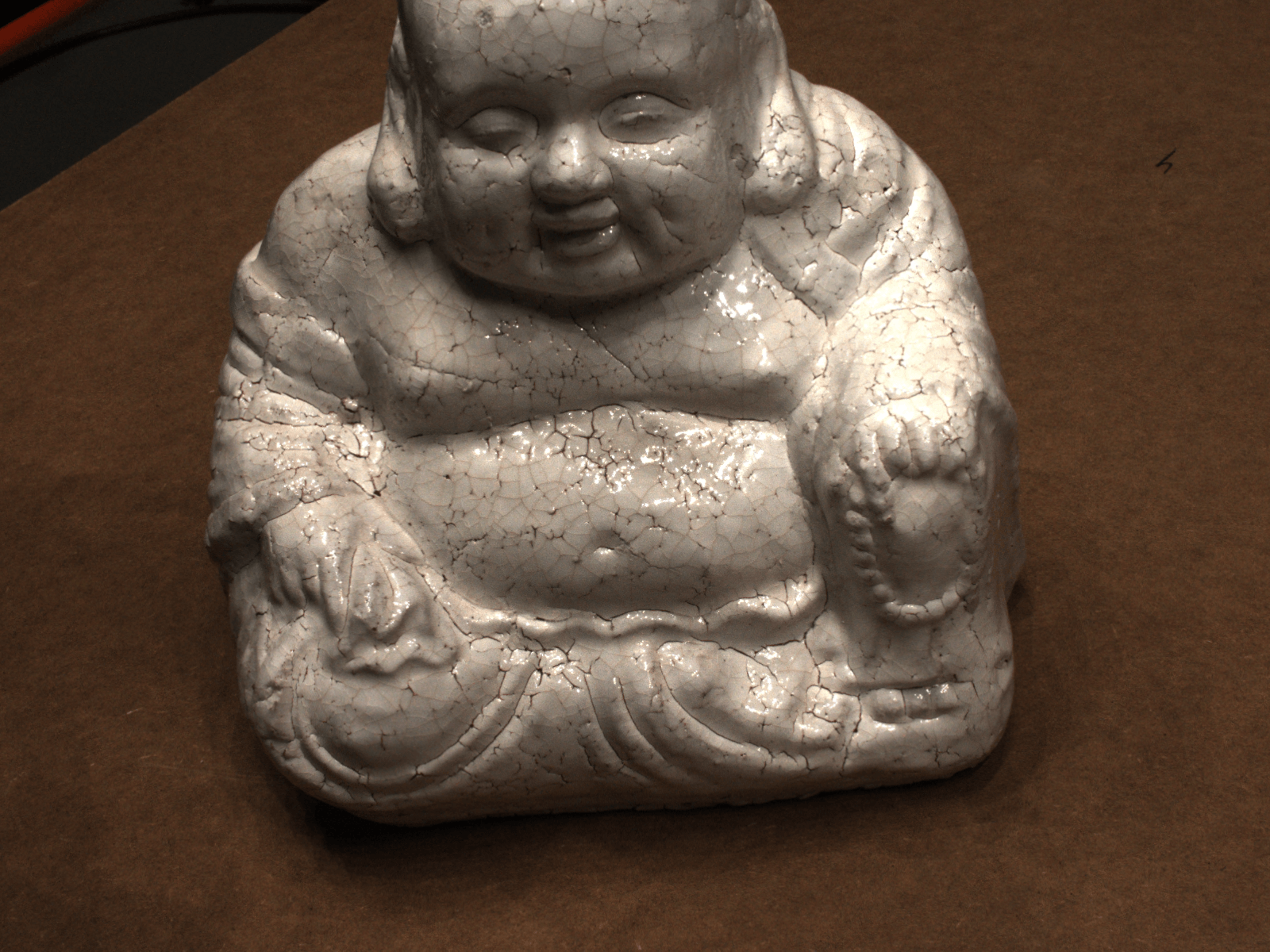}};
		\end{tikzpicture} &
		\begin{tikzpicture}\node[above right, inner sep=0](image) at (0,0) {\includegraphics[width=0.3\linewidth]{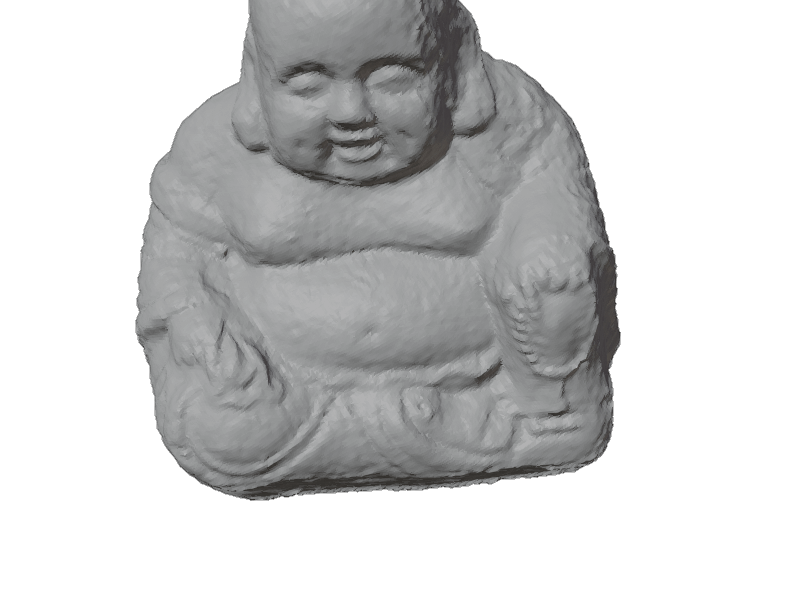}};
		\end{tikzpicture} &
		\begin{tikzpicture}\node[above right, inner sep=0](image) at (0,0) {\includegraphics[width=0.3\linewidth]{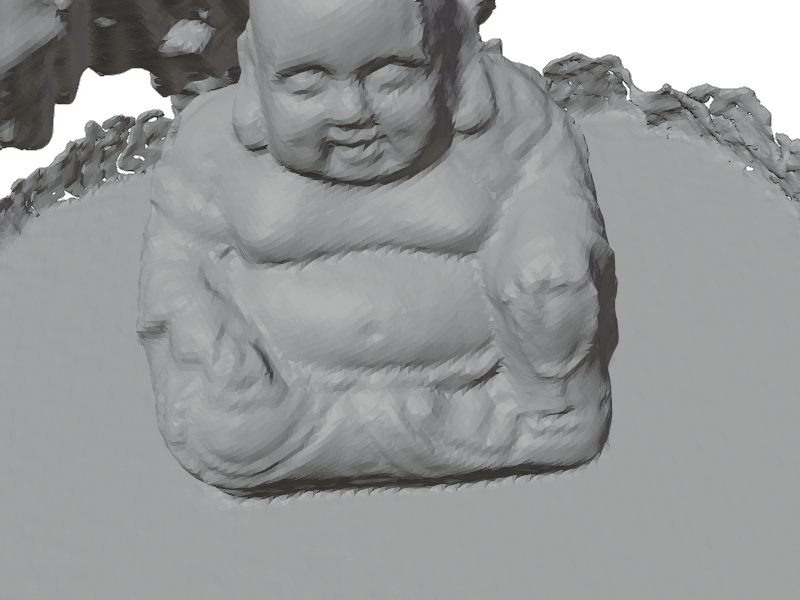}}; 
		\end{tikzpicture} & \\
		
		\begin{tikzpicture}\node[above right, inner sep=0](image) at (0,0) {\includegraphics[width=0.3\linewidth]{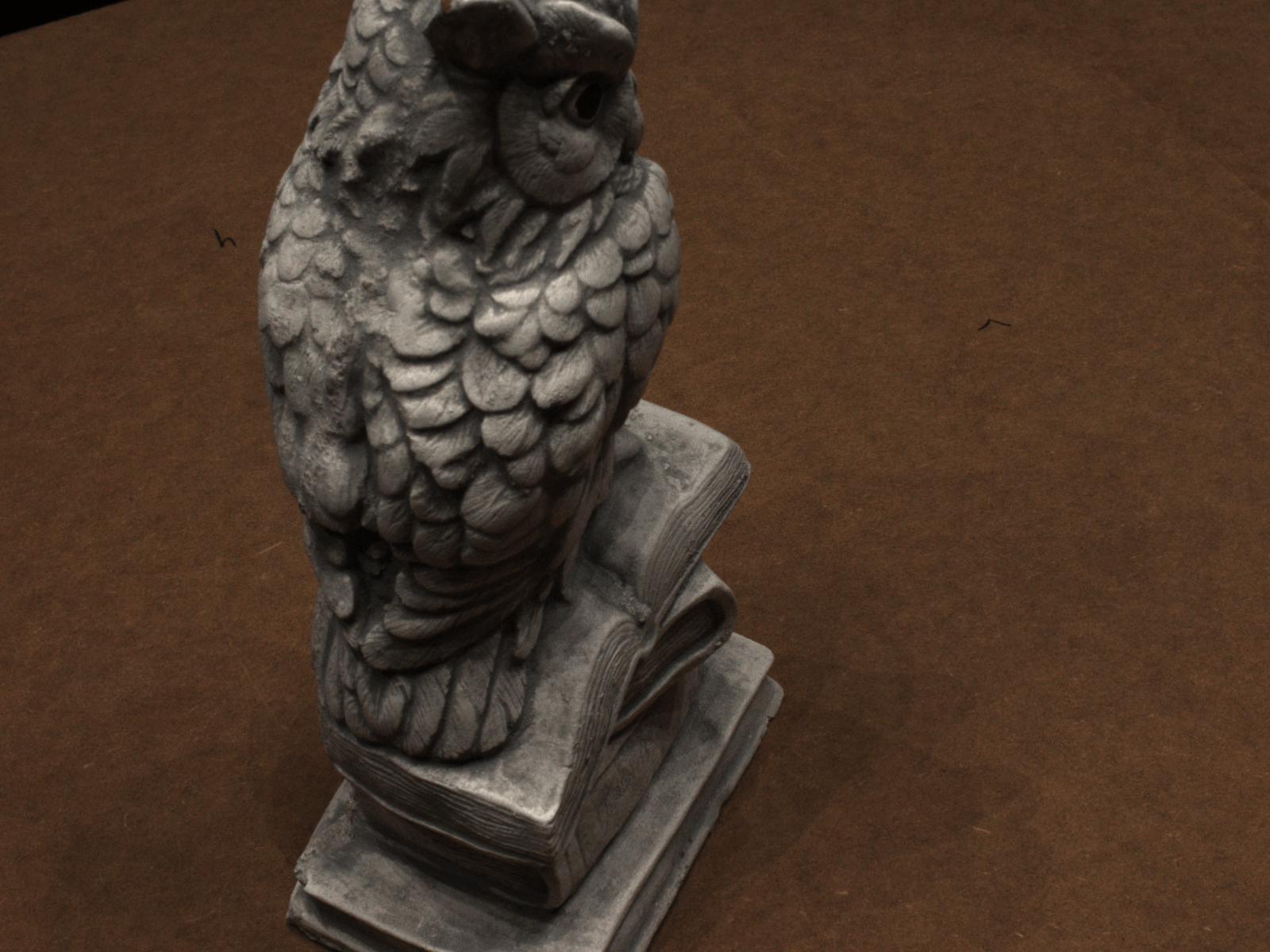}};
		\end{tikzpicture} &
		\begin{tikzpicture}\node[above right, inner sep=0](image) at (0,0) {\includegraphics[width=0.3\linewidth]{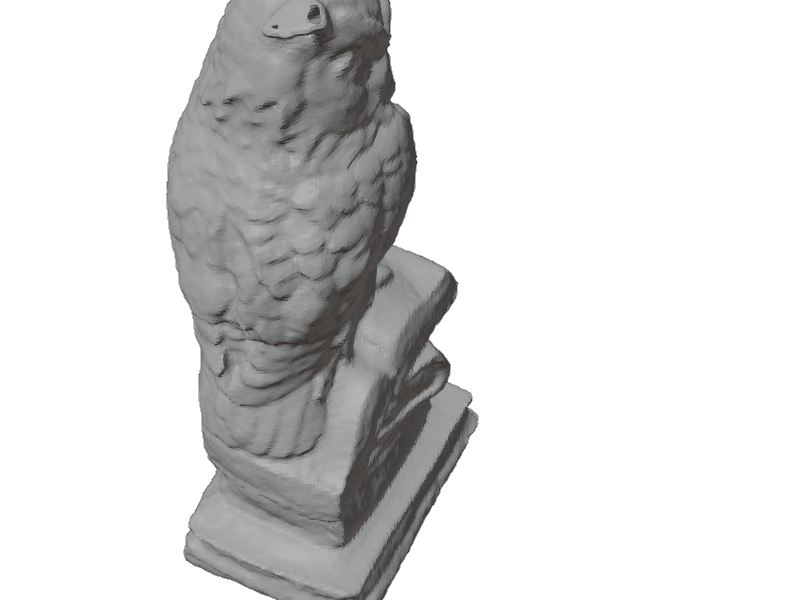}};
		\end{tikzpicture} &
		\begin{tikzpicture}\node[above right, inner sep=0](image) at (0,0) {\includegraphics[width=0.3\linewidth]{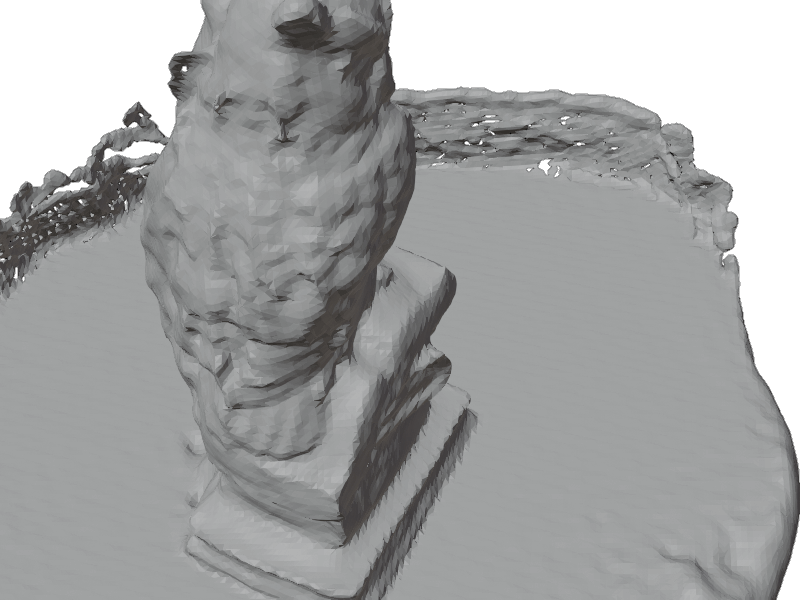}};
		\end{tikzpicture} & \\
		
		Original image & with mask & w/o mask
	\end{tabular}
	\caption{Ablation studies on the silhouette loss. Without silhouette constraints we can get the reconstruction of the whole scene. We can get more detailed reconstruction results by adding silhouette loss with the same resolution of SDF grid.}
	\label{fig:maskabl}
\end{figure}

We investigate the effectiveness of silhouette loss that helps to recover only the parts of interest. Without silhouette constraints, we may reconstruct the whole scene. As depicted in Table~\ref{tab:abl}, we will get higher mean Chamfer distance without mask supervision. This is because the Chamfer distance evaluation will be affected by the surroundings like floor. Although the Chamfer distance above a certain threshold will not be counted in evaluation metrics, there are still some reconstructed parts with the Chamfer distance less than the threshold. As the recovered mesh is extracted from an SDF grid at the resolution of $256^3$, the reconstruction results are affected by the size of scene. Fig.~\ref{fig:maskabl} shows the example of silhouette loss on the reconstruction results. It can be seen that our proposed method also can reconstruct correct results without mask supervision. The meshes obtained without mask supervision are rougher, since the reconstruction range is larger while the resolution of SDF grid does not change. 

\begin{table}[htbp]
	\centering
	\caption{Ablation studies on DTU dataset. We evaluate the effect of the visual hull initialization, depth loss, and silhouette loss.}
	
	\begin{tabular}{l|cccc|c}
		\toprule
		& $ \mathcal{L}_d $ & $ \mathcal{L}_r $ &  $ \mathcal{L}_c $ & $ \mathcal{L}_s $ & Mean Chamfer (mm) \\ \hline
		from sphere & $ \checkmark $  &    & $\checkmark$  & $\checkmark$ & 0.69        \\
		rendering loss  &   & $ \checkmark $  &   & $ \checkmark $   & 3.46         \\
		without mask & $ \checkmark $  &    &  $\checkmark$ &  & 0.97         \\
		full       & $ \checkmark $ &  & $ \checkmark $  & $ \checkmark $  & 0.68         \\
		\bottomrule
	\end{tabular}
	\label{tab:abl}
\end{table}

\begin{figure}[htbp]
	\centering
	\begin{tabular}{@{\hskip2pt}c@{\hskip2pt}@{\hskip2pt}c@{\hskip2pt}@{\hskip2pt}c@{\hskip2pt}@{\hskip2pt}c@{\hskip2pt}@{\hskip2pt}c@{\hskip2pt}}
		
		\begin{tikzpicture}\node[above right, inner sep=0](image) at (0,0) {\includegraphics[width=0.3\linewidth]{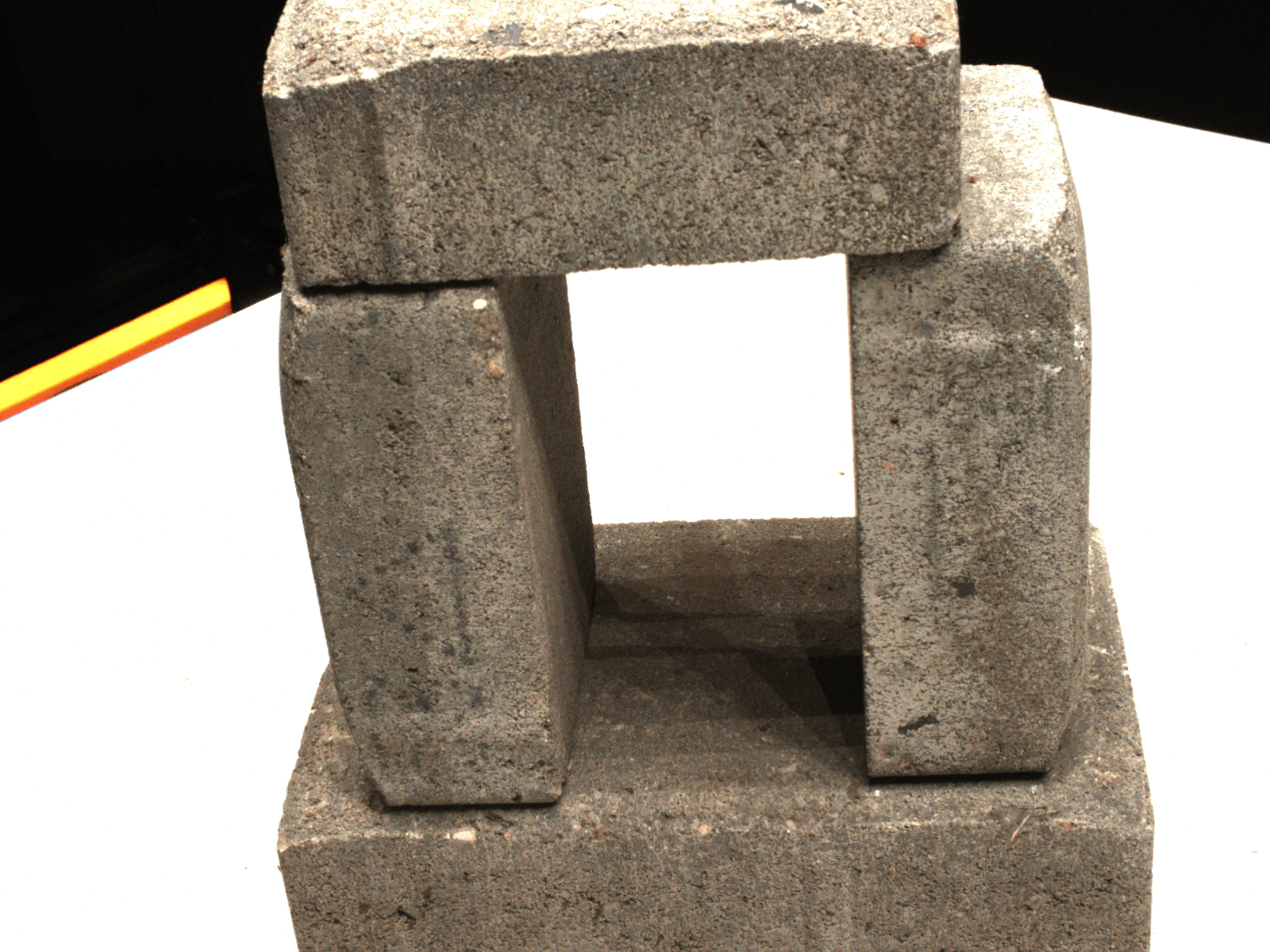}};
		\end{tikzpicture} &
		\begin{tikzpicture}\node[above right, inner sep=0](image) at (0,0) {\includegraphics[width=0.3\linewidth]{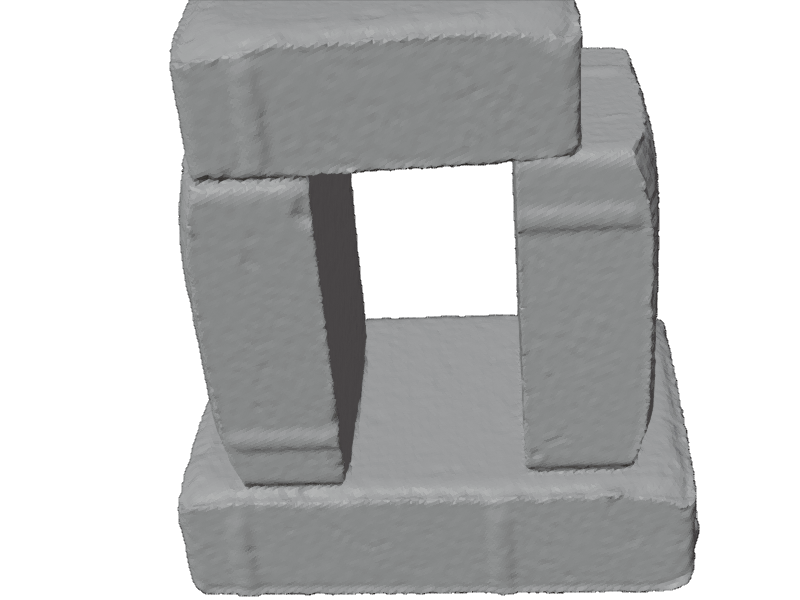}};
		\end{tikzpicture} &
		\begin{tikzpicture}\node[above right, inner sep=0](image) at (0,0) {\includegraphics[width=0.3\linewidth]{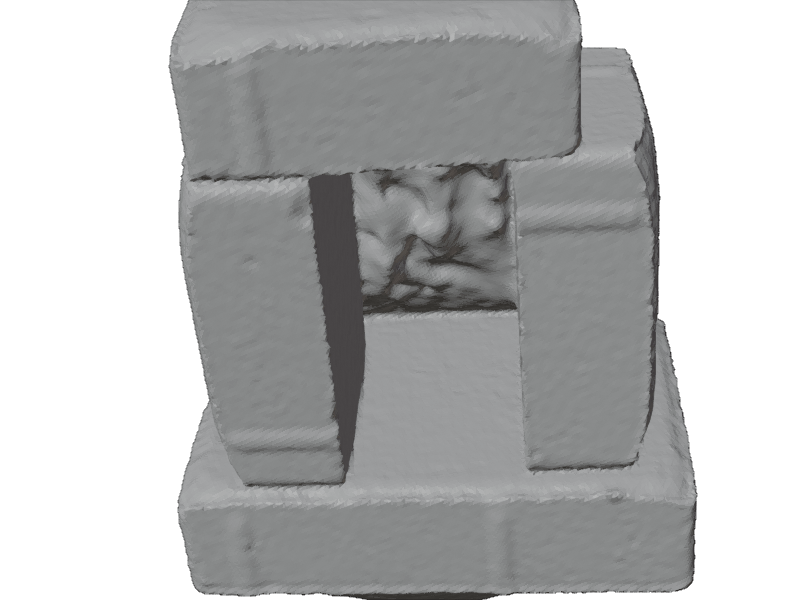}}; 
		\end{tikzpicture} & \\
		
		\begin{tikzpicture}\node[above right, inner sep=0](image) at (0,0) {\includegraphics[width=0.3\linewidth]{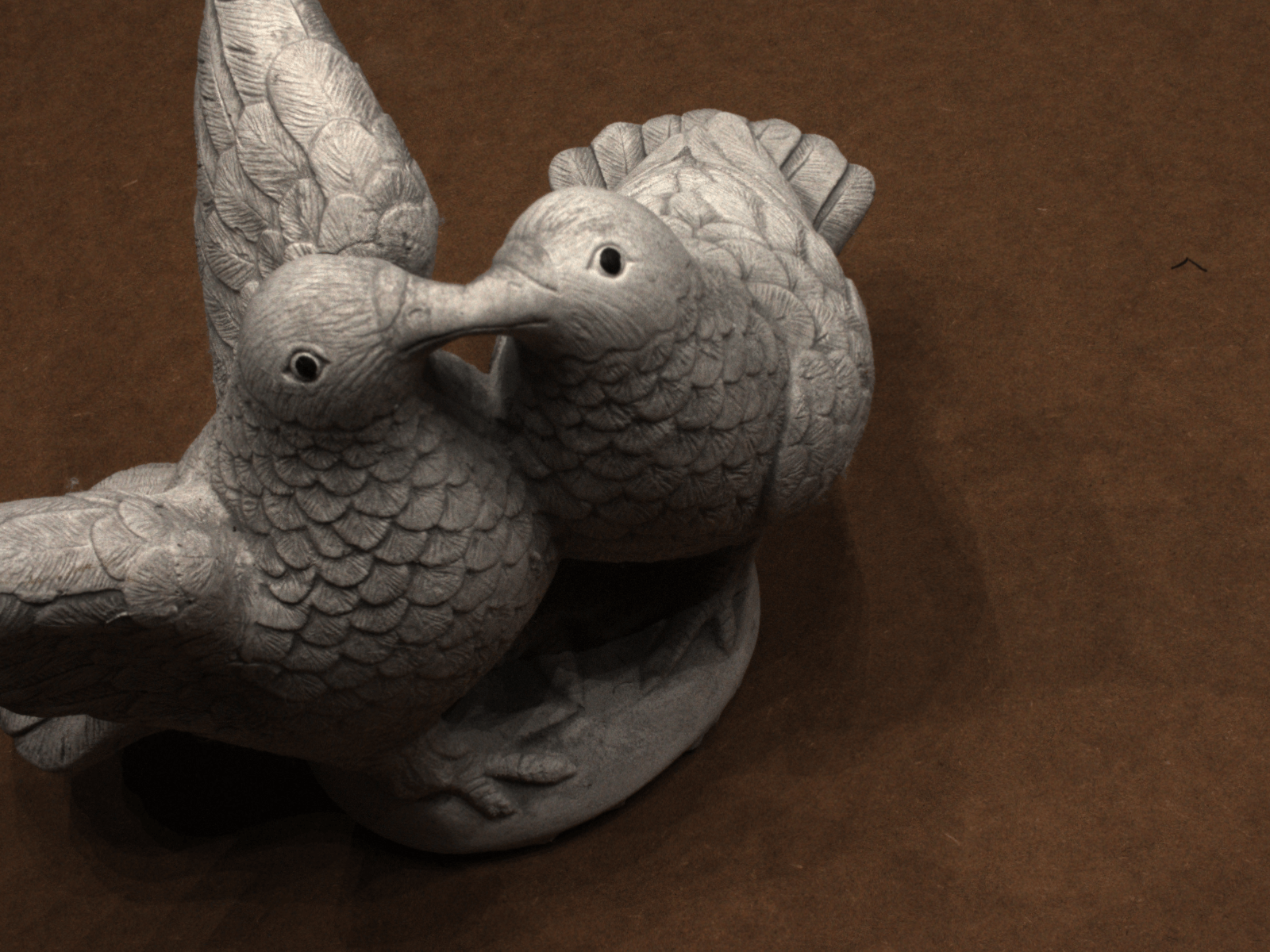}};
		\end{tikzpicture} &
		\begin{tikzpicture}\node[above right, inner sep=0](image) at (0,0) {\includegraphics[width=0.3\linewidth]{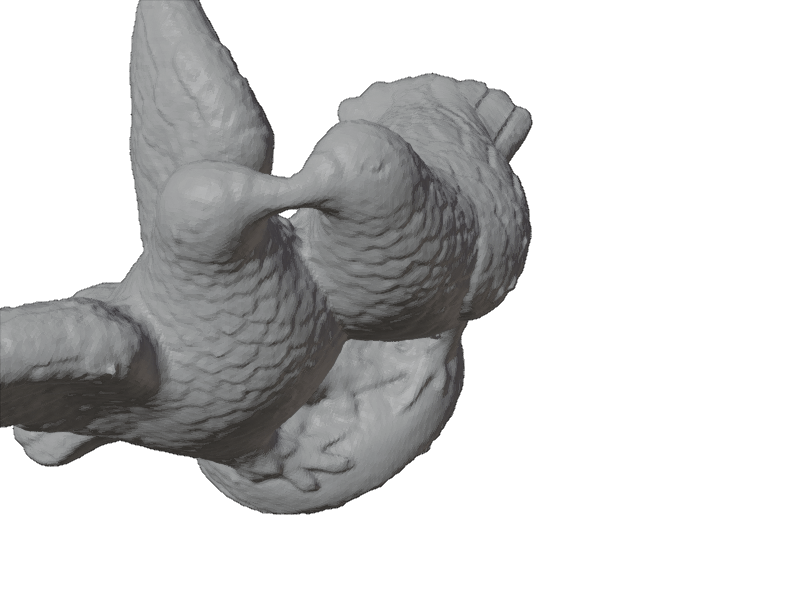}};
		\end{tikzpicture} &
		\begin{tikzpicture}\node[above right, inner sep=0](image) at (0,0) {\includegraphics[width=0.3\linewidth]{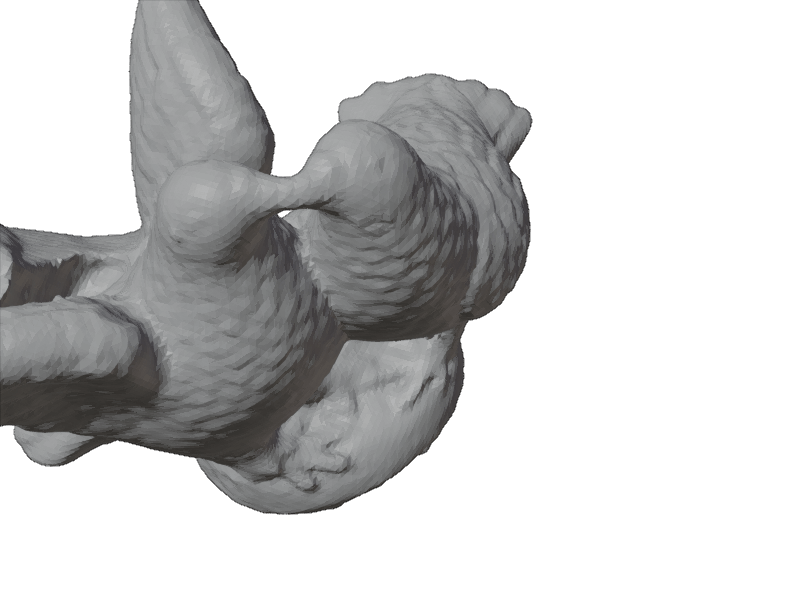}};
		\end{tikzpicture} & \\
		
		Original image & from visual hull & from sphere
	\end{tabular}
	\caption{Ablation studies on the initialization. We also can get accurate reconstruction results from sphere initialization. Since some views are not included in DTU dataset, the result obtained from the ball initialization still has some redundant parts.}
	\label{fig:sphabl}
\end{figure}

We also perform ablation study on the different initialization schemes. We compare the results of sphere and visual hull initialization. Results are shown in Table~\ref{tab:abl}. It indicates that visual hull initialization barely affects the reconstruction results. Moreover, we still can get the accurate meshes from the sphere initialization. Fig.~\ref{fig:sphabl} gives the example results. It is difficult to dig a hole from sphere initialization. This is because the gradients estimated by differentiable renderer is incorrect in this case. Since the view of DTU dataset does not fully cover the object, there is still some extra surfaces when deforming from the sphere.

\begin{table}[htbp]
	\centering
	\caption{Ablation studies on loss types. We evaluate the effect of the $L_1$ norm and $L_2$ norm of depth loss and silhouette loss.}
	
	\begin{tabular}{l|cccc|c}
		\toprule
		& $L_1$ $ \mathcal{L}_d $ &  $L_2$ $ \mathcal{L}_d $ &  $L_1$ $ \mathcal{L}_s $ & $L_2$ $ \mathcal{L}_s $ & Mean Chamfer (mm) \\ \hline
		& $ \checkmark $  &  &  & $\checkmark$ & 0.68        \\
		& $ \checkmark $  &   & $ \checkmark $  &   & 0.69    \\
		&  & $ \checkmark $  &  $\checkmark$ &  & 1.32         \\
		& & $ \checkmark $ &  & $ \checkmark $  & 1.31         \\
		\bottomrule
	\end{tabular}
	\label{tab:losstype}
\end{table}

\begin{figure}[htbp]
	\centering
	\begin{tabular}{@{\hskip2pt}c@{\hskip2pt}@{\hskip2pt}c@{\hskip2pt}@{\hskip2pt}c@{\hskip2pt}@{\hskip2pt}c@{\hskip2pt}@{\hskip2pt}c@{\hskip2pt}}
		
		\begin{tikzpicture}\node[above right, inner sep=0](image) at (0,0) {\includegraphics[width=0.3\linewidth]{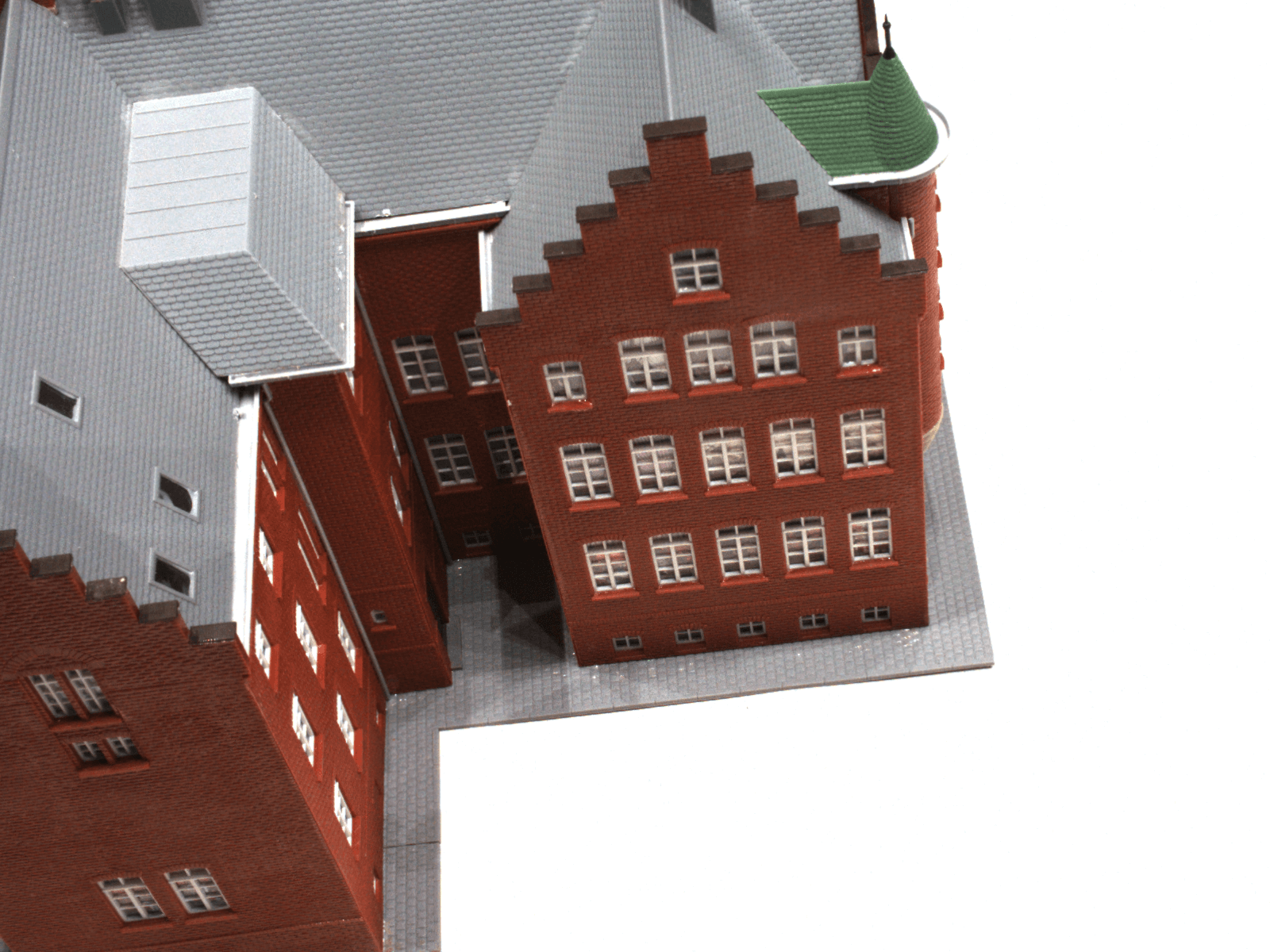}};
		\end{tikzpicture} &
		\begin{tikzpicture}\node[above right, inner sep=0](image) at (0,0) {\includegraphics[width=0.3\linewidth]{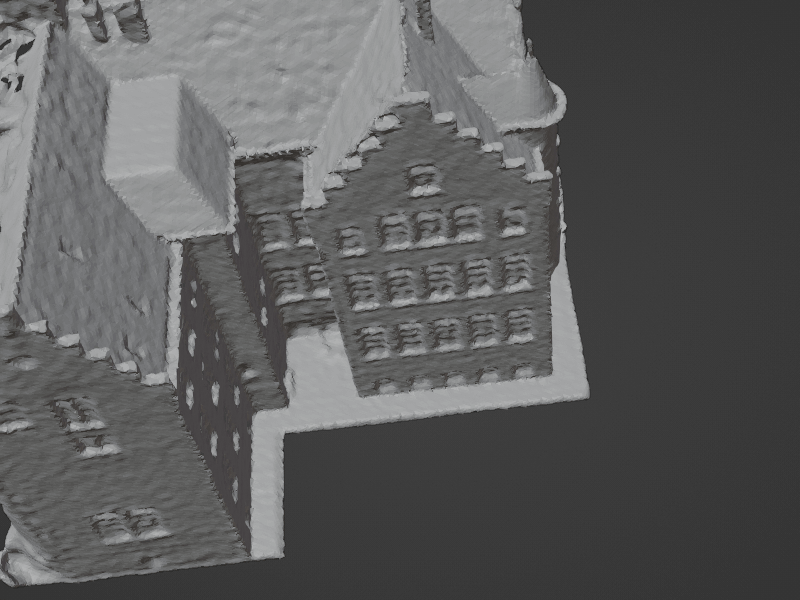}};
		\end{tikzpicture} &
		\begin{tikzpicture}\node[above right, inner sep=0](image) at (0,0) {\includegraphics[width=0.3\linewidth]{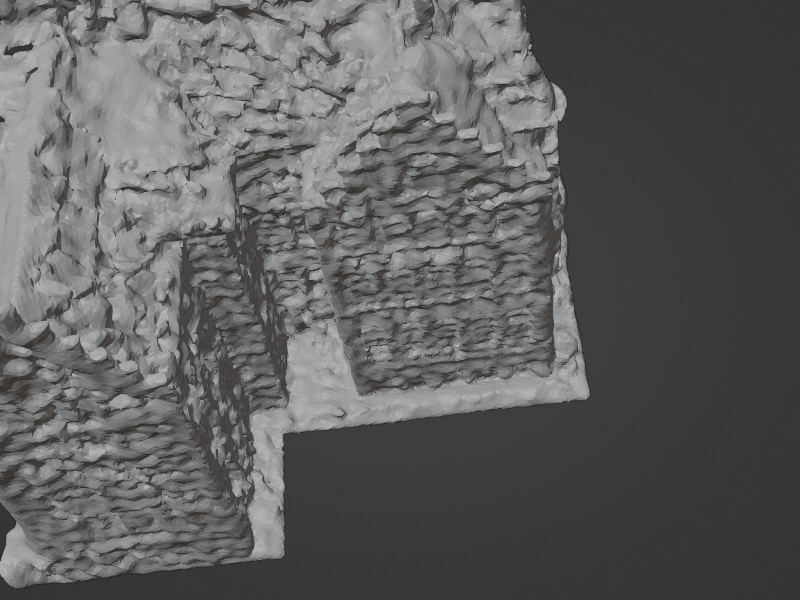}}; 
		\end{tikzpicture} & \\
		
		\begin{tikzpicture}\node[above right, inner sep=0](image) at (0,0) {\includegraphics[width=0.3\linewidth]{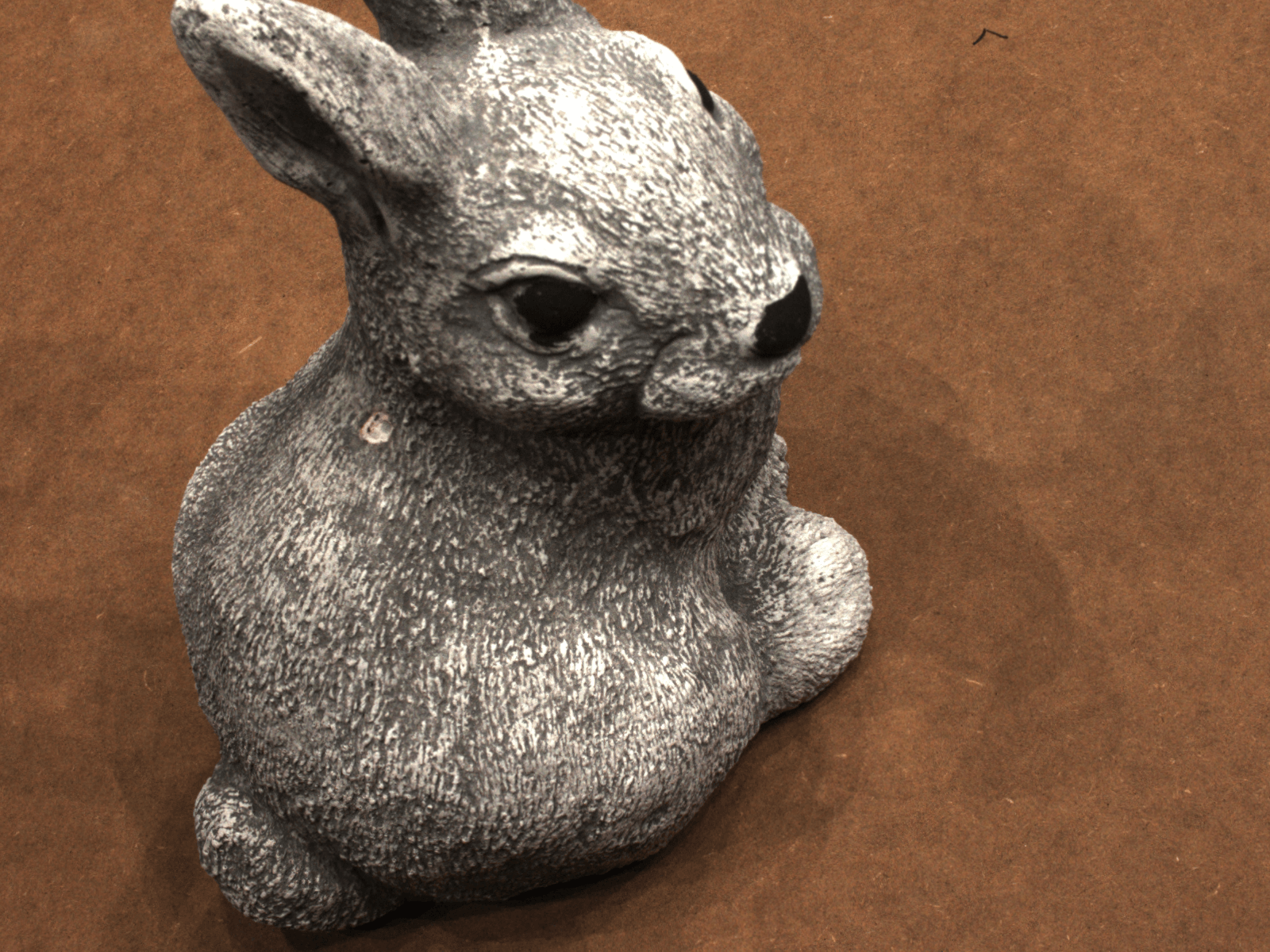}};
		\end{tikzpicture} &
		\begin{tikzpicture}\node[above right, inner sep=0](image) at (0,0) {\includegraphics[width=0.3\linewidth]{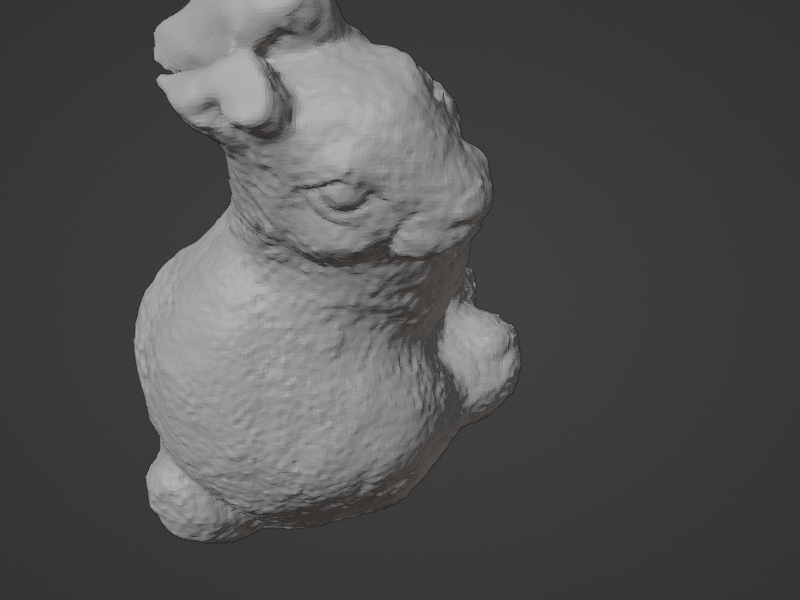}};
		\end{tikzpicture} &
		\begin{tikzpicture}\node[above right, inner sep=0](image) at (0,0) {\includegraphics[width=0.3\linewidth]{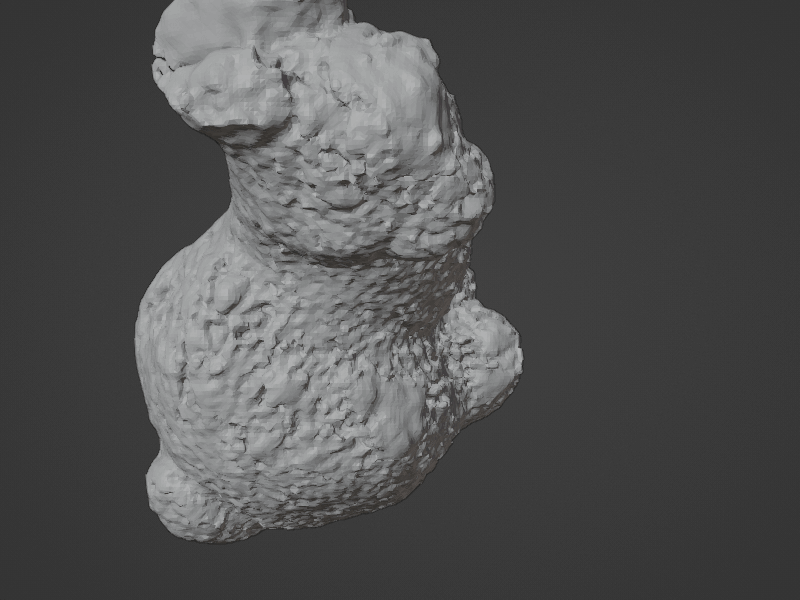}};
		\end{tikzpicture} & \\
		
		Original image & L1 depth loss & L2 depth loss
	\end{tabular}
	\caption{Ablation studies on the depth loss. L1 norm depth loss leads to smooth and accurate reconstruction results, while L2 norm depth loss generates rough and incorrect results.}
	\label{fig:depabl}
\end{figure}

Moreover, we conduct ablation study on different losses. We keep the same weight of the loss, and only change the type of depth loss and silhouette loss. Table~\ref{tab:losstype} shows the quantitative results. It can be seen that the type of depth loss has great influence on the results while the type of silhouette loss has little affect on the results. This may be due to the small value of $L_2$ norm depth loss, resulting in insufficient supervision of the reconstruction results. We use the visual hull initialization so that the silhouette loss is small at the beginning. It has barely effects on the results. The qualitative results are shown in Fig.~\ref{fig:depabl}. It can be seen that we can get smooth and accurate results with $L_1$ norm depth loss while the results of depth loss become rough and incorrect using $L_2$ norm.

\section{Conclusions}

In this paper, we proposed an efficient coarse-to-fine approach to the textured mesh recovery from multi-view images. Oriented point clouds with a differentiable Poisson Solver was used to represent the shape, which produces the topology-agnostic and watertight surfaces. The texture of reconstructed mesh was interpolated from a learnable texture grid. Instead of using the conventional MLP-based neural rendering, we introduced a physically-based inverse rendering scheme to jointly estimate the lighting and reflectance, which is able to render the high resolution image at real-time. We have conducted the extensive experiments on several multiview stereo datasets. The encouraging results showed that our approach can effectively reconstruct the textured mesh.


\bibliographystyle{IEEEtran}
\bibliography{egbib}

\end{document}